\documentclass[11pt]{article}

\usepackage[final]{acl}
\usepackage{comment}
\usepackage{times}
\usepackage{latexsym}
\usepackage{subcaption}
\usepackage{float}
\usepackage{tabularx}
\usepackage[T1]{fontenc}

\usepackage[utf8]{inputenc}

\usepackage{microtype}

\usepackage{inconsolata}

\usepackage{graphicx}

\usepackage{microtype}
\usepackage{hyperref}
\usepackage{url}
\usepackage{booktabs}
\usepackage{multirow}
\usepackage{array}
\usepackage{subcaption}
\usepackage{longtable}
\usepackage{marvosym}
\usepackage{amsmath,amsfonts,bm}
\usepackage{subcaption}
\usepackage{stfloats}

\usepackage{tabularx}
\usepackage{caption}

\title{Paid Voices vs. Public Feeds: Interpretable Cross-Platform Theme-Based Analysis of Climate Discourse} 

\author{
 \textbf{Samantha Sudhoff\thanks{These authors contributed equally.}}, \textbf{Pranav Perumal}, \textbf{Zhaoqing Wu}, \textbf{Tunazzina Islam\footnotemark[1]}
\\
Department of Computer Science, Purdue University, West Lafayette, IN 47907
\\
 \small{
  \texttt{\{ssudhoff, pperuma, wu1828, islam32\}{ @purdue.edu}}
 }
}

\begin{document}
\maketitle
\begin{abstract}
Climate discourse online shapes public understanding of climate change and informs political and policy debate, yet it unfolds across structurally different environments: paid advertising platforms host targeted, institutionally produced messaging, while public social media reflects largely organic, user-driven discussion. 
We present a comparative analysis of climate discourse across paid advertisements on Meta (previously Facebook) and public posts on Bluesky from July 2024 to September 2025. To support it, we develop an interpretable thematic discovery pipeline that clusters texts by semantic similarity and uses large language models (LLMs) to label clusters with concise, human-interpretable themes, requiring \textbf{no} predefined topic inventory or seed set. 
Using these themes, we find the two environments diverge systematically: paid advertising centers on strategic promotion of specific solutions in a formal, forward-looking register, whereas organic discourse centers on systemic critique in a crisis-oriented, scientifically grounded one. 
We also evaluate the utility of the discovered themes through downstream \textit{stance prediction} and \textit{theme-guided retrieval} tasks.
While our analysis focuses on climate communication, the framework generalizes to comparative thematic analysis across heterogeneous communication environments.
\end{abstract}
\section{Introduction}
Online platforms have become a primary arena for climate communication, shaping how climate change is discussed, debated, and acted upon by the public \cite{pera2024shifting,islam2023analysis,bloomfield2019circulation,pearce2019social,abbar2016using}. Climate narratives circulated through social media influence political attitudes, policy preferences, and perceptions of scientific credibility \cite{liu2025scientists,vivion2024misinformation,holder2023climate,treen2020online,lejano2020power,walter2018echo,gupta2018advocacy}, making climate discourse a central object of study for computational social science (CSS) and natural language processing (NLP).

\begin{figure}[t]
\includegraphics[width=.95\columnwidth]{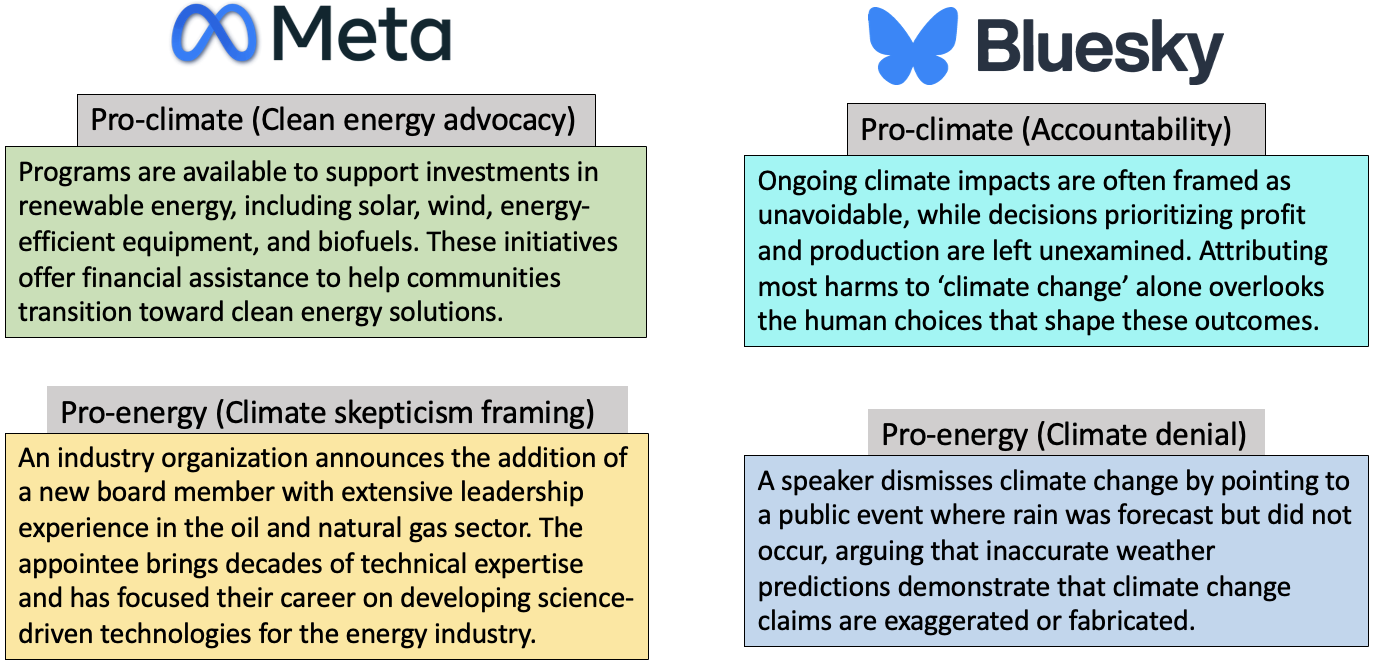}
\caption{Examples of climate discourse across paid and public platforms (paraphrased and anonymized).}
\vspace{-5pt}
    \label{fig:motiv}
\vspace{-5pt}
\end{figure}
However, climate communication does not unfold in a uniform online 
environment. Paid advertising ecosystems such as Meta's advertising 
platform \cite{meta_ad_library} are designed for strategic, targeted 
persuasion in service of political or economic goals 
\cite{islam2023analysis,islam2023weakly,islam2022understanding,capozzi2021clandestino,ribeiro2019microtargeting}, 
while public social media platforms such as Bluesky \cite{bluesky_platform} 
host largely user-driven discourse shaped by personal expression and 
grassroots interaction \cite{quelle2024bluesky}. We treat paid advertising 
as a proxy for institutionally produced, resource-backed messaging and 
public posting as a proxy for organic, user-driven discourse; Fig.~\ref{fig:motiv} illustrates the contrast (motivation only---all empirical claims come from later analyses). Prior computational works have examined climate discourse within individual platforms, focusing on tasks such as topic modeling \cite{dahal2019topic}, stance detection \cite{upadhyaya2023intensity,luo-etal-2020-detecting}, and targeted messaging analysis \cite{islam2025post,islam2025uncovering,islam2025discovering,islam2023analysis}, 
leaving unclear whether thematic patterns are shared across communication environments or specific to where they emerge.

\begin{figure}[t]
\includegraphics[width=1.0\columnwidth]{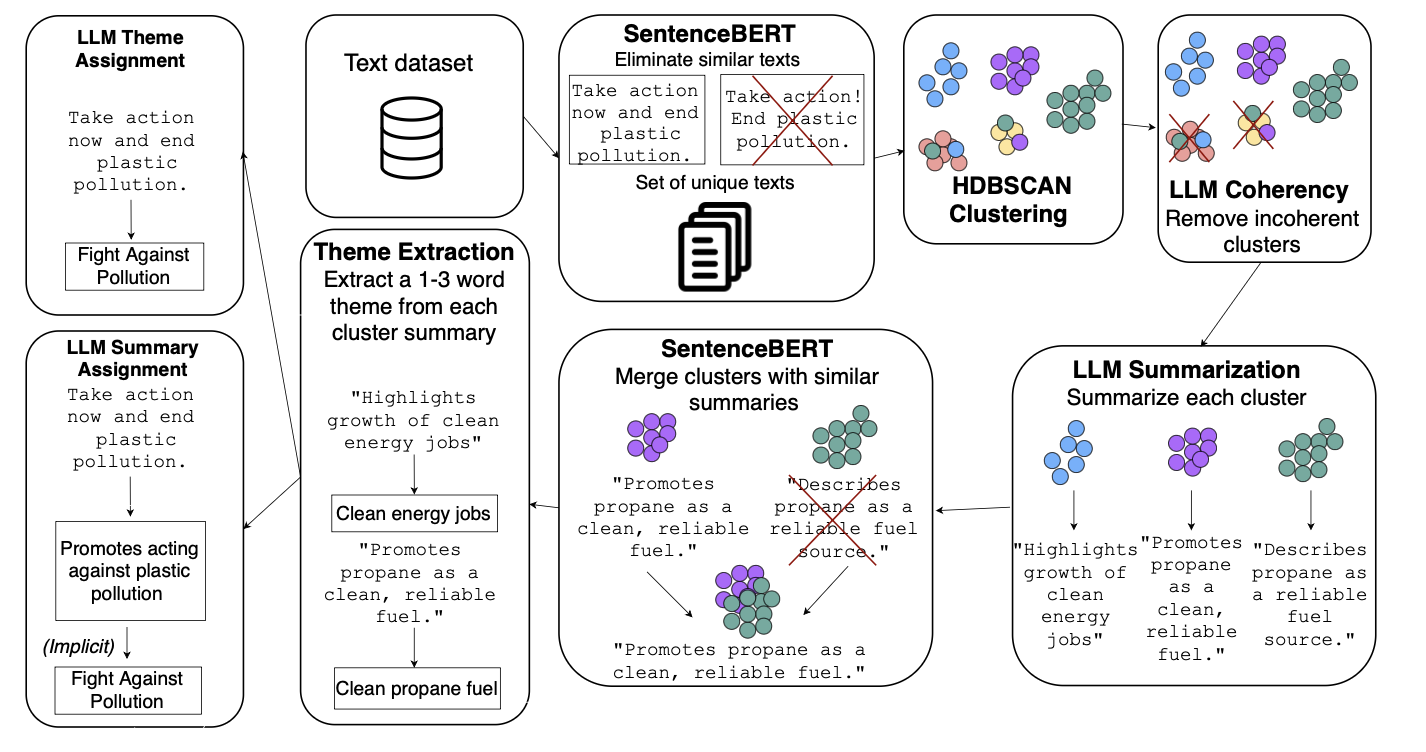}
    \caption{Overview of our framework.}
    \vspace{-5pt}
    \label{fig:pipeline}
    \vspace{-5pt}
\end{figure}

In this paper, we conduct a comparative analysis of climate discourse across paid advertisements on Meta and public posts on Bluesky. Rather 
than proposing a new topic modeling algorithm, we develop a measurement 
framework that supports cross-platform thematic analysis despite 
differences in vocabulary, style, and communicative norms. Our approach first discovers latent semantic clusters and then uses large language models (LLMs) \cite{brown2020language} to generate concise, human-interpretable theme labels. By separating semantic discovery from linguistic interpretation, the framework keeps themes grounded in the data while making them accessible for substantive analysis and downstream use. It also avoids reliance on predefined topic inventories or seed sets (Fig.~\ref{fig:pipeline}). We evaluate the resulting themes against standard topic modeling baselines---LDA \cite{blei2003latent}, BERTopic \cite{grootendorst2022bertopic}, and TopicGPT \cite{pham2024topicgptpromptbasedtopicmodeling}---using both human judgments and an LLM-based evaluator, and probe their utility on two downstream tasks, \textit{stance prediction} and \textit{theme-guided retrieval}, as validation probes for whether the themes are coherent semantic abstractions. We then use the themes to examine thematic prevalence across platforms and how discourse shifts around major real-world events. Our goal is not to make causal claims about platform effects or speaker intent, but to descriptively characterize how climate discourse differs across two communication environments.

Across our analyses, we find that the two environments reflect distinct
communicative orientations. Paid advertising concentrates on the strategic promotion of specific solutions, whereas organic
discourse concentrates on the systemic critique of corporate and political actors. This contrast recurs across thematic structure, rhetorical register, and stance alignment, and the two platforms respond to major events in opposite directions: advertising
contracts following the 2024 U.S. election while public discourse expands.

In summary, this paper makes three primary contributions\footnote{Our datasets and code are publicly available here: \url{https://github.com/srs827/PaidVoices-vs-PublicFeeds}}:
\newline
(1) We construct and release the first multi-platform dataset enabling direct comparison between paid climate advertising and public climate discourse.
\newline
(2) We develop an interpretable thematic discovery pipeline for cross-platform thematic analysis that does not require predefined topic inventories or seed themes.
\newline
(3) We conduct quantitative and qualitative analysis. Our findings show that paid and public climate discourse differ systematically in thematic structure, stance alignment, and temporal responsiveness, illustrating the value of comparative thematic analysis across heterogeneous communication environments.

\section{Related Work}
\textbf{Climate communication and political advertising.}
Prior works examine how climate and political campaigns operate in paid advertising ecosystems. Studies analyze messaging strategies, stance, and targeting in climate-related ads using weak supervision, graph-based advertiser–ad models, and LLM-based analysis \cite{islam2025post,rowlands2024predicting,islam2023analysis,islam2023weakly}. 
The most closely related concurrent work is \cite{cheng2026crossplatformclimateclaims}, which analyzes cross-platform climate discourse by extracting and classifying claims using a Toulmin-inspired argumentative framework \cite{toulmin1958uses}. In contrast, our work takes a coarser-grained, theme-level perspective, examining how climate messaging strategies rather than individual argumentative moves differ across platform contexts. These studies motivate treating paid advertising as a distinct, strategically and institutionally produced communication environment. 
\newline
\textbf{LLMs for theme and narrative discovery.}
Recent work uses LLMs to discover interpretable latent structure in large corpora \cite{islam2026reasoning,brady2025latent,nakshatri2025talkingpointbasedideological}. \citet{islam2025discovering} cluster documents from an initial seed and use LLMs to summarize clusters and label themes; others uncover finer-grained units such as arguments from a predefined theme set \cite{islam2025uncovering}, and concept-induction methods like LLooM \cite{lam2024concept} incorporate human-provided seed concepts. In contrast, our pipeline performs thematic discovery \textbf{without} any seed set.
\newline
\textbf{Framing analysis of climate discourse.}
Framing theory \cite{entman1993framing} describes how communicators emphasize certain aspects of reality to promote particular framing functions. Computational work operationalizes framing through cross-issue codebooks and annotated corpora \cite{boydstun2014tracking, card2015media} and (weakly) supervised frame classification \cite{roy2020weakly, field2018framing}. In the climate domain, framing analyses examine climate and energy policy narratives \cite{vikstrom2023framing} and how multimodal climate communication shapes public understanding \cite{dancygier2023multimodal}. We adopt Entman's four functions across two communicative environments.
\newline
\textbf{Linguistic and psycholinguistic analysis.}
Beyond framing, computational social science has used psycholinguistic features---particularly LIWC \cite{boyd2022liwc}---to characterize how climate discourse varies across regional and organizational contexts \cite{wuraola-etal-2023-linguistic,GULLIVER2021101563}. We extend this work by applying framing lexicons and LIWC variables to enable theme-level rhetorical-register comparison across paid and organic communication.
\newline
\textbf{Stance detection.}
Stance detection is widely used to analyze climate narratives and ideological framing \cite{luo-etal-2020-detecting}, including detection of denial and neutralization techniques in climate skepticism \cite{bhatia2021automatic}. This motivates our stance-based validation: meaningful themes should align systematically with stance labels. 
\newline
\textbf{Cross-platform discourse analysis.}
Recent datasets enable comparison of political advertising across platforms such as Meta and Google \cite{zhang2025comparable}, and other work tracks how climate communication shifts across ecosystems such as YouTube and TikTok \cite{pera2024shifting} or within decentralized platforms like Bluesky \cite{quelle2024bluesky}. However, paid advertising and organic discourse are typically studied separately. We bridge this gap by jointly analyzing themes, framing, register, and temporal dynamics in Meta ads and Bluesky posts.

\section{Dataset}
We analyze climate discourse across two structurally distinct platforms. The Meta Ad Library contains paid advertisements designed for strategic persuasion and includes disclosed advertiser identities and metadata (e.g., impressions, spend, targeting). In contrast, Bluesky is a decentralized social media platform where climate-related content is generated organically by users without monetary promotion or targeting, reflecting grassroots discussion and personal expression.

We collect\footnote{\textcolor{red}{The social media data used in this study may contain offensive, biased, toxic, or harmful language.}} a set of over $113K$ climate-related ads from Meta Ad Library and $1.3M$ Bluesky climate posts from July 2024 to September 2025 using a curated set of climate-related keywords shown in Table \ref{tab:climate-keywords} App. \ref{app:kw}. The climate keywords are taken from a selection constructed by \citet{islam2023analysis} and modified to reflect recent shifts in climate-change related topics.  
Using the Meta Ad Library, a total of $113,493$ climate related ads are collected, which is then reduced to a dataset of size $17,026$ upon using SentenceBERT \cite{reimers-2019-sentence-bert} with a cosine similarity threshold of $0.80$ to eliminate highly similar ads. 

Bluesky, due to its nature as a social networking platform, has many more posts ($\approx 1.3$ million) that match our keywords within the same time frame. From the obtained post database, we randomly sample $20,000$ posts to serve as a similarly-sized comparison to the retrieved Meta dataset. We furthermore apply the same deduplication procedure, yielding $19,182$ unique Bluesky posts. We apply a greedy deduplication strategy for both Meta and Bluesky: for each text, all subsequent texts exceeding the similarity threshold are removed, retaining the first occurring text. The Bluesky sample maintains temporal representativeness of the full corpus. Across all months, the absolute difference between population and sample proportions $\leq 0.39\%$. Furthermore, human validation on $500$ ads and $500$ posts confirms $94.6\%$ are climate-relevant (App.~\ref{app:kw}).

\section{Methodology}
We develop a multi-stage pipeline for discovering the thematic structure of climate-related social-media posts and advertisements (Fig.~\ref{fig:pipeline}). This process is done independently per-platform as described below. 

We first embed texts into a shared semantic space using SentenceBERT (all-MiniLM-L6-v2) with L2 normalization, followed by dimensionality reduction via Principal Component Analysis (PCA) and Uniform Manifold Approximation and Projection (UMAP) \cite{mcinnes2018umap}. Density-based clustering (HDBSCAN) \cite{mcinnes2017hdbscan} is applied to group semantically similar texts. To characterize each cluster, we rank members by their probability scores and extract the top-k (k=5) texts from each cluster as representatives for downstream processing. Because automatically discovered clusters may be noisy, we filter them for semantic coherence with an LLM, prompting it to judge whether each cluster's representative texts reflect a single underlying theme (few-shot details in App.~\ref{app:coher}). Coherence filtering retains $69.1\%$ of the $97$ initial Meta clusters and $52.6\%$ of the $97$ Bluesky clusters.  

For each retained cluster we generate a concise LLM summary capturing its shared theme and intent, using the five texts with highest probability score as few-shot examples (App.~\ref{app:pt} Fig.~\ref{fig:summary_prompt}). We then merge clusters whose summaries exceed a tuned embedding-similarity threshold and assign each merged cluster a short ($1$--$3$ word), human-interpretable theme label (merging and labeling details in App.~\ref{app:merge}, App.~\ref{app:assign}).

Finally, texts are assigned to themes via two complementary strategies: (i) matching a text to its most similar cluster summary and inheriting that summary's theme, and (ii) directly assigning the text to one of the generated theme labels. This yields $\mathbf{47}$ themes for Meta (after de-duplication; App.~\ref{app:assign}) and $\mathbf{46}$ for Bluesky (Table~\ref{tab:thm}, App.~\ref{app:thm_all}).

All LLM steps (coherence filtering, summarization, theme labeling) use Mistral-Large-Instruct-2407 with few-shot prompting, chosen for its strong performance on structured annotation tasks among comparably sized open-weight models within our compute budget \cite{islam2025post}. Full prompts and implementation details are in App.~\ref{app:pt}, App.~\ref{app:experim}. We compare our two assignment variants against three topic-modeling baselines: LDA, BERTopic, and TopicGPT. Assignments are evaluated by both human annotators (App.~\ref{app:human}) and an LLM judge (Fig.~\ref{fig:llm_judge}, App.~\ref{app:pt}).
\begin{table}[t]
\centering
\small
\begin{tabular}{l|cc|cc}
\hline
 & \multicolumn{2}{c|}{\textbf{Meta}}  & \multicolumn{2}{c}{\textbf{Bluesky}} \\
\textbf{Method} & \textbf{Human} & \textbf{LLM} & \textbf{Human} & \textbf{LLM} \\
\hline
LDA & 0.25 & 0.43 & 0.35 & 0.44 \\
BERTopic & 0.25 & 0.37 & 0.26 & 0.25 \\
TopicGPT & 0.95 & 0.96 & 0.93 & 0.88 \\
LLM Summary & 0.45 & 0.60 & 0.62 & 0.46 \\
LLM Theme & 0.67 & 0.46 & 0.67 & 0.51 \\
\hline
\end{tabular}
\caption{Accuracy comparison across methods for Meta and Bluesky data under human and LLM evaluations.}
\vspace{-5pt}
\label{tab:combined-accuracy-comparison}
\vspace{-5pt}
\end{table}
\section{Results}
\subsection{Results: Meta} 
For each ad, we compare theme assignments produced by our two assignment strategies of Text-to-Theme and Text-to-Summary against keywords inferred by three baseline methods: LDA, BERTopic, and TopicGPT. We randomly sample $500$ Meta ads and assess the correctness of each method's assigned theme or keyword set. Each sample is independently evaluated by (1) human annotators (App. \ref{app:human}) and (2) an LLM-based judge (Qwen3-235b-a22b) \cite{yang2025qwen3technicalreport}. The prompt given to the LLM judge is shown in Fig. ~\ref{fig:llm_judge} App. \ref{app:pt}. 
Annotators determine whether the assigned theme or keyword set accurately reflects the central topic and framing of the ad (App. \ref{app:human}). Representative annotation examples are shown in Tables ~\ref{tab:results-table-llm} and ~\ref{tab:results-table-human} App. \ref{app:eval_meta}. We quantify the level of agreement between the two judges by computing Cohen's Kappa \cite{cohen1960coefficient} across both evaluation conditions. For summary-based judgments, we observe 79.23\% raw agreement ($\kappa = 0.58$) between the human and LLM annotator. For theme-based judgments, we see 77.42\% agreement ($\kappa = 0.50$). Both Cohen's Kappa scores fall in the moderate agreement range, which is consistent with what is typically reported for semantic-level annotation tasks of this nature. Results are shown in Table \ref{tab:combined-accuracy-comparison}.

TopicGPT achieves high surface-level accuracy ($96\%$ for the LLM judge) 
However, this result is largely driven by the generality of the topics it produces (e.g., “Environment,” “Energy,” “Climate Change”), which broadly match most posts in a climate-focused corpus. This is reflected in a low Cohen’s $\kappa$ of 0.19, indicating only slight agreement once chance overlap is accounted for. However, there is 93.19\% raw agreement for this dataset. The near-collapse of Kappa to Slight agreement despite high raw agreement score is attributable to the highly skewed label distribution: 94.99\% of human annotations and 96.19\% of LLM annotations are marked “True”. In contrast to TopicGPT, our methods produce more specific and semantically targeted themes (“Water Resilience,” “Climate Legal Accountability,” and “Capitalism’s Climate Doom”) while achieving higher accuracy than LDA and BERTopic across both evaluation settings. App. ~\ref{app:eval_meta} provides additional details.

\subsection{Results: Bluesky } 
We evaluate theme assignment on a randomly sampled set of $500$ Bluesky posts using both human and LLM-based judges. 
Across both evaluation settings (Table \ref{tab:combined-accuracy-comparison}), our methods outperform both LDA and BERTopic, with our Text-to-Theme variant achieving the highest accuracy among our methods. TopicGPT again achieves high surface-level accuracy ($88\%$ for LLM judge) due to its broader topic labels. App. ~\ref{app:eval_bsky} provides a detailed discussion of systematic differences between human and LLM-based judgments on the Bluesky dataset.

\section{Analysis}
Through our analysis, we explore how paid advertising on Meta and organic discourse on Bluesky reflect fundamentally different communicative orientations: strategic persuasion and systemic critique. We see these orientations manifest themselves consistently across thematic structure (\S\ref{sec:thematic}), rhetorical register (\S\ref{sec:rhetoric}), stance alignment (\S\ref{sec:stance-align}), and responsiveness to events (\S\ref{sec:temporal}). 

\subsection{Thematic Structure}
\label{sec:thematic}

\begin{figure}[h]
    \centering
    \includegraphics[width=1.0\linewidth]{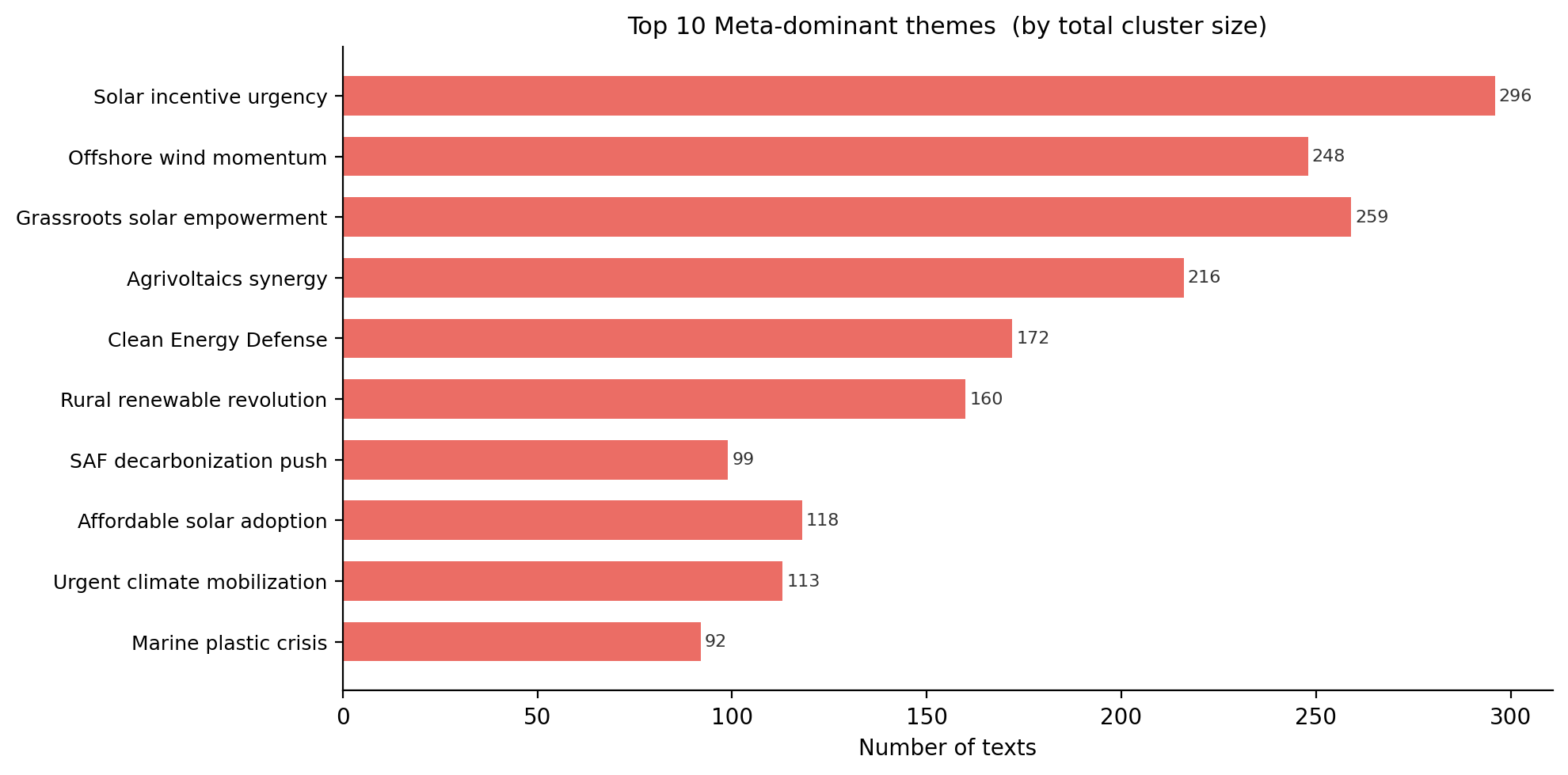}
    \caption{Top 10 themes (joint-clustering) which are Meta-Dominant, meaning that over 70\% of texts in each given theme cluster is a Meta theme.}
    \label{fig:top10-meta-joint}
\end{figure}

To enable direct cross-platform comparison, we re-run our pipeline on the joint combination of all Meta and Bluesky texts ($36,206$ total texts). Modified experimental details are provided in App. ~\ref{app:joint}. The resulting $354$ coherent themes split into three categories based on within-theme platform composition (70\% threshold): $145$ Meta-dominant, 
$112$ Bluesky-dominant, and $97$ shared themes.The top Meta-dominant themes by cluster size are \textit{Solar incentive urgency} ($296$ Meta texts), \textit{Offshore wind momentum} ($248$), and \textit{Grassroots solar empowerment} ($259$)---all renewable-energy promotion (Fig.~\ref{fig:top10-meta-joint}). Meta texts are distributed relatively evenly across the top themes, forming a diverse set of topics for campaign. 

\begin{figure}[h]
    \centering
    \includegraphics[width=1.0\linewidth]{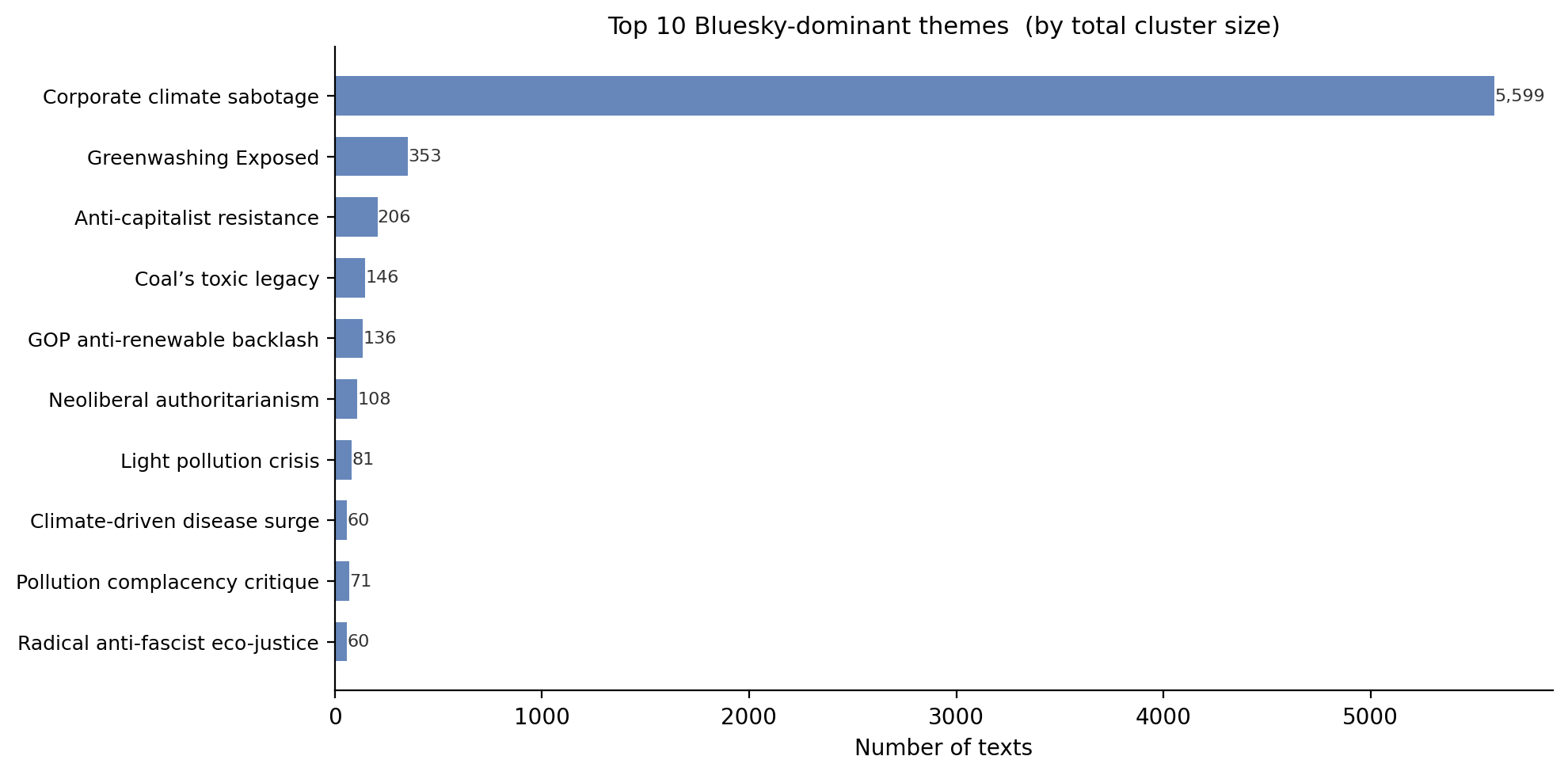}
    \caption{Top 10 themes (joint-clustering) which are Bluesky-Dominant, meaning that over 70\% of texts in each given theme cluster is a Bluesky theme.}
    \label{fig:top10-bluesky-joint}
\end{figure}

The Bluesky-dominant distribution differs greatly from the Meta-dominant distribution. A single theme, \textit{Corporate climate sabotage}, accounts for $5,599$ Bluesky texts; This number is larger than the entire top-four Meta-dominant cluster size combined (Fig.~\ref{fig:top10-bluesky-joint}). The remaining top Bluesky-dominant themes (\textit{Greenwashing exposed}, \textit{Anti-capitalist resistance}, etc.) are organized around systemic critique of corporate, political, and economic actors. The full theme-size distribution is found in App. ~\ref{app:shared-themes} Fig.~\ref{fig:joint-theme-dist}. Topically, we see that Meta-dominant clusters focus on a diverse set of promotional topics such as \textit{Affordable solar adoption}, whereas Bluesky topics are heavily concentrated on a few critique-oriented themes. We include a list of shared themes for Bluesky and Meta in App. ~\ref{app:shared-themes} (Fig. ~\ref{fig:shared-themes}), where each cluster in this category is neither predominantly comprised of Meta or Bluesky texts. A supplementary named-entity analysis (App. ~\ref{app:ner_results}) further reinforces this topical contrast: Meta-dominant clusters emphasize U.S. states and named individuals consistent with targeted promotional campaigns, whereas Bluesky-dominant clusters emphasize countries and collective actors consistent with larger-scale critique and structural framing. 

\subsection{Rhetorical Register}
\label{sec:rhetoric}

We next characterize \textit{how} the two platforms talk about climate using two different lenses: an Entman framing lexicon~\cite{entman1993framing} operationalizing the four framing functions across eight subcategories (crisis/solution, institutional/individual actor, economic/moral, scientific authority/populist skepticism), and the Linguistic Inquiry and Word Count (LIWC-22) framework~\cite{boyd2022liwc} summary variables. Lexicon construction, cluster- and demographic-level breakdowns, and theme-level LIWC profiles are provided in App.~\ref{app:register}. Together, these lenses show an asymmetric register. Paid communication leans formal, authoritative, and solution-oriented; Organic discourse leans personal, crisis-oriented, and scientifically grounded. 

\begin{figure}[t]
    \centering
    \begin{subfigure}{\linewidth}
        \includegraphics[width=\linewidth]{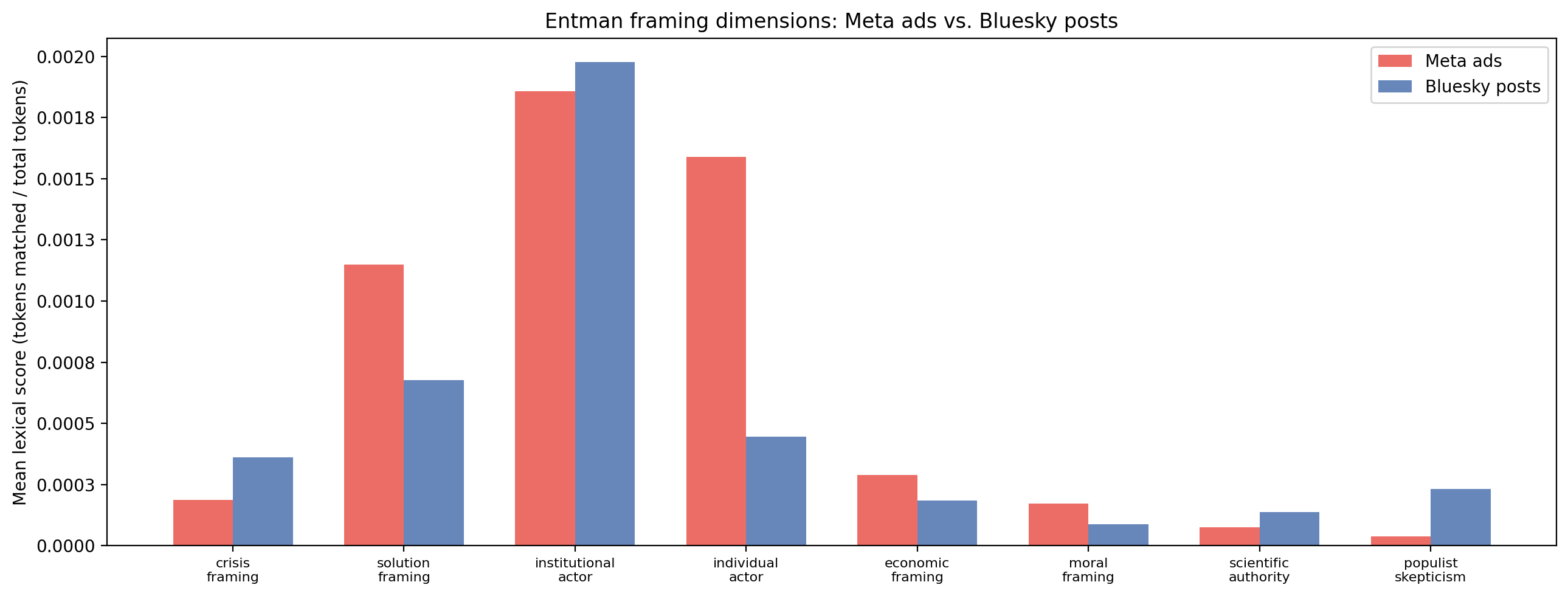}
        \caption{Mean Entman framing scores by platform.}
    \end{subfigure}
    \vspace{0.3em}
    \begin{subfigure}{\linewidth}
        \includegraphics[width=\linewidth]{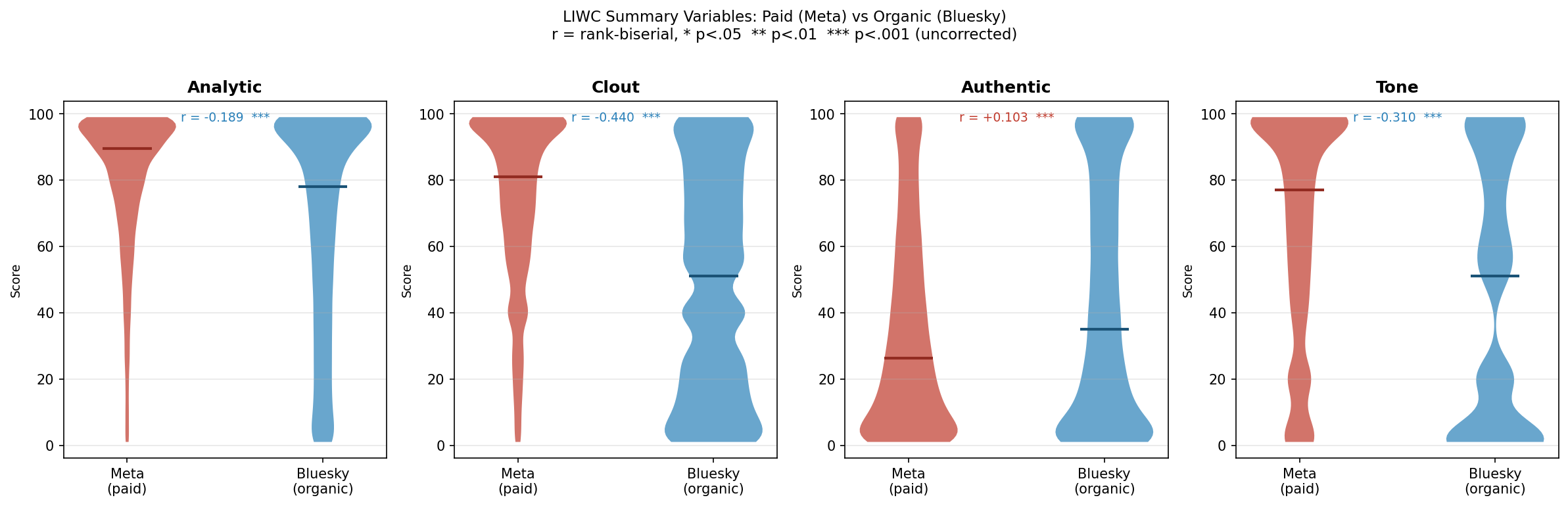}
        \caption{LIWC summary variables by platform.}
    \end{subfigure}
    \caption{Rhetorical register comparison: paid climate communication 
    is forward-looking, authoritative, and emotionally positive; organic 
    discourse is crisis-oriented, scientifically grounded, and affectively 
    mixed.}
    \label{fig:register-combined}
\end{figure}

\paragraph{Framing.} Fig. ~\ref{fig:register-combined} (a) presents mean lexical framing scores across the eight subcategories. Meta ads score substantially higher on solution and individual-actor framing, consistent with calls-to-action targeting the individual viewer. Bluesky posts score higher on crisis framing, scientific authority, and (apparent) populist skepticism---the last likely reflecting criticism \textit{of} skeptics rather than skepticism itself. Cluster-level analysis (App. ~\ref{app:register}) shows Bluesky-dominant clusters peak on emotional urgency (0.79)
 and institutional actor (0.71) framing, while Meta-dominant clusters score lowest on scientific framing (0.33) and highest on emotional hope (0.75) and future framing (0.70), suggesting paid climate advertising favors emotive and future-focused persuasion as opposed to evidence-heavy rhetoric. 
 
\paragraph{LIWC.} Fig. \ref{fig:register-combined} (b) shows Mann-Whitney 
U comparisons across four LIWC summary variables, with effect sizes via 
rank-biserial correlation ($r$). Meta ads score significantly higher on 
\textit{Analytic} ($r=-0.189$, $p<.001$) and \textit{Clout} ($r=-0.440$, 
$p<.001$), indicating a more formal, categorical, and authoritative register. 
The \textit{Clout} difference is the largest effect observed. By contrast, 
Bluesky posts score higher on \textit{Authentic} ($r=+0.103$, $p<.001$). 
Meta ads also exhibit more positive \textit{Tone} ($r=-0.310$, $p<.001$), 
with a median near $75$ compared to roughly $52$ for Bluesky.

\subsection{Stance Alignment}
\label{sec:stance-align}

If induced themes capture meaningful discourse structure, they should be stance-aligned. We manually annotate $500$ Meta ads and $500$ Bluesky posts as \textit{Pro-Climate}, \textit{Pro-Energy}, or \textit{Neutral}, with breakdowns found in App. ~\ref{app:qual_bsky}.  

\begin{figure}[t]
\centering
\begin{subfigure}{.48\columnwidth}
  \centering
  \includegraphics[width=\textwidth]{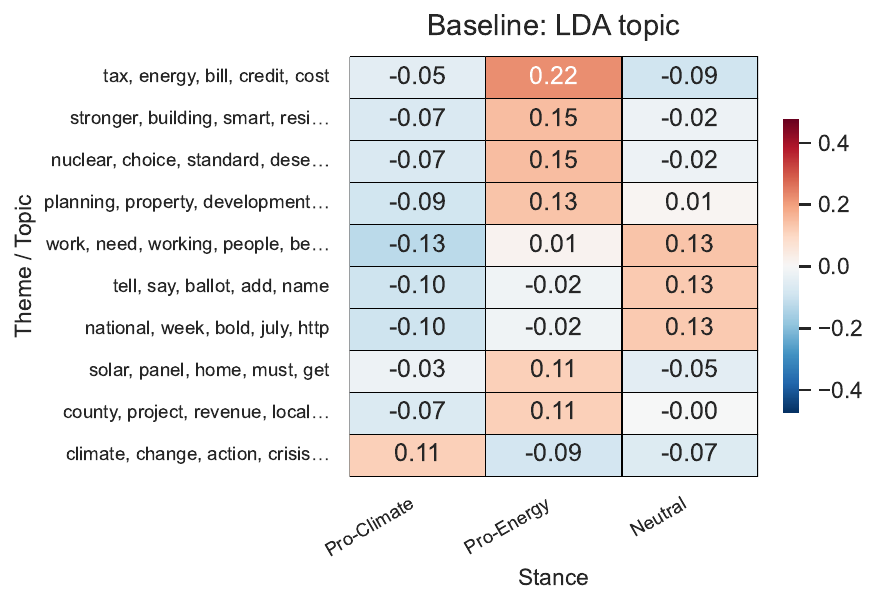}
  \caption{\textbf{Baseline}: LDA.}
\end{subfigure}%
\hfill
\begin{subfigure}{.48\columnwidth}
  \centering
  \includegraphics[width=\textwidth]{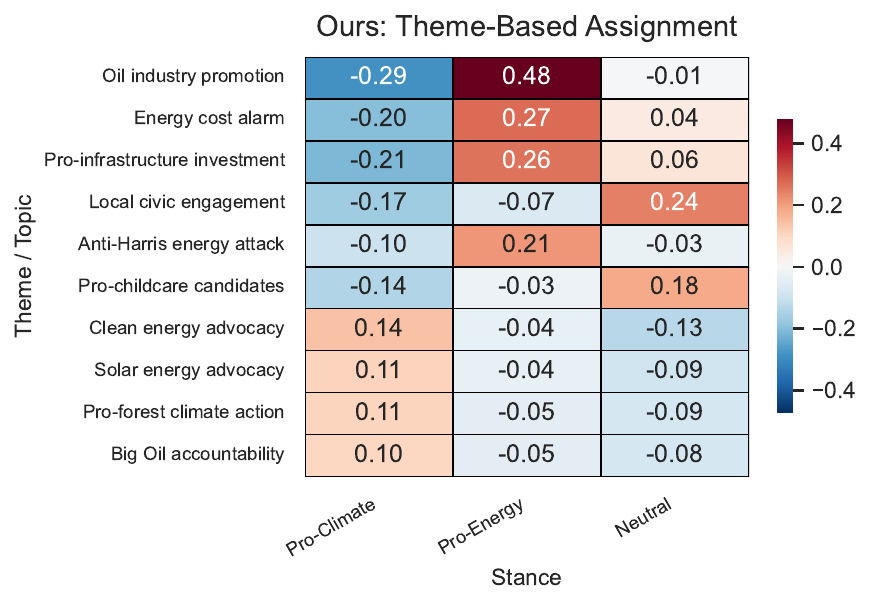}
  \caption{\textbf{Ours}: LLM Theme.}
\end{subfigure}
\caption{Theme--stance correlation on Meta. Our themes produce 
concentrated, interpretable stance alignment, while baselines yield 
diffuse correlations.}
\label{fig:correl_meta}
\end{figure}

Fig.~\ref{fig:correl_meta} presents theme--stance correlations on Meta for 
LDA and our Text-to-Theme variant; full panels for BERTopic, Text-to-Summary, 
and Bluesky are provided in App.~\ref{app:qual_bsky}. The LDA baseline 
produces diffuse, weak correlations across stances. Our themes produce 
markedly more concentrated correlations: Pro-Climate themes align with 
Pro-Climate stance, and Pro-Energy themes with Pro-Energy stance. Bluesky 
correlations are constrained by the platform's stance imbalance but remain 
directionally consistent, whereas LDA baseline correlations are more sensitive 
to noise. Overall, we find the themes encode stance-relevant structure, and the platforms differ not only what they discuss, but also in stance-balance of their discourse.

\subsection{Temporal Responsiveness}
\label{sec:temporal}

Finally, we examine how the two platforms respond to external climate-relevant events: the \textit{November 5, 2024 U.S. presidential election} and the \textit{January 2025 Southern California wildfires}. The two platforms respond to the same events in opposite directions: Paid climate advertising contracts post-election and shifts toward attribution-heavy, less future-oriented framing; Organic discourse expands. These divergent responses are consistent with the platform 
orientations established in \S\ref{sec:thematic}. We report text-volume shifts here; corresponding spend, impression, 
and detailed LIWC dynamics are provided in App.~\ref{app:meta-election} and 
App.~\ref{app:meta-wildfire}. 

\begin{figure}[t]
    \centering
    \includegraphics[width=1.0\linewidth]{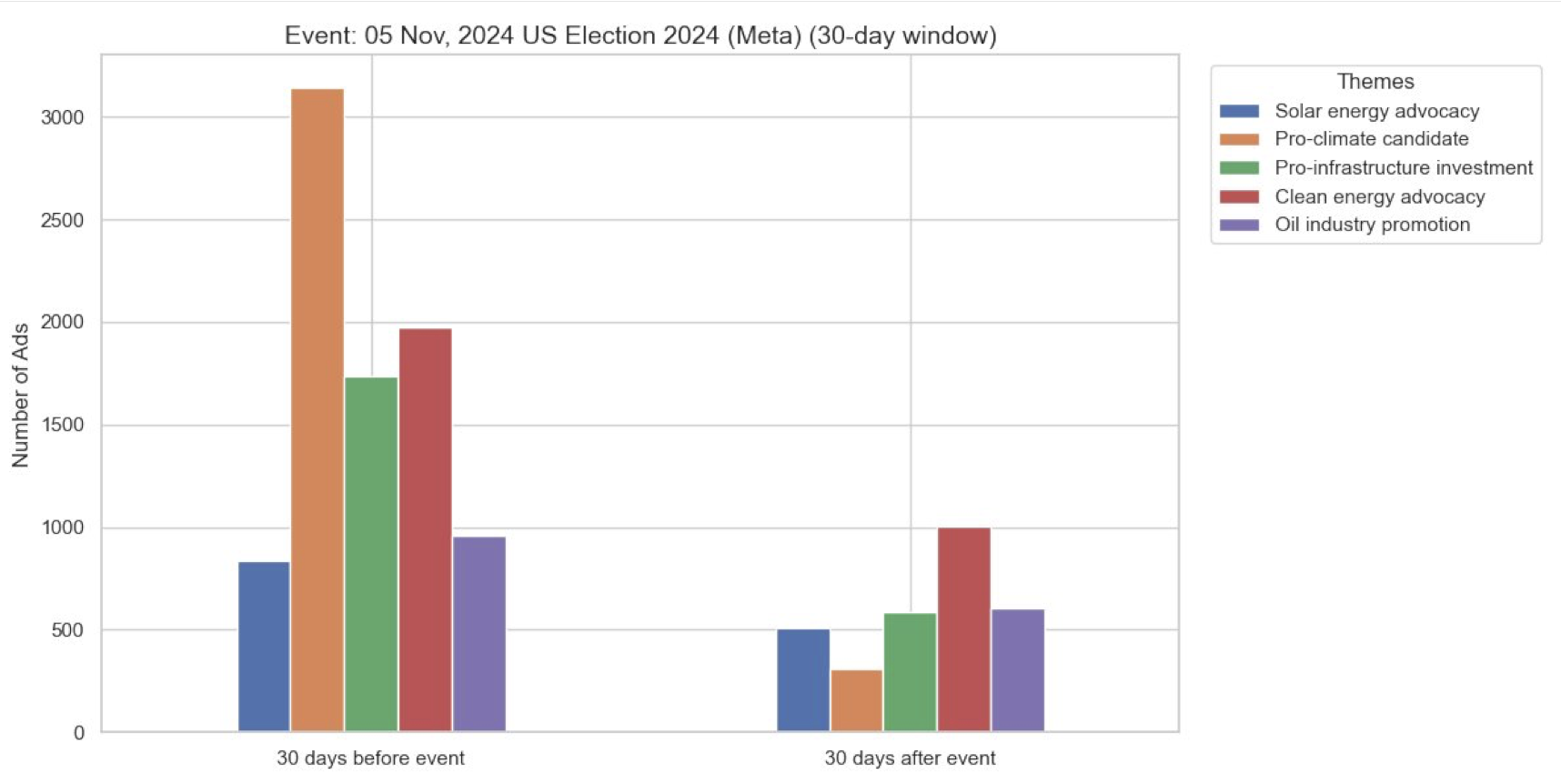}
    \caption{Top Meta themes 30 days before and after the November 2024 
    U.S. election. All five top-shift themes contract post-election.}
    \label{fig:election_themes_main}
\end{figure}

\paragraph{Meta contracts.} As shown in Fig.~\ref{fig:election_themes_main}, 
the 2024 U.S. election is followed by a sharp decline in climate-related 
advertising across the five themes with the largest pre/post shifts. These five themes span both Pro-Climate topics, such as clean-energy advocacy, and Pro-Energy topics, such as oil-industry promotion. This suggests that the post-election pullback is not necessarily stance-specific. Linguistically, we see that \textit{Cause} language increases 
significantly post-election, whereas \textit{moral}, \textit{risk}, and \textit{focusfuture} all decline (App. ~\ref{app:meta-election} Fig.~\ref{fig:liwc-election-meta}). The urgency and consequence-driven appeals used to promote candidates pre-election are dampened once the electoral cycle closes. Interestingly, a decreased \textit{power} and increased \textit{tentat} suggest uncertainty post-election. The 2025 wildfire window produces a different linguistic signature: \textit{power} language characteristic of heightened institutional action rises sharply, while \textit{Tone} becomes more negative (App.~\ref{app:meta-wildfire}).

\begin{figure}[t]
\centering
\begin{subfigure}{\linewidth}
    \includegraphics[width=\linewidth]{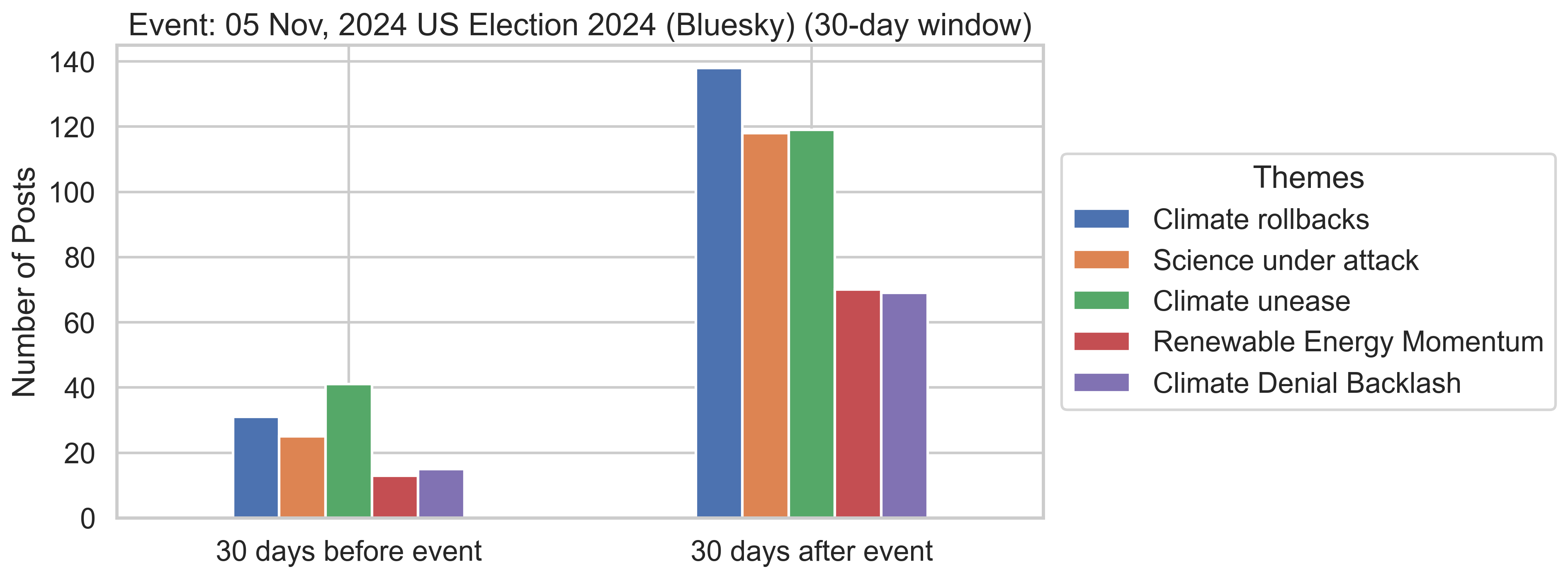}
    \caption{Top Bluesky themes $\pm 30$ days around U.S. election.}
\end{subfigure}
\vspace{0.3em}
\begin{subfigure}{\linewidth}
    \includegraphics[width=\linewidth]{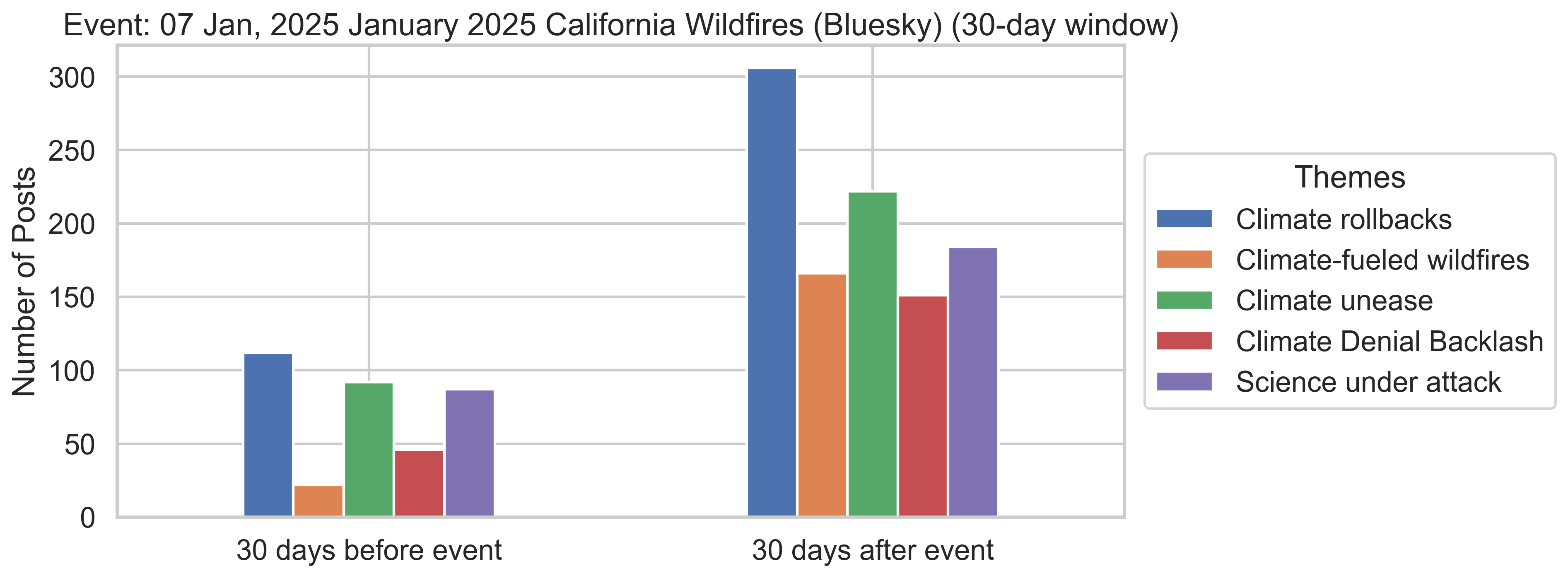}
    \caption{Top Bluesky themes $\pm 30$ days around CA wildfires.}
\end{subfigure}
\caption{Thematic shifts on Bluesky surrounding major events: rapid 
emergence of critique-oriented themes that were absent pre-event.}
\label{fig:bluesky_event_themes}
\end{figure}

\paragraph{Bluesky amplifies.} Fig.~\ref{fig:bluesky_event_themes} 
illustrates the inverse pattern. Within the $\pm 30$-day window around the 
2024 U.S. election, every one of the top-shift themes---\textit{Climate 
rollbacks}, \textit{Science under attack}, \textit{Climate unease}, 
\textit{Renewable Energy Momentum}, and \textit{Climate Denial 
Backlash}---rises sharply post-election. The same expansion pattern 
recurs around the 2025 California wildfires, with \textit{Climate 
rollbacks} and \textit{Science under attack} again among the steepest 
risers. Linguistically, the two Bluesky events diverge (App.~\ref{app:liwc-event-bluesky} Fig.~\ref{fig:liwc-bluesky-events}). Around the election, LIWC register variables (\textit{Analytic}, \textit{Clout}, \textit{Authentic}, \textit{Tone}) remain stable, and only \textit{cause} shifts significantly. Around the wildfires, \textit{Tone} declines and \textit{cause} increases, indicating that organic discourse around this acute disaster is both more negatively valenced and more oriented toward attribution and responsibility.

\section{Downstream Tasks}
We evaluate the utility of generated themes on two downstream tasks:

\noindent \textbf{(i) A classification task} where models predict stance (e.g., \textit{Pro-Climate}, \textit{Pro-Energy}, and \textit{Neutral}) based on content. We compare performance using (a) text-only, (b) theme-only, and (c) combined features.

\noindent \textbf{(ii) A retrieval task} where themes act as queries to retrieve semantically relevant posts using dense embedding similarity. Precision@k is used to assess how often retrieved posts reflect theme-relevant content, using automatic keyword overlap for evaluation.
\subsection{Stance Prediction}
Stance Prediction is designed as a three-way classification task. We use \textbf{stratified} sampling to construct train, validation, and test splits (80/20 train–test, then 80/20 train–validation), preserving stance label distributions. Across model families and platforms, incorporating themes provides complementary signals and yields stronger gains in more balanced settings (e.g., Meta), while remaining competitive under severe class imbalance (Bluesky), as shown in Table~\ref{tab:stance_climate}. Theme-only models perform competitively, indicating that themes capture stance-relevant signal beyond surface text. On Bluesky, where class imbalance is more severe, improvements are smaller but remain consistent for simpler models. Full experimental details are provided in the App. \ref{app:stance}.

\begin{table}[t]
\centering
\small
\begin{tabular}{lcccc}
\toprule
\textbf{\textsc{Model}} 
& \multicolumn{2}{c}{\textbf{\textsc{Meta}}} 
& \multicolumn{2}{c}{\textbf{\textsc{Bluesky}}} \\
\cmidrule(lr){2-3} \cmidrule(lr){4-5}
& \textbf{\textsc{Acc}} & \textbf{\textsc{F1}} 
& \textbf{\textsc{Acc}} & \textbf{\textsc{F1}} \\
\midrule
$LR\textsubscript{tf-idf}_{txt}$ & \textbf{0.78} & 0.53 & 0.89 & 0.46\\
$LR\textsubscript{tf-idf}_{thm}$ & 0.72 & 0.59 & 0.90 & 0.47 \\
$LR\textsubscript{tf-idf}_{txt+thm}$ & 0.75 & \textbf{0.61} & \textbf{0.91} & \textbf{0.52}\\
\midrule
$RoBERTa_{txt}$        & 0.73 & 0.63 & 0.88 & 0.82 \\
$RoBERTa_{thm}$         & 0.72 & 0.60 & 0.88 & 0.82 \\
$RoBERTa_{txt+thm}$ & \textbf{0.75} & \textbf{0.67} & 0.88 & 0.82 \\
\midrule
$Kimi_{txt}$           & 0.85 & 0.74 & 0.84 & \textbf{0.52} \\
$Kimi_{thm}$            & 0.80 & 0.63 & 0.83 & 0.41 \\
$Kimi_{txt+thm}$          & \textbf{0.86} & \textbf{0.75} & \textbf{0.86} & 0.51 \\
\midrule
$Llama3_{txt}$            & 0.76 & 0.58 & 0.66 & 0.41 \\
$Llama3_{thm}$             & 0.66 & 0.54 & 0.73 & 0.36\\
$Llama3_{txt+thm}$         & \textbf{0.78} & \textbf{0.63} & \textbf{0.81} & \textbf{0.42} \\
\bottomrule
\end{tabular}
\caption{Contribution of theme to stance classification on test data. txt: text, thm: theme.}
\vspace{-5pt}
\label{tab:stance_climate}
\vspace{-5pt}
\end{table}

\begin{table}[t]
\centering
\begin{tabular}{lcc}
\toprule
\textbf{Platform} & \textbf{P@1} & \textbf{P@5} \\
\midrule
Meta    & 0.783 & 0.691 \\
Bluesky     & 0.818 & 0.736 \\
\bottomrule
\end{tabular}
\caption{Precision@$k$ for retrieving texts.} 
\vspace{-5pt}
\label{tab:retrieval}
\vspace{-10pt}
\end{table}
\subsection{Theme-Guided Retrieval}
We assess the semantic alignment between LLM-derived themes and content (texts) via a retrieval-based evaluation. For each unique theme, we embed both the theme and all ads/posts using SBERT. To qualitatively assess the alignment between theme and text embeddings, we project their vector representations into 2D using t-SNE \cite{vanDerMaaten2008} (Fig. ~\ref{fig:tsne_theme_text} App. \ref{app:tsne}). Across both datasets, theme embeddings (\textcolor{blue}{$\times$}) cluster near semantically related texts (\textcolor{orange}{$\bullet$}), visually supporting their utility for semantic retrieval. Meta shows a denser, centralized theme-ad structure, suggesting more coordinated messaging; while Bluesky reveals broader dispersion, reflecting diverse, user-driven discourse (Fig. ~\ref{fig:tsne_theme_text} App. \ref{app:tsne}).

We then compute cosine similarity and retrieve the top-k most similar posts per theme. Relevance is evaluated by checking whether retrieved posts contain keywords extracted from the theme (simple frequency-based filtering over normalized tokens). Table \ref{tab:retrieval} reports Precision@1 and Precision@5 as aggregate metrics. 
These results indicate that theme embeddings effectively retrieve semantically relevant content, even when evaluated with simple unsupervised metrics. Retrieved examples illustrate stylistic contrasts, i.e., Meta ads are coordinated and campaign-driven (Table~\ref{tab:fb_retrieval_examples} App. \ref{app:tsne}), while Bluesky posts reflect diverse, user-driven discourse (Table~\ref{tab:bsky_retrieval_examples} App. \ref{app:tsne}).

We additionally perform small-scale human relevance assessment on a random sample of $20$ themes (top-5 retrieved posts per theme) for each platform. Using a rubric (2 = clearly relevant, 1 = partially relevant, 0 = not relevant), the resulting P@5 (binary $\geq 1$) is $\textbf{0.60}$ for Meta and $\textbf{0.73}$ for Bluesky, with mean relevance scores of $\textbf{0.68}$ and $\textbf{0.88}$, respectively. For Meta, the result is comparable but more conservative than the keyword-based estimate reported in Table \ref{tab:retrieval}. Notably, the semantic P@5 for Bluesky closely matches the keyword-based estimate reported in Table \ref{tab:retrieval}, indicating that theme-guided retrieval captures semantic alignment beyond keyword overlap.

\section{Conclusion}
We present a comparative analysis of climate discourse across paid advertising and public social media, examining Meta advertisements and Bluesky posts from July 2024 to September 2025 through an interpretable thematic discovery framework that labels semantic clusters with concise, human-interpretable themes. Under both human and LLM-based evaluation, the induced themes are more coherent and stance-consistent than standard topic-modeling baselines, and downstream stance prediction and retrieval confirm they capture meaningful semantic structure rather than surface lexical patterns.
Applying these themes, we find the two environments reflect fundamentally different communicative orientations. Paid advertising concentrates on the strategic promotion of specific solutions in a formal, future-focused register, while organic discourse concentrates on the systemic critique of corporate and political actors in a crisis-oriented, scientifically grounded one. This contrast propagates consistently across thematic structure and responsiveness to events---most strikingly, the two move in opposite directions around the 2024 U.S. election, where advertising contracts and public discourse expands. More broadly, this work demonstrates the value of interpretable theme modeling for comparative thematic analysis across heterogeneous communication environments. While our analysis focuses on climate discourse, the framework can support cross-platform studies of narratives in other domains.

\section{Limitations}
While our framework enables interpretable cross-platform analysis of climate discourse, several limitations remain. First, our study focuses on two platforms: Meta’s advertising ecosystem and Bluesky as representative cases of paid and organic communication environments. Although these platforms exhibit clear structural differences, our findings may not directly generalize to other advertising systems or social media platforms due to different user bases, moderation policies, or affordances. We acknowledge the existence of confounding factors, and that the two platforms differ across many dimensions beyond simply paid versus organic discussion. These dimensions include user demographics, moderation policies, platform age, and more. To account for these compounds, we include supplemental demographic-level analyses on the Meta dataset, as well as note confounds for which no comparable data is available---such as per-user demographics on Bluesky---limit the interpretability of cross-platform comparisons, and that observed differences should not be attributed to the paid versus organic distinction alone.

Second, our thematic modeling pipeline relies on pre-trained large language models for cluster coherence filtering, summarization, and theme labeling. While we employ open-source LLMs to promote reproducibility and reduce cost barriers, these models may encode biases present in their training data \cite{islam2026gets,blodgett2020language,brown2020language}, which could influence theme interpretation. We do not explicitly measure or mitigate such biases in this work. Besides, we only used pre-trained LLMs and did not consider fine-tuning due to the resource constraints. Relatedly, our coherence filtering combines LLM-based judgments with quantitative cluster cohesion (mean pairwise cosine similarity), but the underlying model choices and thresholds were calibrated on our climate corpus; because LLM-based coherence judgments are model- and prompt-dependent, the procedure may require re-tuning to generalize to other domains, platforms, or datasets.

Lastly, \textbf{our analysis is descriptive rather than causal}. Although we identify systematic differences in thematic structure and temporal dynamics across platforms, we do not attribute intent to individual actors or infer causal effects of platform incentives on discourse. Future work could combine our framework with causal or experimental designs to further investigate these mechanisms.

%


\section{Ethics Statement}
To the best of our knowledge, we did not violate any ethical code while conducting the research work described in this paper. We report the technical details for the reproducibility of the results, as well as release the datasets and codes publicly. The author's personal views are not represented in any results we report, as it is solely outcomes derived from machine learning or AI models.

The social media data used in this study may contain offensive, biased, toxic, or harmful language. Such content reflects user-generated discourse and does not represent the views of the authors or institutions. All analyses were conducted for research purposes only.

We used LIWC-22 under a research license. Our use is consistent with its intended academic research purpose. Any derivatives or outputs involving LIWC features should not be used outside of research contexts.
\bibliography{custom}

\appendix


\section{Curated Keywords}
\label{app:kw}
The curated set of climate-related keywords shown in Table \ref{tab:climate-keywords} was collected from \cite{islam2023analysis} and modified to reflect recent shifts in climate-change related topics. 
\begin{table}[h]
    \centering
    \textbf{Climate-Related Keywords}
    \begin{tabular}{|l|}
        \hline
        climate change, climate denial, sustainable, \\
        fossil fuel, green new deal, oil and gas industry, \\
        solar energy, renewable energy, greenhouse gas, \\
        pollution, clean energy, green energy, \\
        global warming, deforestation, carbon emission, \\ drilling, fracking, coal mining,
        junk science, \\ climate misinformation, greenwashing \\
        \hline
    \end{tabular}
    \caption{Climate change related keywords for Meta and Bluesky data collection.}
    \label{tab:climate-keywords}
\end{table}

To evaluate resulting relevance of the datasets obtained through keyword extraction, we conduct human validation on a random sample of 500 Meta ads and 500 Bluesky posts. Among the Meta sample, 23 items (4.6\%) are judged to be irrelevant to climate-related content. Among the Bluesky sample, 31 items (6.2\%) are judged irrelevant. As such, 94.6\% of sampled items across both platforms are confirmed to be substantively climate-related. Note that for Meta, the estimated relevance rate of 95.4\% yields a 95\% CI of (93.6\%, 97.2\%). For Bluesky, the estimated relevance rate of 93.8\% yields a 95\% CI of (91.7\%, 95.9\%). These intervals are tight around the observed proportions, and so the high climate-relevance rates observed in our samples are reliable estimates of the relevance rates in the full datasets. 

\section {Experimental Details}
\label{app:experim}
In this section, we now describe the experimental configuration used to instantiate and evaluate the proposed framework.
We compute sentence-level embeddings using SentenceBERT (all-MiniLM-L6-v2) and then applying L2-normalization on each row. We then reduce dimensionality of the original embeddings by first applying Principle Component Analysis (PCA) and retaining the top 100 components, and then applying Uniform Manifold Approximation and Projection (UMAP) to project the data to $20$ dimensions. Through this two-step dimensionality reduction, we suppress noise and outlier sensitivity, allowing for a more easily separable representation to be used for clustering. 
We perform density-based clustering using HDBSCAN using a minimum cluster size of $20$ and a minimum number of samples of five. Clusters are characterized using membership probabilities, and the \textit{top-5} texts per cluster are selected as representatives.

\subsection{Cluster Coherency and Summarization}
\label{app:coher}
Cluster coherency checking and cluster summarization are performed using Mistral-Large-Instruct-2407. We provide few-shot examples for coherence evaluation. We look at the texts with the five highest probabilities per cluster and use them to determine whether the text is coherent or incoherent. A few-shot example given to LLM is shown in Table \ref{tab:coherency-ex}. We provide the LLM with three coherent examples and three incoherent examples. Then LLM labels cluster coherence based on the prompt shown in Fig. \ref{fig:coherency_prompt}. Upon removing incoherent examples, we prompt the LLM to generate a summary from the top five texts from each coherent cluster (Fig. \ref{fig:summary_prompt}).
\subsection{Cluster Merging}
\label{app:merge}
We use cosine similarity with SBERT (all-mpnet-base-v2) to determine which clusters are semantically similar and should be merged. We perform a grid search with threshold ($\tau \in \{0.60, 0.65, 0.70, 0.75, 0.80, 0.85, 0.90 \}$) and use Silhouette Score ($S$) and Davies-Bouldin Index (DBI) as guideline metrics for determining best threshold value upon merging. For the Meta data, we find that a threshold value of $0.8$ (S=0.161, DBI=0.665) obtains the best merging results. We began with $67$ total clusters and merged $16$ clusters for a total of $51$ clusters. This clustering resulted in 3 duplicate themes, which are also removed (see \ref{app:assign}). 
For the Bluesky data, we also find that a threshold value of $0.8$ ($S$=0.484, DBI=0.444) obtains the best merging results. We began with $51$ total clusters and merged $5$ of them, for a final total of $46$ clusters. 

\subsection{Theme Labeling and Assignment}
\label{app:assign}
Theme labels are generated from cluster summaries using few-shot prompting with Mistral-Large-2407. The prompt used to generate each theme is the following: {\small \texttt{``Produce a short theme label in 1–3 words for the given summary. This theme label should capture the central idea, stance, and/or topic of the text. Examples: `Against climate policy'......"}} shown in Fig. \ref{fig:theme_prompt}. As a final pre-processing step, clusters with duplicate themes are removed. This yields no removals for Bluesky and eliminates $3$ clusters from Meta. We have a total of $\mathbf{47}$ themes for Meta and $\mathbf{46}$ themes for Bluesky. Our final themes for Meta and Bluesky datasets are shown in Table \ref{tab:thm}. 
Once we obtain summaries and themes for each coherent cluster, we assign each text to its best-fitting theme through the following two methods:

\textbf{\textit{(i) Text to Summary}} where we prompt LLM with the full set of cluster-level summaries. The model selects the single summary which best-fits the text. Because each summary was previously mapped to a theme during the Theme Label Generation step, the ad/post automatically inherits the corresponding theme once its summary is chosen. For efficiency, assignments are processed in batches of ten texts per LLM call. The exact prompt used for this procedure is shown in Fig. ~\ref{fig:summary_assignment}.

\textbf{\textit{(ii) Text to Theme}} where we assign each ad to a theme directly by providing Mistral-Large-2407 the full set of cluster themes. This step is analogous to the summary assignment but uses cluster themes rather than cluster summaries. Again, a batch size of ten ads per LLM call is used. The prompt given to the LLM for this step is shown in Fig. ~\ref{fig:assign_theme_llm}.

We assign by semantic fit rather than original cluster membership, since geometric proximity in embedding space does not always match the most appropriate theme, particularly for boundary-spanning texts.

\subsection{Prompting}
\label{app:pt}
\begin{figure}[h]
    \centering
    \includegraphics[width=1.0\linewidth]{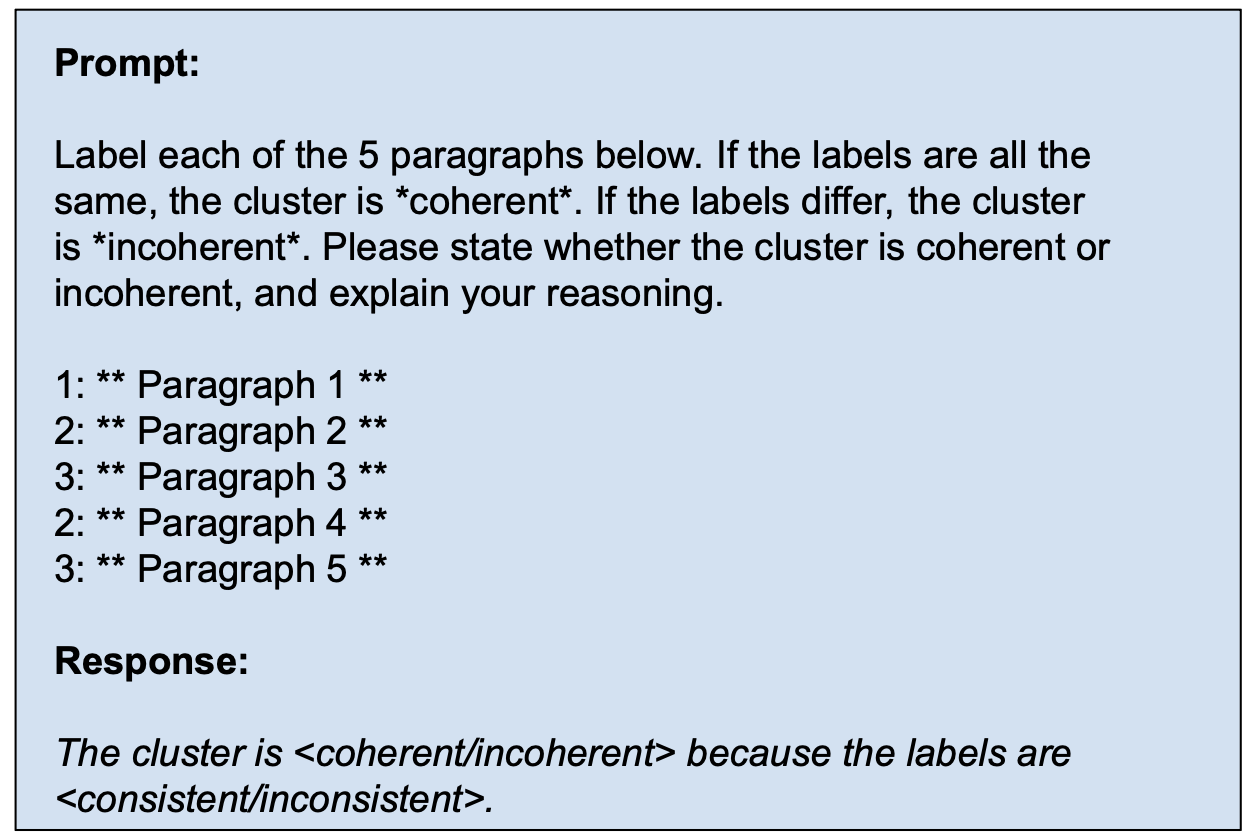}
    \caption{Coherency prompt with example response from LLM. }
    \label{fig:coherency_prompt}
\end{figure}

Table \ref{tab:coherency-ex} shows the few-shot example given to LLM as a prompt for the coherency check. We provide the LLM with three coherent examples and three incoherent examples. LLM then labels cluster coherence based on the prompt shown in Fig. \ref{fig:coherency_prompt}.

\begin{figure}[h]
            \centering
            \includegraphics[width=1.0\linewidth]{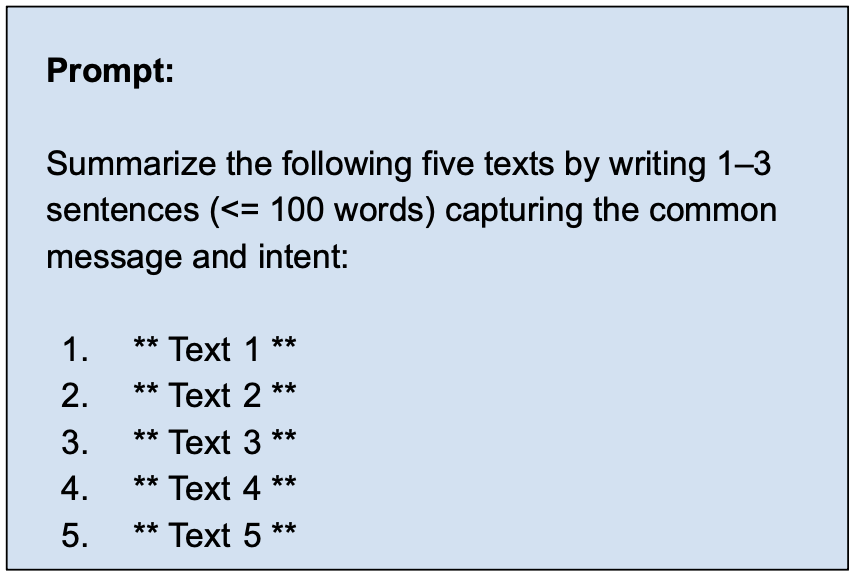}
            \caption{Prompt used for generating cluster summaries. }
            \label{fig:summary_prompt}
        \end{figure}

\begin{figure}[h]
    \centering
    \includegraphics[width=1.0\linewidth]{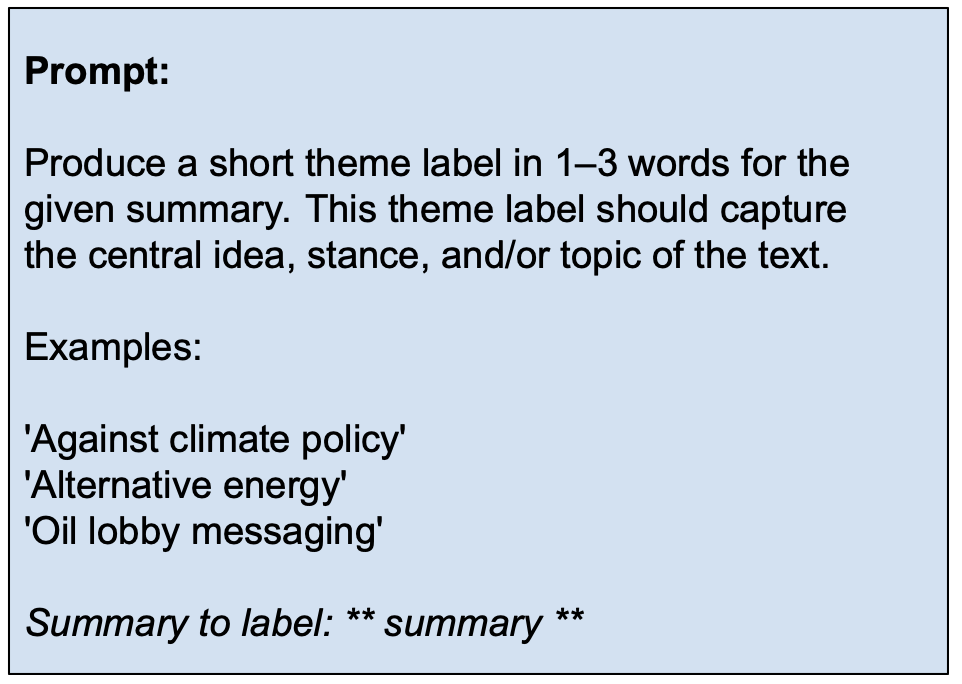}
    \caption{Prompt used for generating cluster themes from cluster summaries.}
    \label{fig:theme_prompt}
\end{figure}

\begin{figure}[h]
    \centering
    \includegraphics[width=1.0\linewidth]{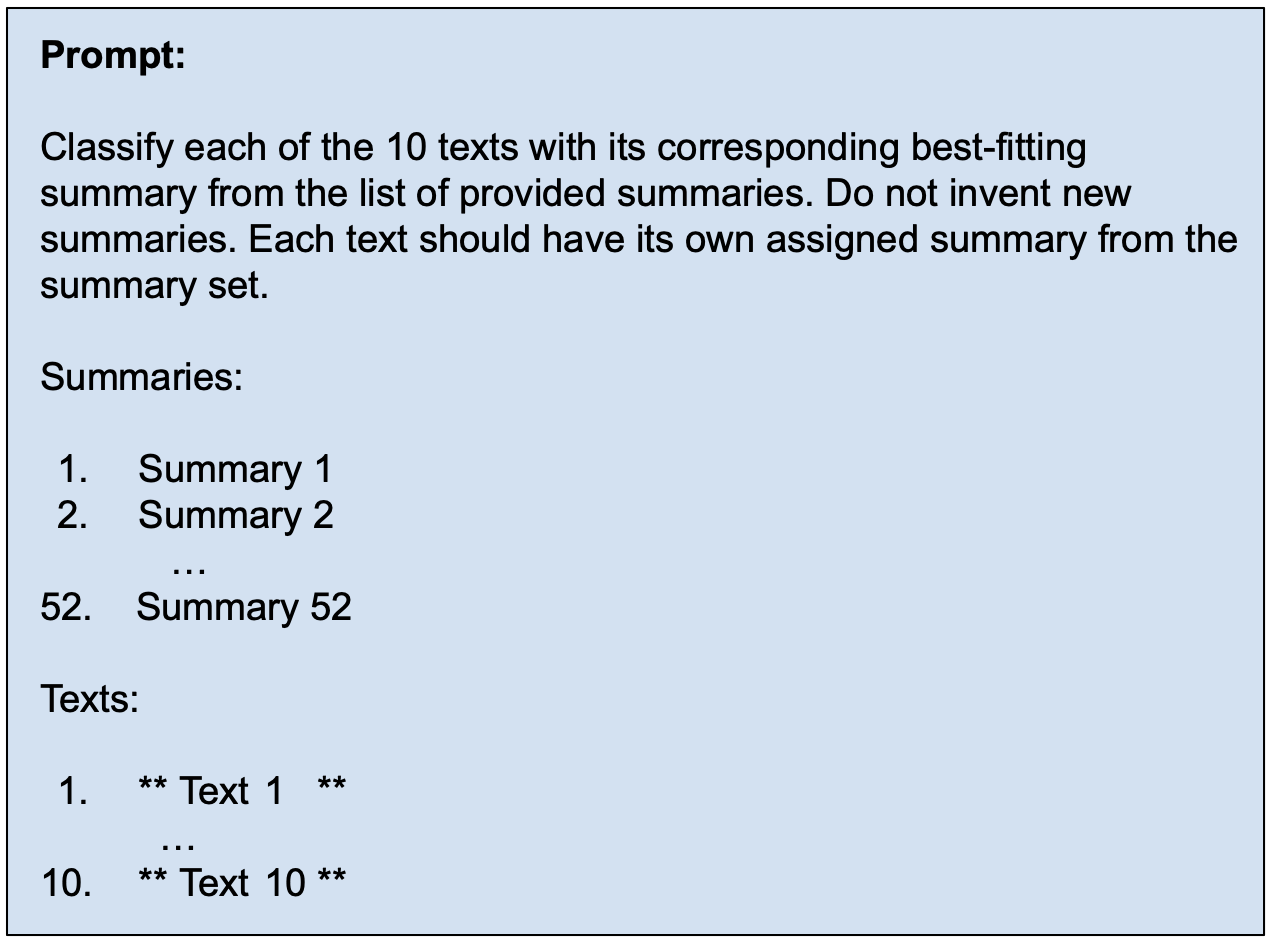}
    \caption{Prompt used for assigning text to corresponding summaries. }
    \label{fig:summary_assignment}
\end{figure}

\begin{figure} [h]
            \centering
            \includegraphics[width=1.0\linewidth]{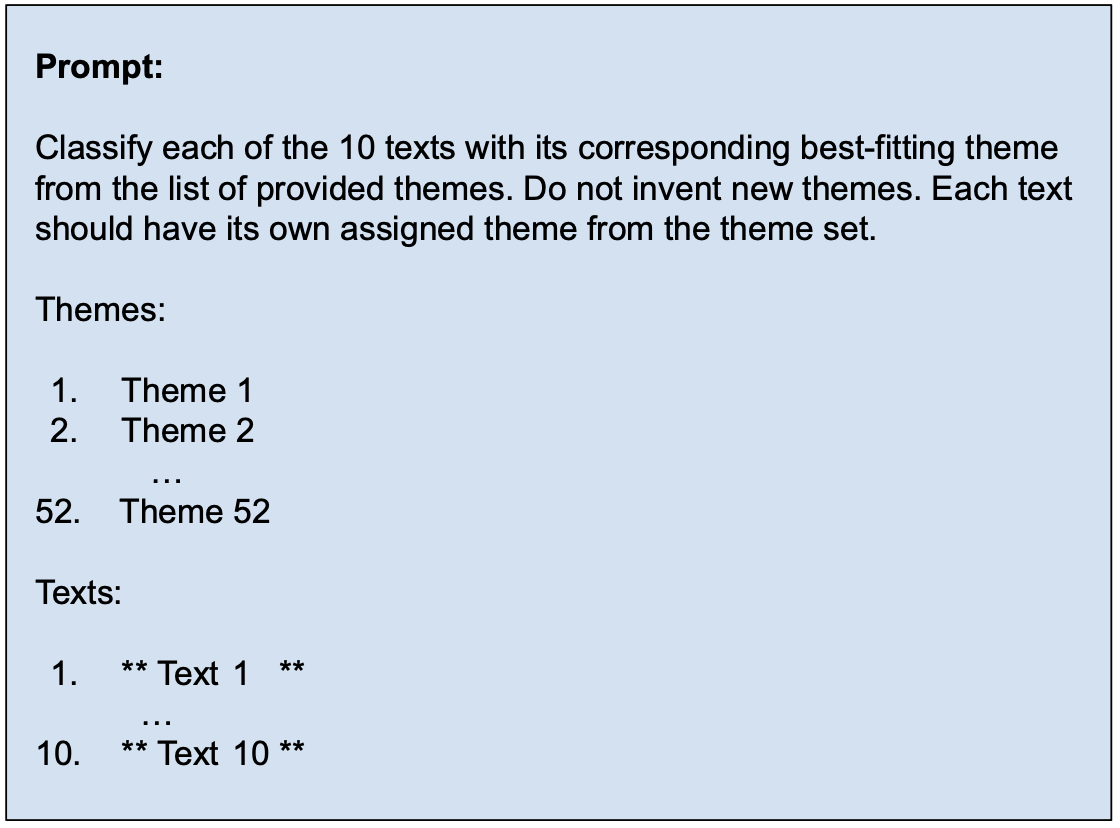}
            \caption{Prompt to assign each text to its corresponding theme. }
            \label{fig:assign_theme_llm}
        \end{figure}

\begin{figure}[h]
    \centering
    \includegraphics[width=1.0\linewidth]{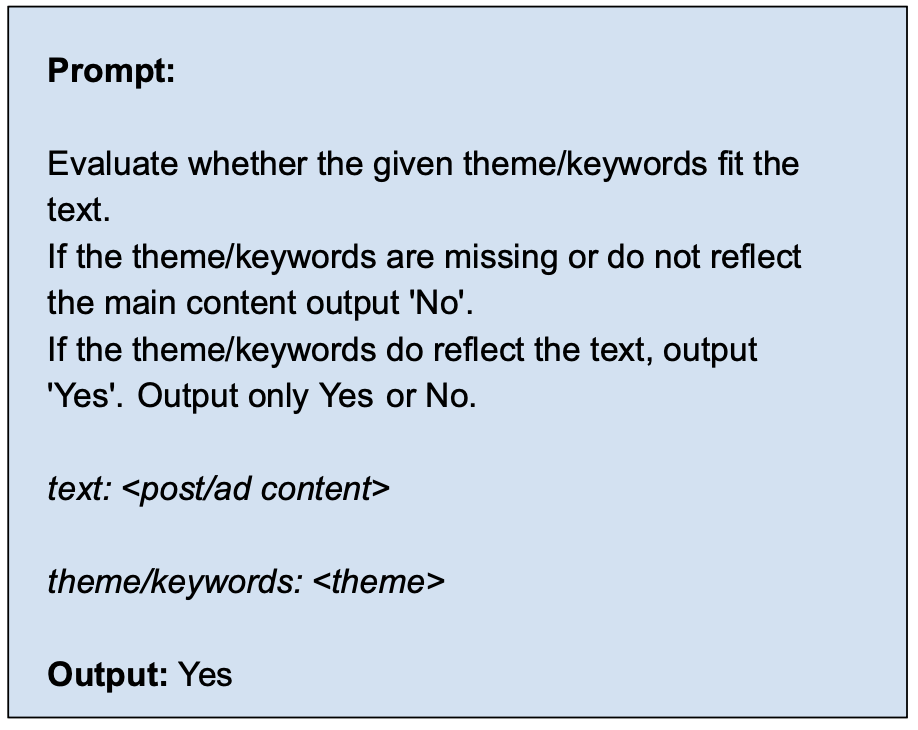}
    \caption{Prompt for LLM-based judge to annotate results based on the correctness of assigned keywords/theme.}
    \label{fig:llm_judge}
\end{figure}

Upon removing incoherent examples, we prompt Mistral-Large-Instruct-2407 with the prompt shown in Fig. \ref{fig:summary_prompt}. Each one-sentence summary is generated from the bodies of the top five texts from each coherent cluster. The prompt for generating themes is shown in Fig. \ref{fig:theme_prompt}, and the prompt used for assigning ads to corresponding summaries is shown in Fig. ~\ref{fig:summary_assignment}. Fig. ~\ref{fig:assign_theme_llm} shows the prompt to assign each text to its corresponding theme. Lastly, Fig. ~\ref{fig:llm_judge} shows the prompt to evaluate results using an LLM-based judge. 
\begin{table*}[t]
\centering
\begin{tabular}{p{0.04\textwidth} p{0.6\textwidth} p{0.25\textwidth}}
\hline
\textbf{\#} & \textbf{Paragraph} & \textbf{Label} \\
\hline
1 & Copper has become the world’s third most consumed metal, but uncertain demand levels and supply constraints have created a volatile outlook. & Against copper mining \\
2 & Highlights widespread support for copper mining in Arizona, emphasizing its role in renewable energy, national security, and supply chain resilience. & Supports copper mining \\
3 & Describes the Kennecott Mine’s long history, production levels, and contribution to sustainable technologies like wind turbines and electric vehicles. & Supports copper mining \\
4 & Discusses Alaska’s Estelle Gold Project and its potential for gold and antimony production. & Promotes gold mining \\
5 & Promotes mining in Minnesota as a modern STEM career, focusing on innovation in iron mining. & Supports iron mining \\
\hline
\multicolumn{3}{p{0.92\textwidth}}{\textbf{Cluster Coherency:} Incoherent} \\
\multicolumn{3}{p{0.92\textwidth}}{\textbf{Reasoning:} The cluster is incoherent because the labels are inconsistent.} \\
\hline
\end{tabular}
\caption{Few-shot example for coherency prompting.}
\label{tab:coherency-ex}
\end{table*}

\subsection{Discovered Themes}
\label{app:thm_all}
\begin{figure*}
    \centering
    \includegraphics[width=1.0\linewidth]{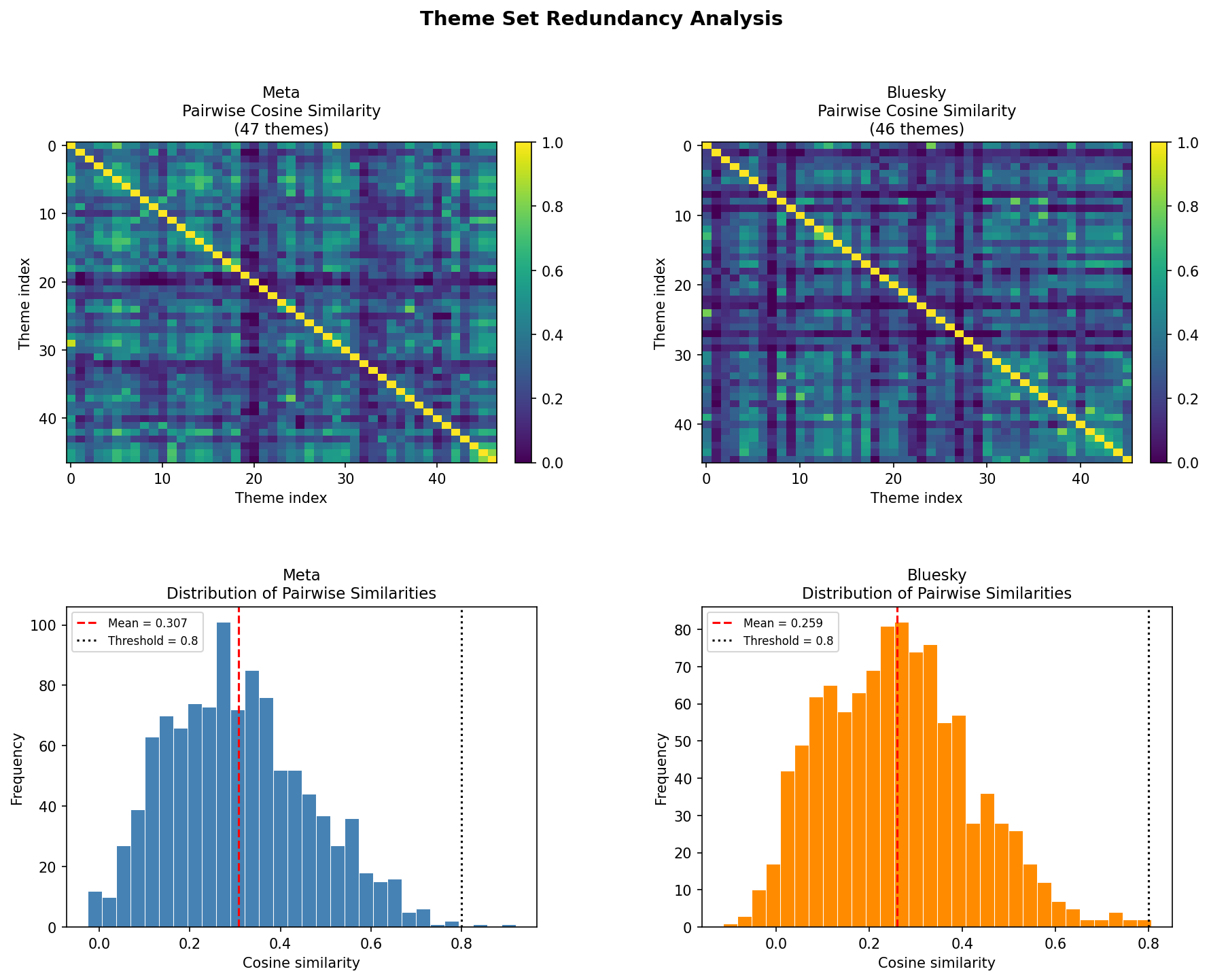}
    \caption{Correlation between and distribution of pairwise similarities for themes of the Meta and Bluesky datasets.}
    \label{fig:theme-similarity}
\end{figure*}

Table \ref{tab:thm} shows the resulting themes using our method on Meta and Bluesky datasets. 
To analyze redundancy of the derived theme sets, we compute pairwise cosine similarities between theme labels using SentenceBERT (sentence-transformers/all-mpnet-base-v2) and we find mean cosine similarity of 0.31 for the Meta theme set and 0.26 for the Bluesky theme set. From this evaluation, we find that the vast majority of theme pairs are semantically distinct. In total, only 3 theme pairs out of 2116 total possible pairs combined across the two theme sets are flagged for high similarity above a threshold value of 0.80, which is a redundancy rate of 0.14\%.  The distribution of pairwise similarities and correlation matrices for both datasets can be found in Fig. ~\ref{fig:theme-similarity}.

\begin{figure*}
    \centering
    \includegraphics[width=1.0\textwidth]{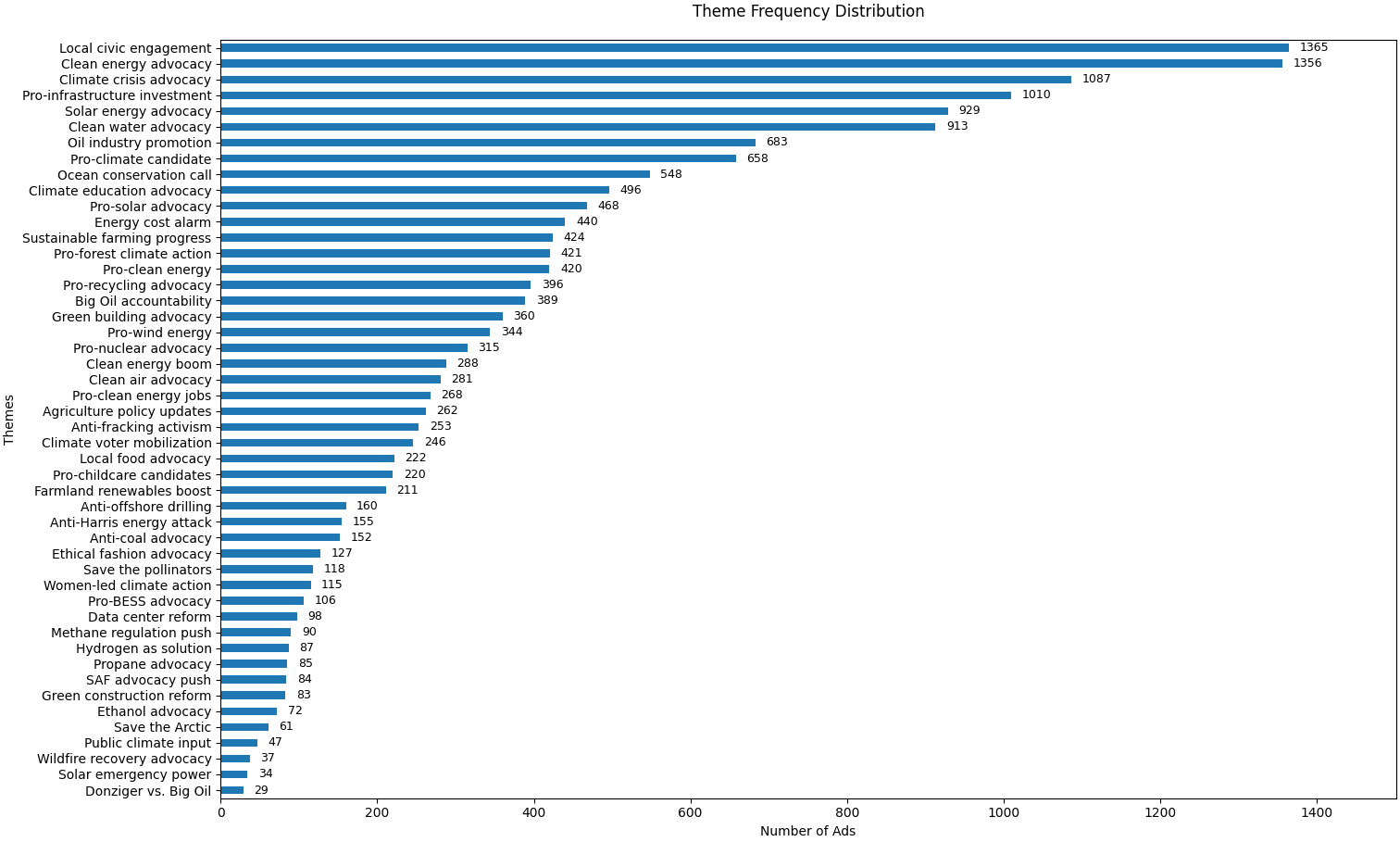}
    \caption{Distribution of themes across Meta dataset.}
    \label{fig:metathemedist}
\end{figure*}

In Fig. ~\ref{fig:metathemedist} and Fig. ~\ref{fig:bluesky-themes}, the distribution of texts into each theme from the full theme sets is shown for Meta and Bluesky, respectively. For Meta, the theme that receives the least number of assignments is “Donziger vs. Big Oil” with 29 assignments, whereas the theme with the greatest number of assignments is “Local civic engagement: with 1365 assignments. We see a mean of 354.4 assignments per theme. 23 themes are needed to cover 80\% of the Meta dataset, whereas 9 themes are needed to cover 50\% of the dataset. Respectively, we see a minimum number of assignments of 78 for the Bluesky dataset with the theme “Marine Climate Crisis” and a maximum of 2051 with “Climate rollbacks”. For this theme set, the average number of texts per theme is 416.9,  where 23 themes are needed to cover 80\% of the Bluesky dataset and 10 themes are needed to cover 50\%.

\begin{table*}
\begin{center}
 \scalebox{.9}{\begin{tabular}{>{\arraybackslash}m{1.9cm}|>{\arraybackslash}m{13.5 cm}}
\toprule
       \textbf{\textsc{Platforms}}  &  \textbf{\textsc{Themes}}\\
 \midrule
 \multirow{3}{*}{Meta} &  Agriculture policy updates, Anti-coal advocacy, Anti-fracking activism, Anti-Harris energy attack, Anti-offshore drilling, Big Oil accountability, Clean air advocacy, Clean energy advocacy, Clean energy boom, Clean water advocacy, Climate crisis advocacy, Climate education advocacy, Climate voter mobilization, Data center reform, Donziger vs. Big Oil, Energy cost alarm, Ethanol advocacy, Ethical fashion advocacy, Farmland renewables boost, Green building advocacy, Green construction reform, Hydrogen as solution, Local civic engagement, Local food advocacy, Methane regulation push, Ocean conservation call, Oil industry promotion, Pro-BESS advocacy, Pro-childcare candidates, Pro-clean energy, Pro-climate candidate, Pro-forest climate action, Pro-infrastructure investment, Pro-nuclear advocacy, Pro-recycling advocacy, Pro-solar advocacy, Pro-wind energy, Propane advocacy, Public climate input, SAF advocacy push, Save the Arctic, Save the pollinators, Solar emergency power, Solar energy advocacy, Sustainable farming progress, Wildfire recovery advocacy, Women-led climate action. \\
 \midrule
   \multirow{2}{*}{Bluesky}  & Anti-renewable hypocrisy, Aviation decarbonization, Capitalism's climate doom, China's green leap, Clean Energy Push, Cleaning fatigue, Climate Denial Backlash, Climate action collaboration, Climate injustice exposed, Climate justice action, Climate legal accountability, Climate migration crisis, Climate rollbacks, Climate unease, Climate-fueled wildfires, Coffee climate crisis, Corporate Climate Accountability, Deadly Heat Crisis, Deforestation \& Wildlife Crisis, EV transition barriers, End fossil fuels, Fascism’s fatal flaws, Fossil Fuel Overreach, Fossil fuel corruption, Fossil fuel deception, Green Energy Sovereignty, Greenwashing deception, Independent investigative journalism, Insurance Climate Urgency, MAGA Climate Deception, Marine climate crisis, Oil's Deceptive Greed, Plastic treaty failure, Renewable Energy Momentum, Renewable energy surge, Resource expansion drilling, Responsible tourism future, Science under attack, Sound's dual impact, Starry night preservation, Sustainable luxury fashion, Trees for Climate, Water pollution crisis, Water resilience, Wealth vs. Welfare, Wildlife climate crisis \\
 \bottomrule
\end{tabular}}
\caption{Resulting themes on Meta and Bluesky datasets.}
\label{tab:thm}
\end{center}
\end{table*}
\section{Evaluations}
\subsection{Evaluation on Meta Dataset} 
\label{app:eval_meta}
Table \ref{tab:results-table-human} shows representative annotation examples.
Table~\ref{tab:combined-accuracy-comparison} reports quantitative results across both evaluation settings (human and LLM judges). Under human evaluation, Text-to-Theme achieves the highest accuracy (0.673) of our methods, substantially outperforming Text-to-Summary (0.452), LDA (0.25), and BERTopic (0.25). While LDA attains comparatively higher accuracy under LLM-based evaluation, its performance drops under human judgment. 
Because LDA, BERTopic, and TopicGPT each assign multiple keywords or topics per document, LLM-based evaluation scores may vary depending on how the judge interprets partial correctness — for instance, whether a label set is marked correct only when all assigned topics are relevant, or when at least one is. This potential difference in interpretability should be considered when examining the results. BERTopic performs poorly under both evaluation settings, likely due to its high proportion of outlier assignments. 
TopicGPT performs best overall; however, much of this
accuracy might be attributed to the generic themes derived from the TopicGPT assignments, as shown in Tables \ref{tab:meta-level1-topics}, \ref{tab:meta-level2-topics}.

In contrast to the human evaluation results, the LLM judge assigns the highest accuracy to Text-to-Summary (0.598), followed by Text-to-Theme (0.456), despite both methods ultimately assigning labels from the same fixed theme set. Consequently, the observed difference in LLM-based accuracy does not stem from lexical differences in the presented labels, but rather from differences in the upstream assignment process.

Specifically, Text-to-Summary relies on an intermediate and implicit summary-to-theme mapping that is influenced by large context windows during summary assignment. This design can degrade assignment accuracy, as shown by the LLM and human evaluation results for Bluesky, as well as for the human annotation results for Meta. However, the Text-to-Summary method may also introduce patterns that align more closely with the LLM judge’s internal decision heuristics, leading to higher apparent accuracy under LLM evaluation. As such, the LLM judge may be sensitive to systematic properties of the assignment pipeline rather than to true semantic alignment between the text and the assigned theme. For this reason, we treat human annotation as the primary and more reliable evaluation signal.

\begin{table*}[t]
\centering
\scriptsize
\setlength{\tabcolsep}{1.5pt}   
\renewcommand{\arraystretch}{1.1}
\begin{tabular}{|p{0.16\textwidth}|p{0.14\textwidth}|p{0.14\textwidth}|p{0.07\textwidth}|p{0.07\textwidth}|p{0.08\textwidth}|p{0.08\textwidth}|p{0.07\textwidth}|p{0.07\textwidth}|}
\hline
\textbf{ad\_text} &
\textbf{lda\_keywords} &
\textbf{bert\_keywords} &
\multicolumn{2}{c|}{\textbf{LLM-Assigned Themes}} &
\multicolumn{4}{c|}{\textbf{LLM Judge Assignments}} \\
\cline{4-9}
 &  &  &
\textbf{Theme-Based} &
\textbf{Summary-Based} &
\textbf{lda\_correct} &
\textbf{bert\_correct} &
\textbf{Theme Correct} &
\textbf{Summary Correct} \\
\hline

Data can’t tell the whole story, but it’s a powerful tool for understanding changes on our planet. Grab this article from our free collection of resources to teach your students essential data literacy skills. &
free, call, tool, generation, receive &
students, teaching, teaching climate, climate education, classroom, climate project, teach, educators, lesson, prep &
Climate education advocacy &
Climate education advocacy &
No & Yes & Yes & Yes \\
\hline

Stop Kamala’s Radical Plan for America Before It’s Too Late. Kamala Harris is dangerously liberal, and her extreme policies threaten to fundamentally weaken our nation. She advocates for abolishing fracking, vote harris, patsy, liberal. &
american, job, economy, america, michigan &
kamala, kamala harris, harris, fracking, fighter, abolishing, ban fracking, vote harris, patsy, liberal &
Anti-Harris energy attack &
Anti-Harris energy attack &
No & No & Yes & Yes \\
\hline

Whether you’ve been considering solar panels, harnessing the power of the wind, upgrading to energy-efficient equipment, or taking advantage of federal rebate programs, learn how clean energy can benefit your household. &
program, wind, provide, grant, farming &
infrastructure, investment, clean energy projects, rural america, infrastructure investment and jobs act, bipartisan infrastructure law, rural clean energy, credits, investment, projects &
Clean energy advocacy &
Clean energy advocacy &
Yes & No & Yes & Yes \\
\hline

Utah families are being charged for lawmakers’ fossil fuel favoritism. Contact your legislator today before it’s too late. &
tax, energy, bill, credit, cost &
utah, investinginamerica, iija, act utah, legislation, legislature, lawmakers, legislature's decision, act, bill &
Energy cost alarm &
Energy cost alarm &
Yes & No & Yes & Yes \\
\hline

Already struggling with your electrical bill? It might get worse if lawmakers roll back clean energy incentives. Call your legislator today to protect American energy innovation. &
tax, energy, bill, credit, cost &
credits, rural energy, energy tax, legislation, credit, act, taxes, utah, bill, change &
Solar energy advocacy &
Energy cost alarm &
Yes & No & No & Yes \\
\hline

\end{tabular}
\caption{Evaluation of topic-model keywords and LLM theme assignments using LLM judge (Meta).}
\label{tab:results-table-llm}
\end{table*}
\begin{table*}[t]
\centering
\scriptsize
\setlength{\tabcolsep}{2.5pt}   
\renewcommand{\arraystretch}{1.1}
\begin{tabular}{|p{0.18\textwidth}|p{0.14\textwidth}|p{0.14\textwidth}|p{0.07\textwidth}|p{0.07\textwidth}|p{0.08\textwidth}|p{0.08\textwidth}|p{0.07\textwidth}|p{0.07\textwidth}|}
\hline
\textbf{ad\_text} &
\textbf{lda\_keywords} &
\textbf{bert\_keywords} &
\multicolumn{2}{c|}{\textbf{LLM-Assigned Themes}} &
\multicolumn{4}{c|}{\textbf{Human Judge Assignments}} \\
\cline{4-9}
 &  &  &
\textbf{Theme-Based} &
\textbf{Summary-Based} &
\textbf{lda\_correct} &
\textbf{bert\_correct} &
\textbf{Theme Correct} &
\textbf{Summary Correct} \\
\hline

Data can’t tell the whole story, but it’s a powerful tool for understanding changes on our planet. Grab this article from our free collection of resources to teach your students essential data literacy skills. &
free, call, tool, generation, receive &
students, teaching, teaching climate, climate education, classroom, climate project, teach, educators, lesson, prep &
Climate education advocacy &
Climate education advocacy &
No & Yes & Yes & Yes \\
\hline

Stop Kamala’s Radical Plan for America Before It’s Too Late. Kamala Harris is dangerously liberal, and her extreme policies threaten to fundamentally weaken our nation. She advocates for abolishing fracking, vote harris, patsy, liberal. &
american, job, economy, america, michigan &
kamala, kamala harris, harris, fracking, fighter, abolishing, ban fracking, vote harris, patsy, liberal &
Anti-Harris energy attack &
Anti-Harris energy attack &
No & Yes & Yes & Yes \\
\hline

Whether you’ve been considering solar panels, harnessing the power of the wind, upgrading to energy-efficient equipment, or taking advantage of federal rebate programs, learn how clean energy can benefit your household. &
program, wind, provide, grant, farming &
infrastructure, investment, clean energy projects, rural america, infrastructure investment and jobs act, bipartisan infrastructure law, rural clean energy, credits, investment, projects &
Clean energy advocacy &
Clean energy advocacy &
Yes & No & Yes & Yes \\
\hline

Utah families are being charged for lawmakers’ fossil fuel favoritism. Contact your legislator today before it’s too late. &
tax, energy, bill, credit, cost &
utah, investinginamerica, iija, act utah, legislation, legislature, lawmakers, legislature's decision, act, bill &
Energy cost alarm &
Energy cost alarm &
Yes & No & Yes & Yes \\
\hline

Already struggling with your electrical bill? It might get worse if lawmakers roll back clean energy incentives. Call your legislator today to protect American energy innovation. &
tax, energy, bill, credit, cost &
credits, rural energy, energy tax, legislation, credit, act, taxes, utah, bill, change &
Solar energy advocacy &
Energy cost alarm &
Yes & No & No & Yes \\
\hline

\end{tabular}

\caption{Evaluation of topic-model keywords and LLM theme assignments using human judge (Meta).}
\label{tab:results-table-human}
\end{table*}

\subsection{Evaluation on Bluesky Dataset}
\label{app:eval_bsky}
Table ~\ref{tab:combined-accuracy-comparison} shows the human and LLM judge evaluation results of theme-assignment on the Bluesky dataset. According to both the human and LLM judgment settings, our Text-to-Theme method outperforms Text-to-Summary, as well as the LDA and BERTopic baseline methods of assignment. This result mirrors the core finding that we obtain from our Meta dataset that concise, theme-level labels and prompting often provide a more robust mapping than the full cluster summaries. 

Another key point is that BERTopic's lower (and in some cases comparative) accuracy to LDA is often a result of its high rate of outliers or clusters that do not map cleanly to topics. One such topic that exemplifies this is: [`alt', `alt text', `note', `text', `spicy', `plan fix', `ouch', `aim make', `salty', `thank doing']. In total, 9418 posts were assigned as outliers using BERTopic, whereas for our summary-based assignment, only 1 post is left unassigned, and for the theme-based assignment, only 6 posts are unassigned. For LDA, each post gets assigned a set of keywords, even if this set is not necessarily fitting to the post text. Therefore, the overall accuracy for LDA might appear higher than for BERTopic with both the human and LLM judges, but this is due to the large number of unassigned posts with BERTopic. 

An additional observation is the overall increase in accuracy for summary-based assignment on Bluesky compared to Meta. Although both theme-based methods show around 67\% accuracy according to the human judge (66.6\% for Bluesky, 67.3\% for Meta), there is a discrepancy in summary-based accuracy. This discrepancy is also very clear with the LLM evaluation results, where Bluesky Text-to-Summary shows 45.8\% accuracy and Meta Text-to-Summary shows 59.8\% accuracy. Bluesky has $46$ total themes, whereas Meta has $47$, and as such the corresponding summary sets differ in both size and informational density. Because summary-based assignment requires presenting the LLM with a list of candidate summaries during text-to-summary matching, a smaller summary set effectively reduces the prompt context window and the number of competing semantic options the model must consider. This reduction appears to benefit Bluesky in particular, where fewer summaries lead to less semantic overlap and lower cognitive load for the LLM during assignment. The difference in summary-based accuracy for Meta may also be affected by the type of dataset: Meta's dataset contains a larger number of more narrowly defined themes, often driven by campaign-specific messaging and targeted issue framing. As a result, summary-based assignment on Meta may be more sensitive to semantic mismatches between individual posts and their corresponding cluster summaries, leading to lower overall accuracy relative to Bluesky.

When comparing human judgments with those produced by the LLM judge, we observe systematic differences in the relative accuracy ranges across methods. LDA and BERTopic produce compact keyword lists, which align closely with the token-level evidence an LLM judge is optimized to evaluate. As a result, the LLM judge may systematically favor these methods based on the presence of salient keywords, even when the overall semantic framing of the post is only weakly captured. In contrast, human annotators place greater emphasis on interpretability and semantic coherence. The descriptive, high-level theme labels generated by our approach more directly convey the underlying intent and meaning of the posts, making them easier for humans to evaluate accurately.
This divergence in evaluation criteria helps explain why the LLM judge exhibits a relative bias toward LDA. The effect is less pronounced for BERTopic, likely due to its high proportion of outlier assignments and loosely formed clusters, which reduce the consistency of its keyword sets. We further observe that both Text-to-Summary and Text-to-Theme achieve lower accuracy under LLM-based evaluation than under human judgment. This gap likely occurs because LLM judges rely more on direct word overlap between the label and the text than on deeper semantic interpretation, which makes abstract theme labels harder to assess than explicit keyword-based representations.

We see that, although our Text-to-Summary and Text-to-Theme assignment variants achieve higher accuracy than LDA and BERTopic, TopicGPT, much like for Meta, achieves the highest accuracy overall by far. However, much of this accuracy can be attributed to the highly generic themes derived from the TopicGPT assignments. For example, "Energy" applies to a very wide range of the climate-related dataset. We provide the full lists of TopicGPT generated topics for Bluesky and Meta such that the generality of the topic set can be noticed in Tables ~\ref{tab:level1-topics-bluesky}, \ref{tab:level2-topics-bluesky} \ref{tab:meta-level1-topics}, \ref{tab:meta-level2-topics}.

\begin{table}[h]
    \centering
    \textbf{Level 1 Topics}
    \begin{tabular}{|p{0.9\linewidth}|}
        \hline
        Agriculture, Climate Science, Climate Change Impacts and Adaptation,
        Emergency Response, Energy, Energy Policy and Transition,
        Environment, Climate Change, Health,
        Social Justice, Climate Justice, Sustainable Development
        \\
        \hline
    \end{tabular}
    \caption{TopicGPT Level 1 topic labels for Bluesky}
    \label{tab:level1-topics-bluesky}
\end{table}

\begin{table}[h]
    \centering
    \textbf{Level 2 Topics}
    \begin{tabular}{|p{0.9\linewidth}|}
        \hline
        Agricultural Technology, Clean Energy and Technological Responses,
        Climate Change, Climate Change Communication and Public Perception,
        Climate Change Impacts and Adaptation, Climate Change Policy and Regulation,
        Climate Justice, Climate Justice and Socioeconomic Issues,
        Climate and Environment, Disaster Management,
        Disease Prevention and Health Behavior, Economic and Social Equity,
        Energy, Energy Policy and Transition, Environment,
        Environmental Impact of Agriculture, Environmental Impact on Society and Health,
        Environmental Policy and Governance, Fossil Fuels, Health,
        Health Policy and Social Issues, Pollution, Public Health and Environment,
        Renewable Energy, Social Justice, Social and Economic Sustainability,
        Sustainable Agriculture, Sustainable Development,
        Sustainable Infrastructure and Mobility
        \\
        \hline
    \end{tabular}
    \caption{TopicGPT Level 2 topic labels for Bluesky}
    \label{tab:level2-topics-bluesky}
\end{table}

\begin{table}[h]
    \centering
    \textbf{Level 1 Topics}
    \begin{tabular}{|p{0.9\linewidth}|}
        \hline
        Agriculture, Sustainable Farming, Economics, Energy,
        Energy Policy and Regulation, Energy Technology and Innovation,
        Renewable Energy, Environment, Conservation and Natural Resources,
        Sustainable Energy and Technology, Health and Well-being,
        Human Experience, Social Issues and Community Well-being, Infrastructure,
        Transportation Infrastructure, Politics,
        Political Campaigns and Elections, Political Discourse and Public Sentiment,
        Public Policy and Governance, Science
        \\
        \hline
    \end{tabular}
    \caption{TopicGPT Level 1 topic labels for Meta}
    \label{tab:meta-level1-topics}
\end{table}

\begin{table}[h]
    \centering
    \textbf{Level 2 Topics}
    \begin{tabular}{|p{0.9\linewidth}|}
        \hline
        Agriculture and Land Use, Climate Change and Emissions,
        Conservation and Natural Resources, Cultural and Personal Expression,
        Economic Development, Economic Education, Economics,
        Education and Leadership Development, Energy,
        Energy Access and Affordability, Energy Policy and Regulation,
        Energy Technology and Innovation, Energy and Community Impact,
        Environment, Environmental Justice and Community Action,
        Environmental Science and Climate Research,
        Environmental and Climate Concerns, Health and Well-being,
        Housing and Urban Development Infrastructure, Infrastructure,
        Political Campaigns and Elections,
        Political Discourse and Public Sentiment, Politics,
        Public Policy and Governance, Pollution and Waste Management,
        Renewable Energy, Science, Scientific Research and Education,
        Social Issues and Community Well-being,
        Social and Environmental Justice, Sustainable Energy and Technology,
        Sustainable Farming, Transportation Infrastructure, Water Infrastructure
        \\
        \hline
    \end{tabular}
    \caption{Level 2 topic labels for Meta}
    \label{tab:meta-level2-topics}
\end{table}

To validate our annotation results for the Bluesky dataset, we look at Inter-annotator agreement scores for our methods as well as TopicGPT. We see that for our method, there is 71.4\% raw agreement between the human and LLM judge ($\kappa = 0.44$) for the summary-based judgments, and furthermore, a 70.4\% raw agreement ($\kappa = 0.40$) for theme-based judgments. Both Cohen's Kappa scores fall in the moderate agreement range. Further analysis suggests that the disagreements are not random: the LLM judge is systematically more conservative, achieving higher precision on positive labels (~0.86–0.87) but lower recall (~0.64). This means the LLM judge tends to miss true positives flagged by human annotators, rather than hallucinating false positives. This asymmetry points to a somewhat consistent difference in labeling threshold between the two judges, rather than unreliable or noisy behavior. For TopicGPT, we see 90.2\% agreement between the human and LLM judge with a Cohen's Kappa score of 0.43.
\begin{figure}[t]
    \centering
    \includegraphics[width=1.0\linewidth]{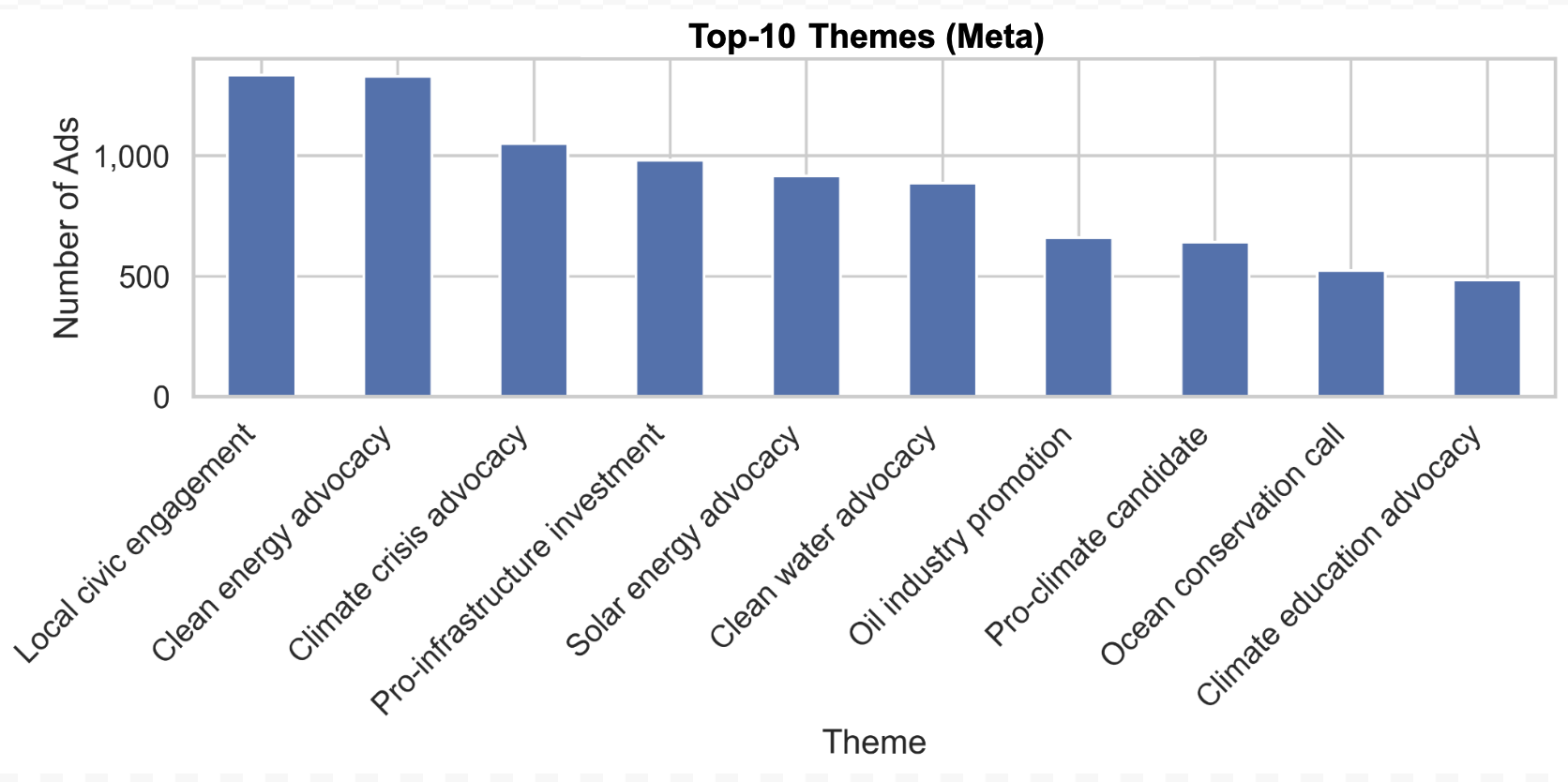}
    \caption{Top 10 themes by number of ads for Meta data. }
    \label{fig:meta-themes}
\end{figure}
\subsection{Human Validation}
\label{app:human}
Seven NLP and CSS researchers, age range 20-40, 3 males and 4 females, did three $\approx1.5$-hour sessions to validate whether the generated and refined themes, as well as text $\rightarrow$ theme mapping, are correct or not for each dataset. In total, it took $\approx 11$ hours for both datasets. Annotators were provided with the same instructions that were provided to the LLM prompt. The annotators included advanced undergraduate students, graduate students, and faculty.
\begin{figure*}
    \centering
    \includegraphics[width=1.0\linewidth]{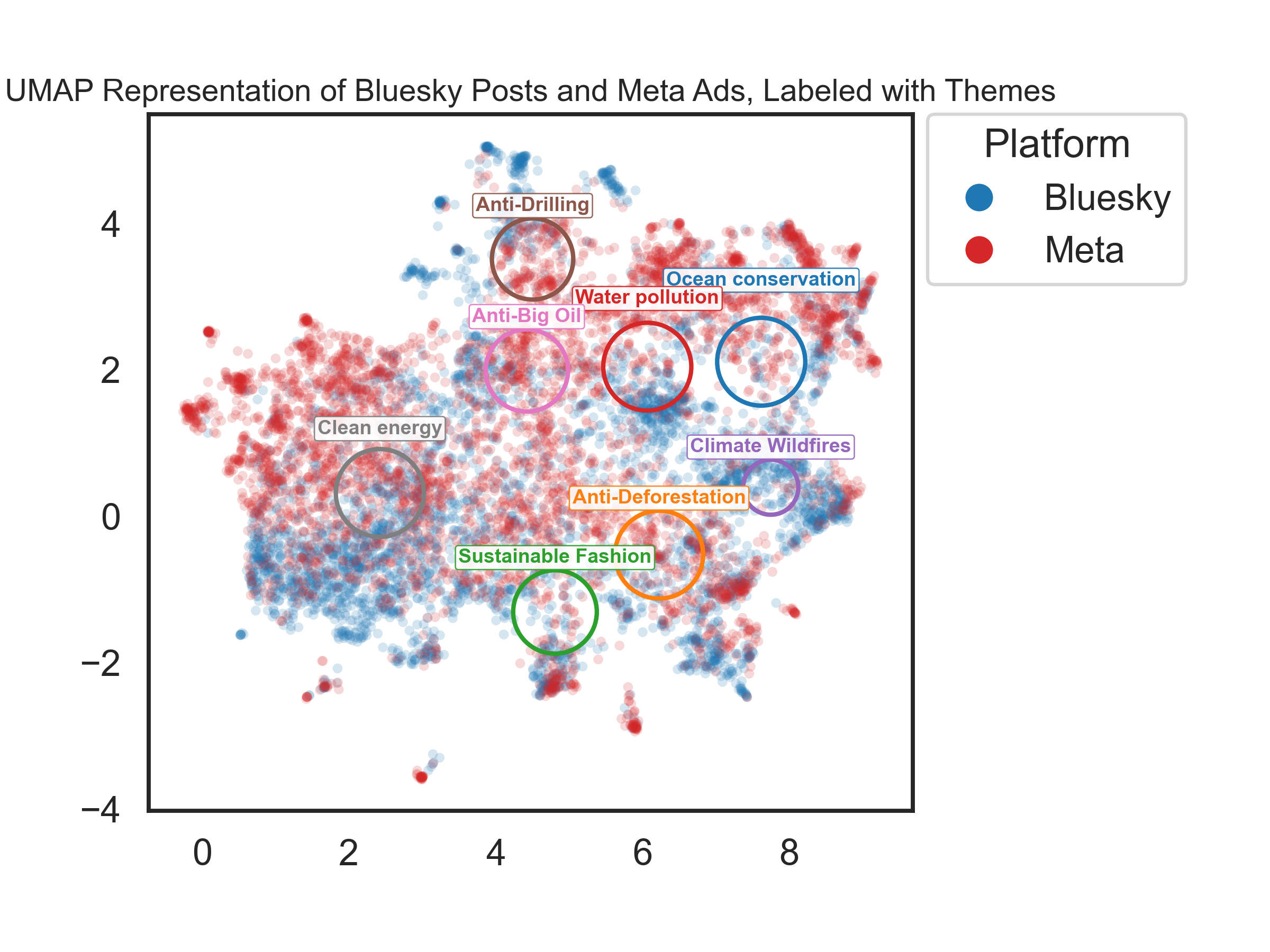}
    \caption{UMAP representation of Meta ads and Bluesky posts, annotated with themes.} 
    \vspace{-10pt}
    \label{fig:umap}
    \vspace{-5pt}
\end{figure*}

\section{Joint-Clustering Details}
\label{app:joint}

To support direct cross-platform comparison in \S\ref{sec:thematic}, 
we re-run a modified version of our clustering pipeline on the joint 
combination of all Meta and Bluesky texts ($36{,}206$ total). The 
pipeline follows the same stages as the per-platform pipeline 
(App.~\ref{app:experim}) but operates on the merged embedding space. 
We retain hierarchical structure by extracting both top-level clusters 
and sub-clusters within each top-level group, producing $140$ top-level 
clusters and $471$ sub-clusters. Coherence filtering, summarization, 
and theme labeling are then applied to sub-clusters as in the 
per-platform pipeline, yielding $354$ coherent themes.

We define three theme categories based on within-theme platform 
composition. A theme is \textit{Meta-dominant} if at least $70\%$ of 
its texts come from Meta, \textit{Bluesky-dominant} if at least $70\%$ 
come from Bluesky, and \textit{shared} otherwise. The $70\%$ threshold 
ensures that dominant themes reflect platform-specific discourse rather 
than minority representation, while still admitting a meaningful number 
of shared themes. The resulting partition contains $145$ Meta-dominant, 
$112$ Bluesky-dominant, and $97$ shared themes.

\subsection{Shared Themes}
\label{app:shared-themes}

\begin{figure}[h]
    \centering
    \includegraphics[width=1.0\linewidth]{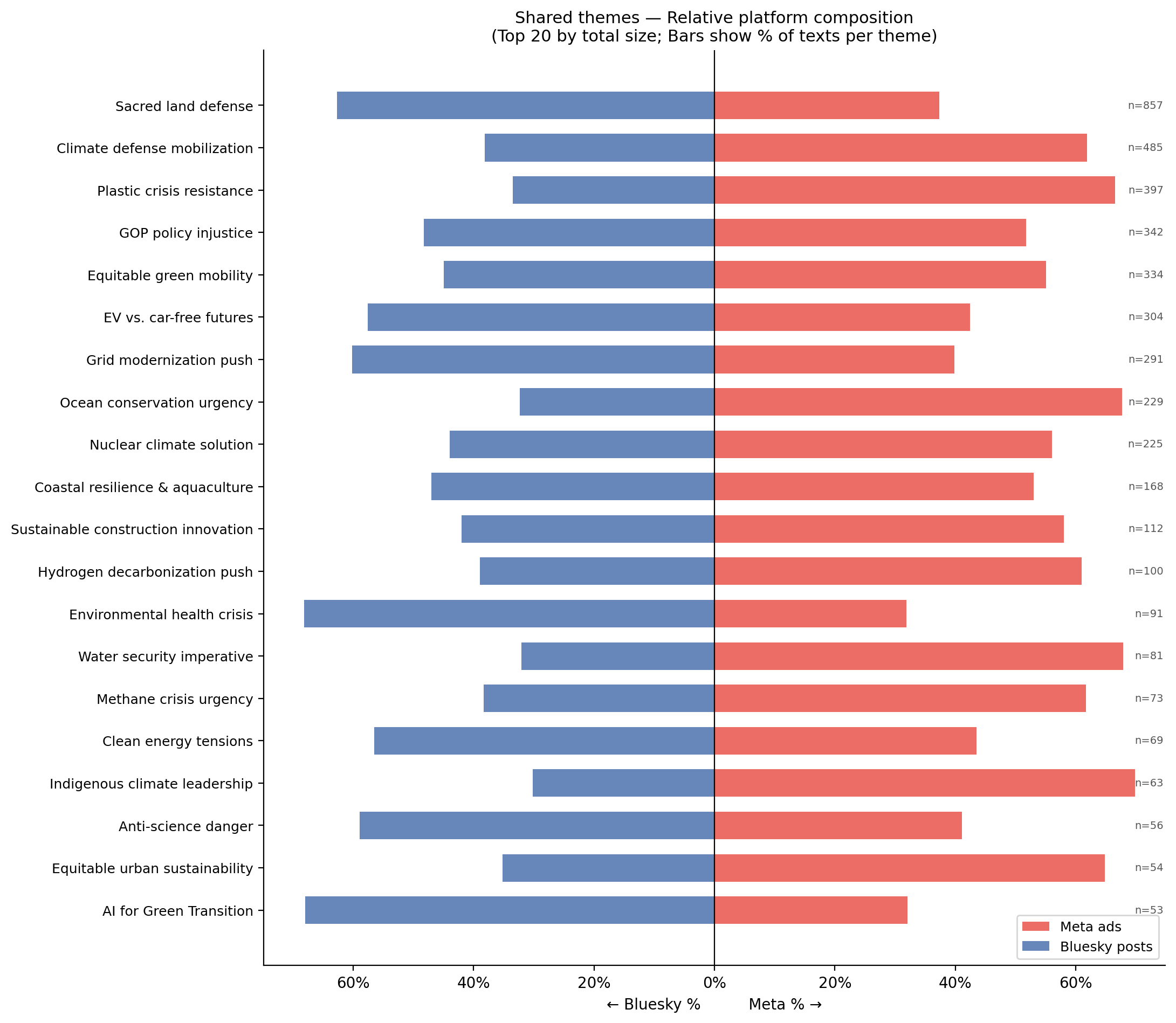}
    \caption{Top 20 shared themes with similar numbers of Meta and Bluesky texts.}
    \label{fig:shared-themes}
\end{figure}

\begin{figure}[h]
    \centering
    \includegraphics[width=1.0\linewidth]{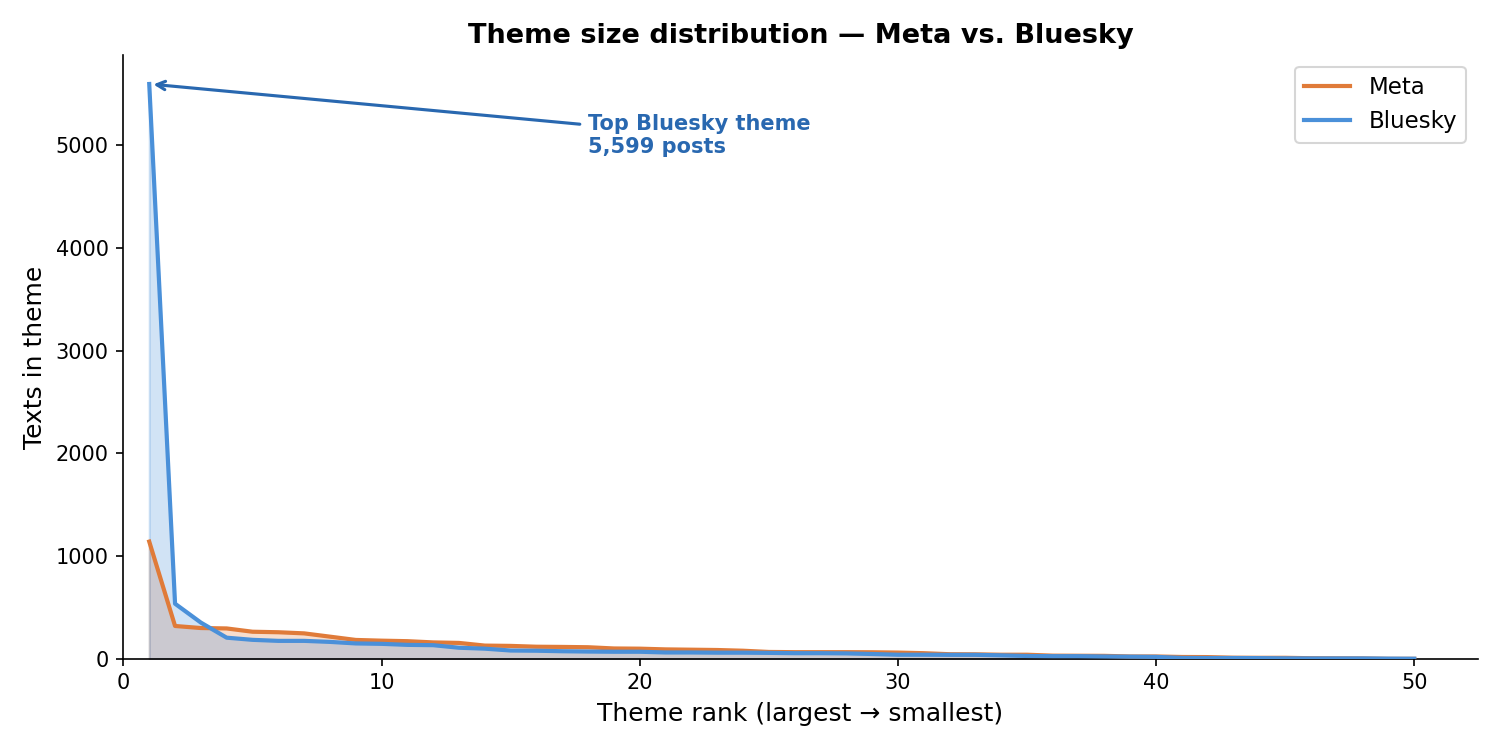}
    \caption{Distribution of joint themes for Meta vs. Bluesky}
    \label{fig:joint-theme-dist}
\end{figure}

Fig.~\ref{fig:shared-themes} shows the top $20$ shared themes ranked by total cluster size. Shared themes span a range of climate-relevant 
topics (e.g., \textit{Sacred land defense}, \textit{Climate defense 
mobilization}, \textit{Plastic crisis resistance}) and provide a 
contrast to the strongly platform-dominant themes discussed in 
\S\ref{sec:thematic}. Furthermore, Fig.~\ref{fig:joint-theme-dist} shows the distribution of joint themes for the Meta and Bluesky datasets. We see that Bluesky discourse concentrates heavily on a small number of top themes, whereas Meta discourse does not see as sharp of a long-tailed distribution.

\section{Named Entity Recognition}

\subsection{Pipeline}
\label{app:ner_pipeline}

We apply a Named Entity Recognition pipeline combining SpaCy's pretrained \texttt{en\_core\_web\_lg} model with a custom climate-domain entity dictionary and alias normalization map. The pipeline runs in two stages per text: a domain-dictionary pass that matches multi-word climate-relevant phrases that SpaCy might miss (e.g., \textit{climate change}, \textit{Inflation Reduction Act}, \textit{Paris Agreement}), followed by SpaCy's NER for general-purpose entity extraction. 

Extracted entities are normalized via an alias map that consolidates phrases which refer to the same entity: people (e.g., \textit{Trump}, 
\textit{President Trump}, \textit{Donald Trump} $\rightarrow$ 
\textit{Donald Trump}), places (\textit{U.S.}, \textit{USA}, 
\textit{America}, \textit{United States of America} $\rightarrow$ 
\textit{United States}), organizations (\textit{GOP}, 
\textit{Republicans}, \textit{Republican Party} $\rightarrow$ 
\textit{Republican Party}), and policy/issue concepts 
(\textit{climate crisis}, \textit{global warming} $\rightarrow$ 
\textit{climate change}; \textit{IRA} $\rightarrow$ 
\textit{Inflation Reduction Act}).

We aggregate recognized entities into four high-level analytic types by mapping SpaCy's labels and our custom dictionary entries into a common taxonomy: PERSON (SpaCy's PERSON); ORG (SpaCy's ORG, plus domain-specific organizational entries such as \textit{fossil fuel industry} and \textit{MAGA}); GPE/LOC/GROUP (SpaCy's GPE, LOC, NORP, and FAC); and POLICY/EVENT/ISSUE (SpaCy's EVENT and LAW, plus all custom climate-policy and climate-issue entries). SpaCY labels for dates, quantities, money, products, etc. are excluded as not informative for our analysis. Within each text, entity mentions are deduplicated such that repeated references to the same entity count once per text. Reported entity-salience percentages in App. ~\ref{app:ner_results} are computed as the proportion of texts mentioning each entity at least once. 

\subsection{Results}
\label{app:ner_results}

\begin{figure}[h]
    \centering
    \includegraphics[width=1.0\linewidth]{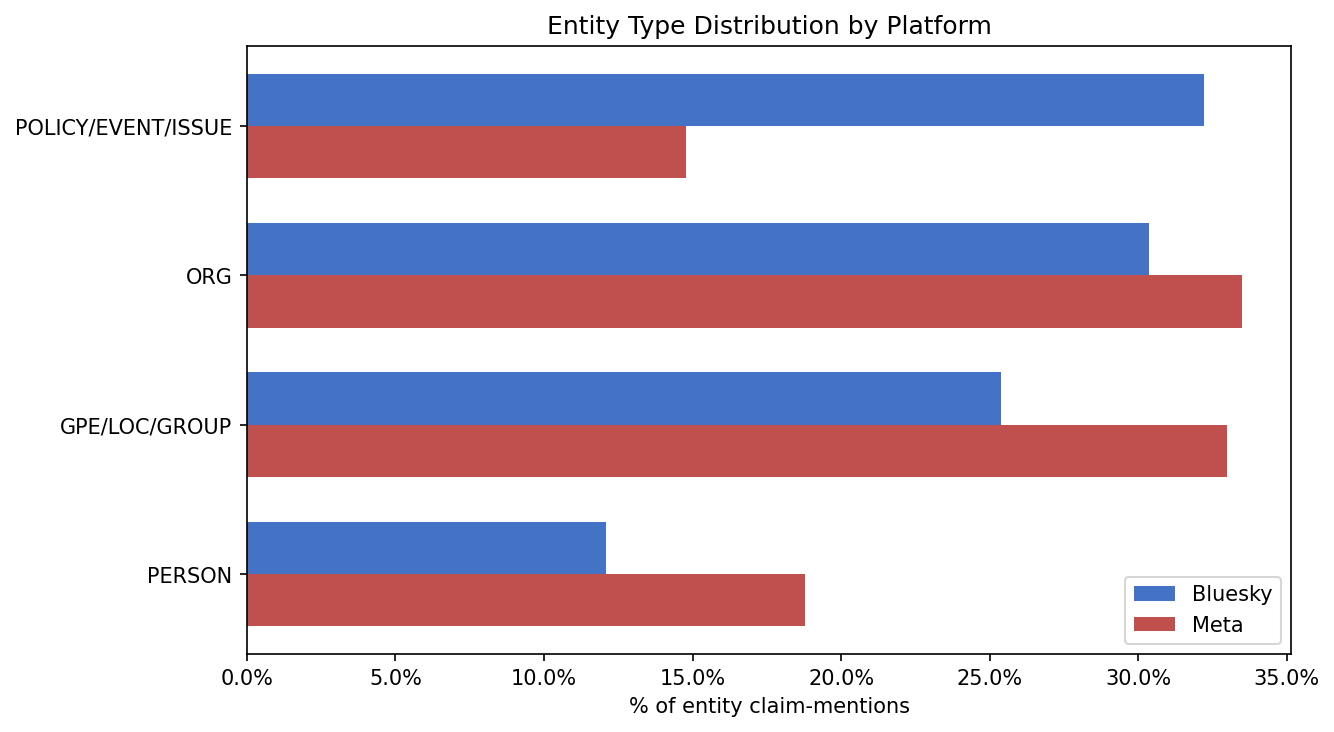}
    \caption{Named Entity Type Distribution for the Bluesky and Meta datasets.}
    \label{fig:entity-type-distribution}
\end{figure}

\paragraph{Entity Type Distribution.}
Entity-type prevalence differs sharply by platform (Fig. ~\ref{fig:entity-type-distribution}). Bluesky contains substantially more POLICY/EVENT/ISSUE entities than Meta, while Meta dominates the GPE/LOC/GROUP and PERSON categories.  This suggests that Meta ads more frequently invoke specific groups, locations, and named 
individuals, while Bluesky discourse more frequently invokes general policy concepts. ORG entities are common on both platforms.

\begin{figure*}
    \centering
    \includegraphics[width=1.0\linewidth]{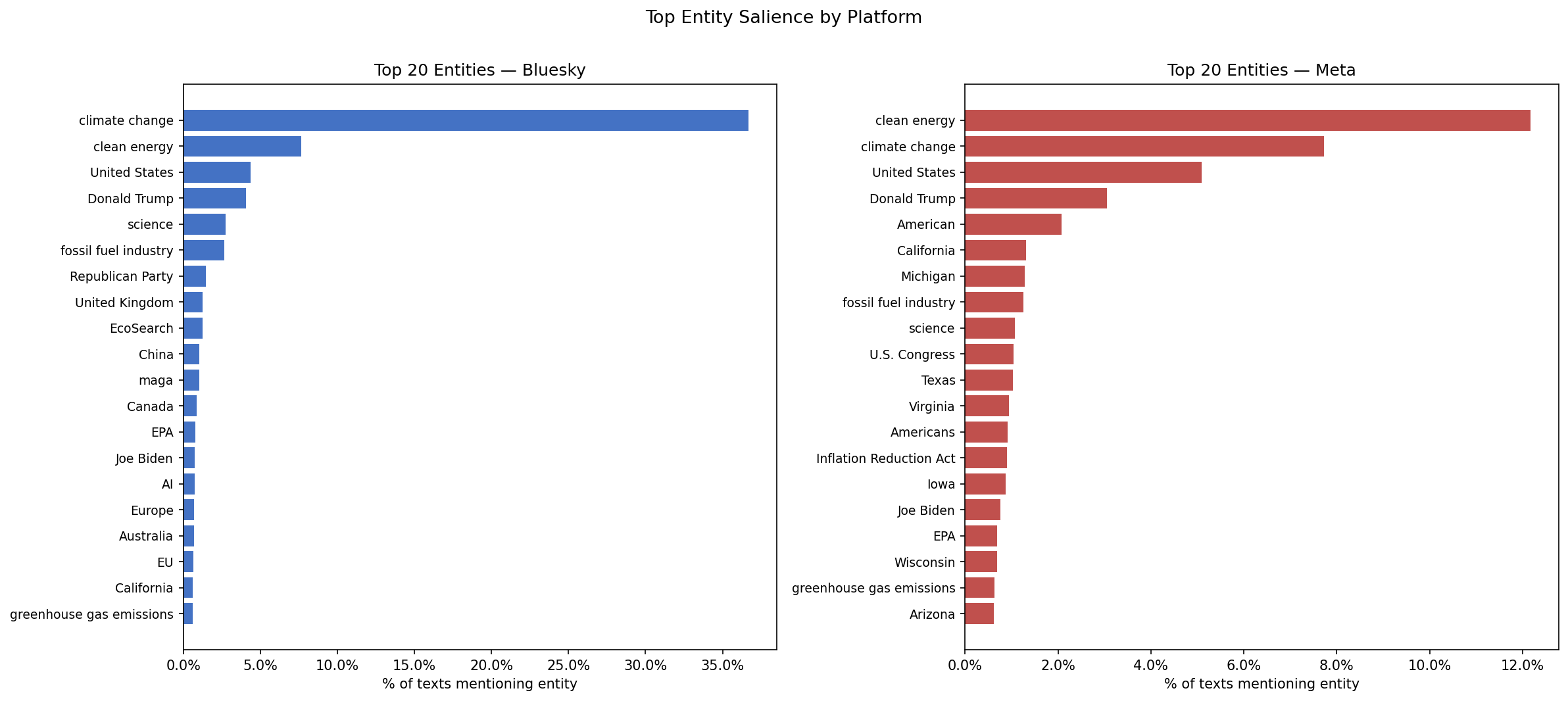}
    \caption{Top 20 entities (\% of total texts) by platform.}
    \label{fig:top-entity-salience}
\end{figure*}

\paragraph{Top Entity Salience.}
The most frequent entities on both platforms are general 
(\textit{clean energy}, \textit{climate change}), with long-tailed 
distributions thereafter (Fig.~\ref{fig:top-entity-salience}). Bluesky 
features prominent references to countries and continents 
(\textit{United States}, \textit{United Kingdom}, \textit{Europe}, 
\textit{Australia}), while Meta features specific U.S. states 
(\textit{Wisconsin}, \textit{Iowa}, \textit{Michigan}, 
\textit{California}), consistent with state-targeted advertising 
delivery.

\begin{figure*}
    \centering
    \includegraphics[width=1.0\linewidth]{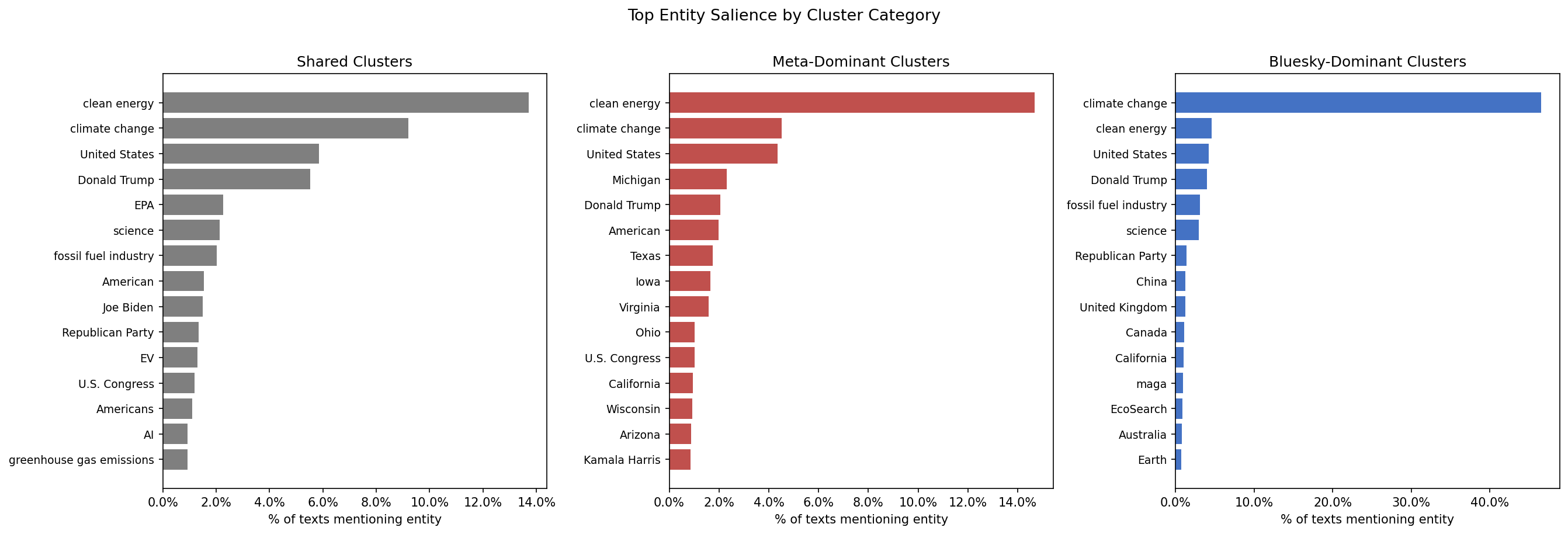}
    \caption{Top 10 entity salience for the three cluster categories: Shared, Meta-Dominant, and Bluesky-Dominant.}
    \label{fig:entity-cluster-category}
\end{figure*}

\paragraph{Entity Salience by Cluster Category.}
Disaggregating by joint-clustering category 
(Fig.~\ref{fig:entity-cluster-category}), Meta-dominant clusters 
contain more U.S. state names while Bluesky-dominant clusters center 
on countries and large-scale collective groups (including 
\textit{Earth}). This reinforces the thematic-structure finding in 
\S\ref{sec:thematic}: Meta-dominant discourse is found to be geographically 
targeted, while Bluesky-dominant discourse operates at a larger systemic scale.

\section{Rhetorical Register: Additional Analyses}
\label{app:register}

This appendix provides the lexicon construction details, cluster- and 
demographic-level framing breakdowns, and theme-level LIWC profiles 
that support the main-text analysis in \S\ref{sec:rhetoric}.

\subsection{Entman Framing Lexicon Construction}
\label{app:lexicon}

We construct a framing lexicon seeded with five hand-defined terms per 
subcategory, each accompanied by a category description. We then 
prompt Mistral-Large-Instruct-2407 to generate $40$ additional terms 
conditioned on each category's description and seed examples, yielding 
a lexicon of approximately $45$ terms per subcategory. Framing scores 
are computed as the proportion of lexicon-matched tokens to total 
tokens in each text.

\subsection{Cluster-Level Framing}
\label{app:cluster-framing}

\begin{figure*}
    \centering
    \includegraphics[width=1.0\linewidth]{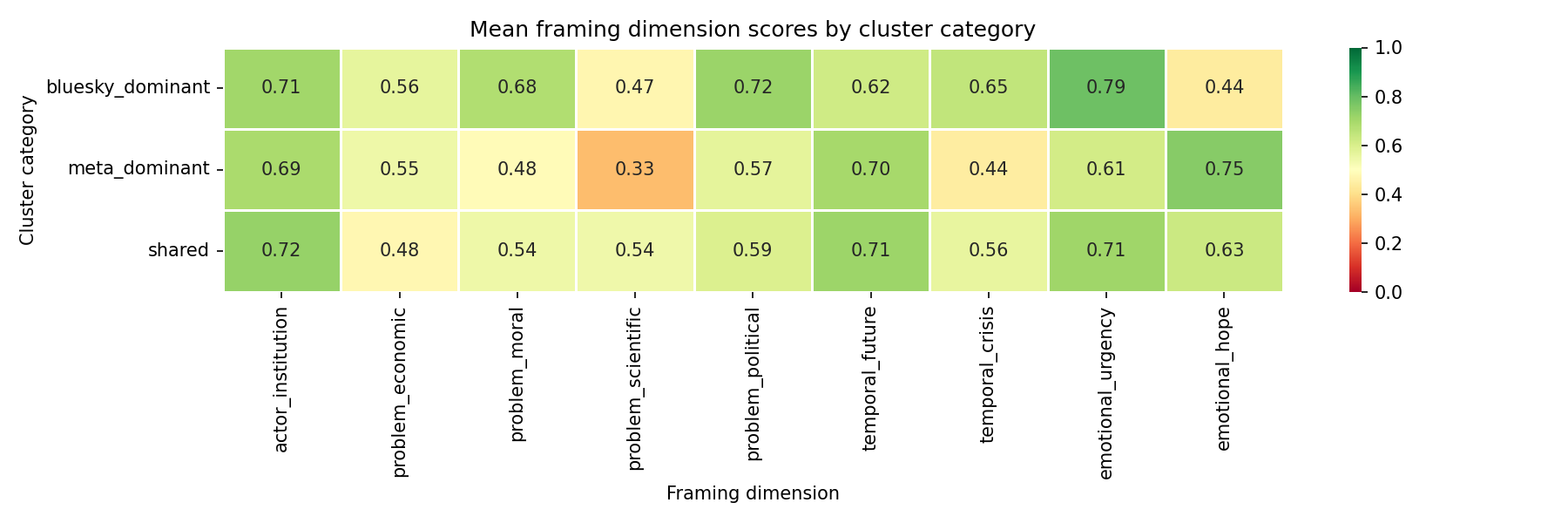}
    \caption{Mean Entman framing dimension scores per cluster category.}
    \label{fig:framing-heatmap}
\end{figure*}

Fig.~\ref{fig:framing-heatmap} presents mean framing scores across 
nine dimensions, aggregated by joint-clustering category 
(Bluesky-dominant, Shared, Meta-dominant). Bluesky-dominant clusters 
score highest on emotional urgency framing ($0.79$), political problem 
framing ($0.72$), and institutional actor framing ($0.71$), suggesting 
a discourse centered on systemic accountability and urgent risk. 
Meta-dominant clusters score lowest on scientific framing ($0.33$); 
this is notable given that scientific authority framing grounds claims 
in empirical evidence, and suggests that paid climate advertising 
favors emotive and community-oriented persuasion over evidence-heavy 
rhetoric. Shared clusters exhibit the most balanced profile, with 
elevated future framing ($0.71$), institutional actor framing ($0.72$), 
and emotional urgency framing ($0.71$).

\subsection{Demographic Targeting Patterns (Meta)}
\label{app:demo-framing}

The Meta Ad Library exposes certain delivery and demographic metadata (age, gender, state, spend) that allows us to examine how framing varies across audience segments. We compute delivery-weighted framing scores for each segment. Bluesky does not expose comparable metadata, so this analysis is necessarily Meta-only. 

\begin{figure*}
    \centering
    \begin{subfigure}{0.48\linewidth}
        \centering
        \includegraphics[width=\linewidth]{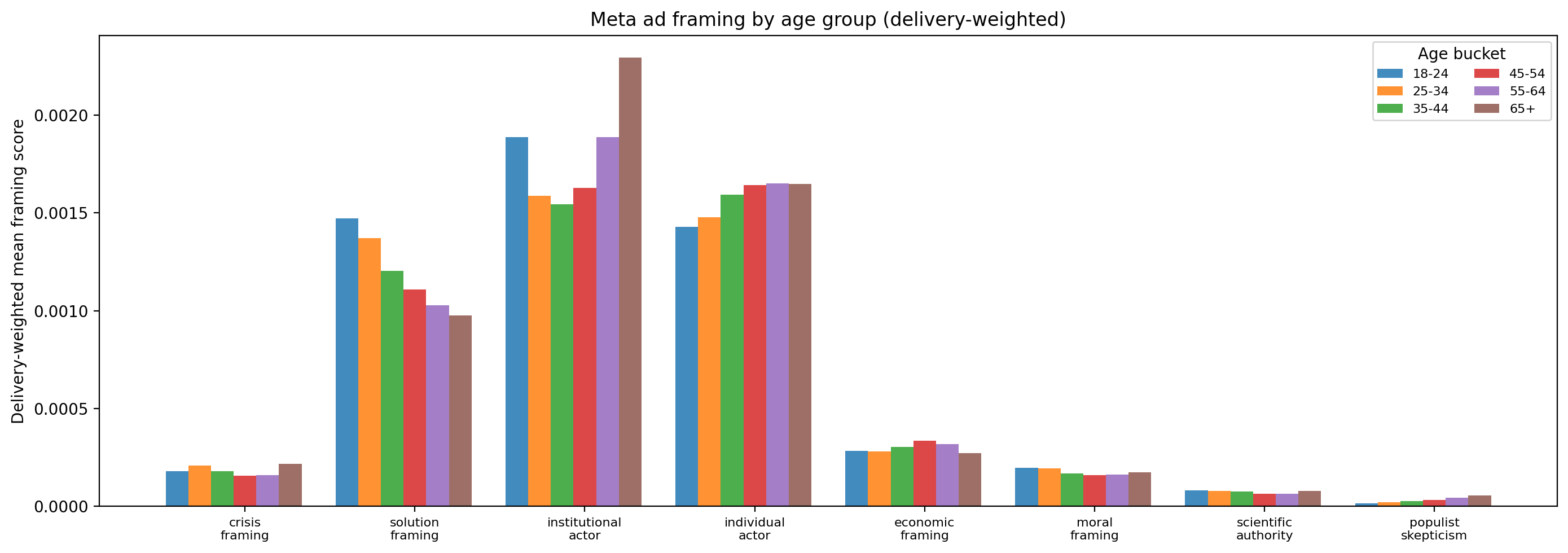}
        \caption{By age group.}
    \end{subfigure}
    \hfill
    \begin{subfigure}{0.48\linewidth}
        \centering
        \includegraphics[width=\linewidth]{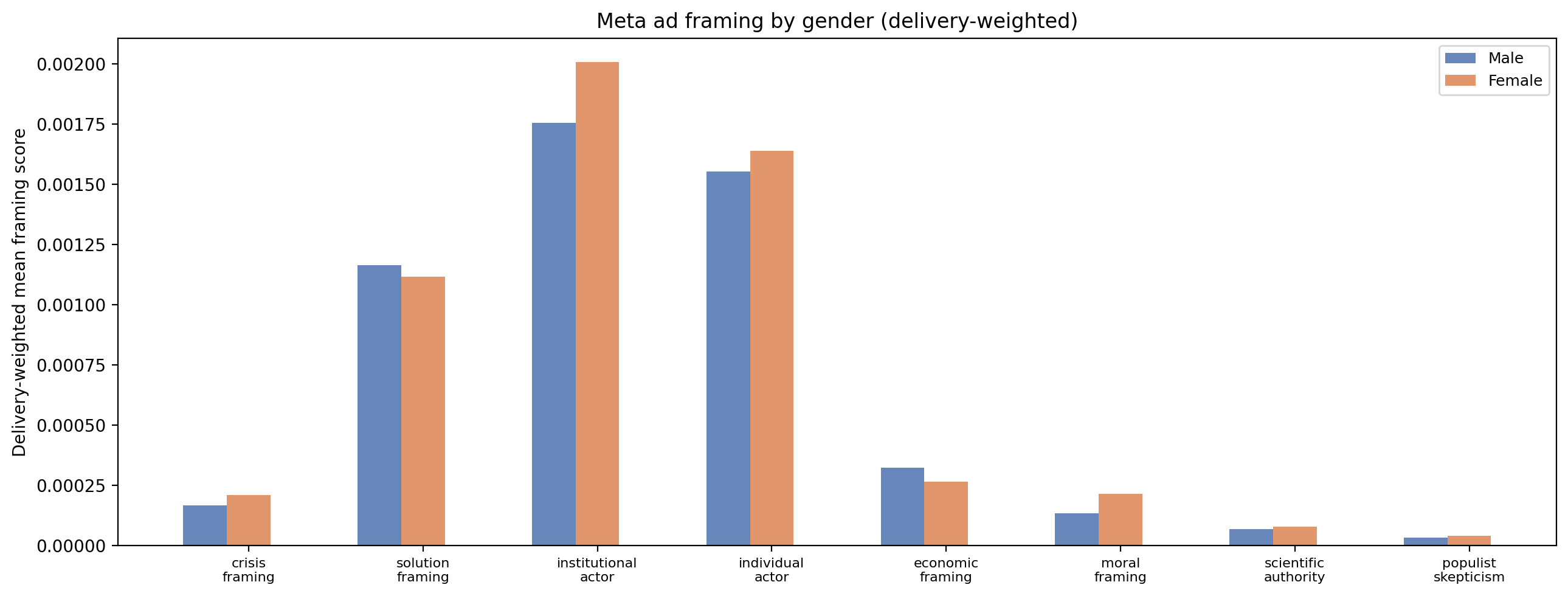}
        \caption{By gender.}
    \end{subfigure}
    \vspace{0.3em}
    \begin{subfigure}{0.48\linewidth}
        \centering
        \includegraphics[width=\linewidth]{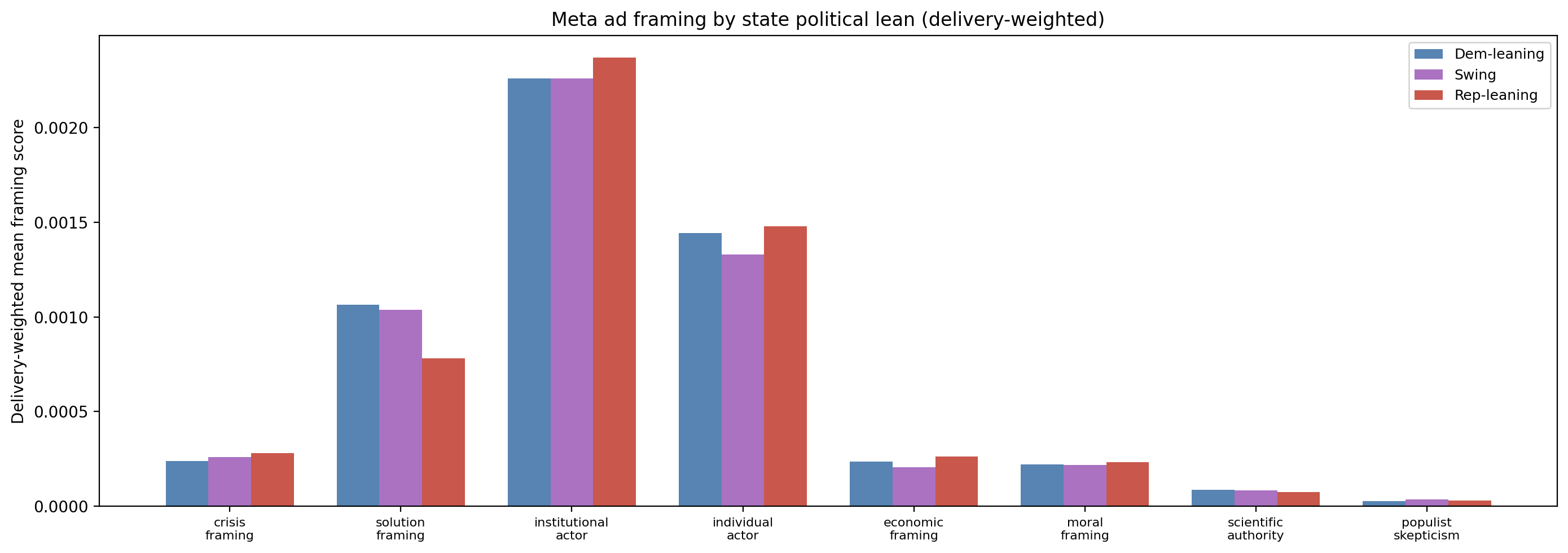}
        \caption{By state political lean.}
    \end{subfigure}
    \hfill
    \begin{subfigure}{0.48\linewidth}
        \centering
        \includegraphics[width=\linewidth]{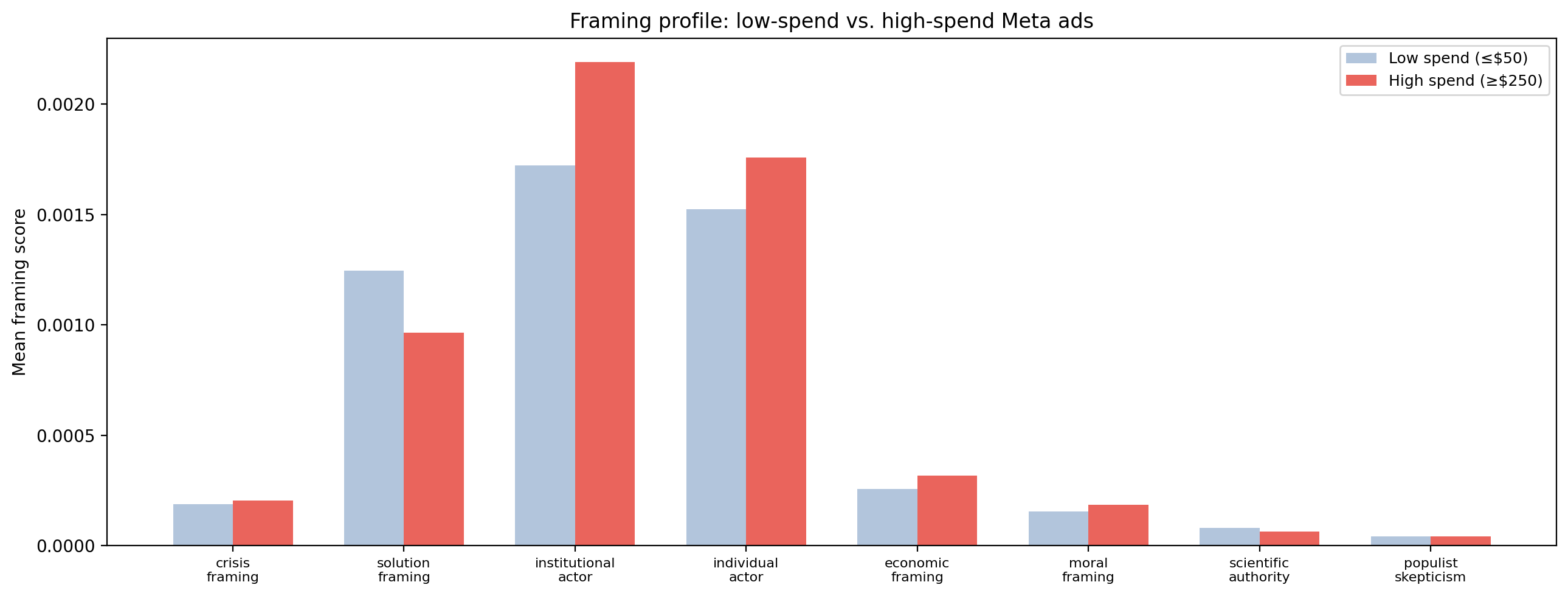}
        \caption{By ad spend.}
    \end{subfigure}
    \caption{Delivery-weighted Entman framing for Meta ads across audience segments.}
    \label{fig:meta-framing-demographic}
\end{figure*}

\paragraph{Age and gender.}
Fig.~\ref{fig:meta-framing-demographic}(a, b) shows several monotone 
trends. Solution framing decreases steadily with age, while individual 
actor framing and populist skepticism increase with age. Institutional 
actor framing forms a U-shape: highest in the 18--24 group, lowest in 
middle-aged groups, and rising again in 65+. Female-targeted ads 
exhibit substantially higher institutional actor framing and moral 
framing than male-targeted ads; male-targeted ads exhibit higher 
economic and solution framing.

\paragraph{Political lean and ad spend.}
Fig.~\ref{fig:meta-framing-demographic}(c) shows that 
democratic-leaning audiences receive the most solution framing and 
scientific authority framing, while republican-leaning audiences 
receive more institutional actor and crisis framing. 
Fig.~\ref{fig:meta-framing-demographic}(d) shows that low-spend ads 
carry more solution framing than high-spend ads, while both 
institutional and individual actor framing are concentrated in 
high-spend ads.

\begin{figure*}
    \centering
    \includegraphics[width=1.0\linewidth]{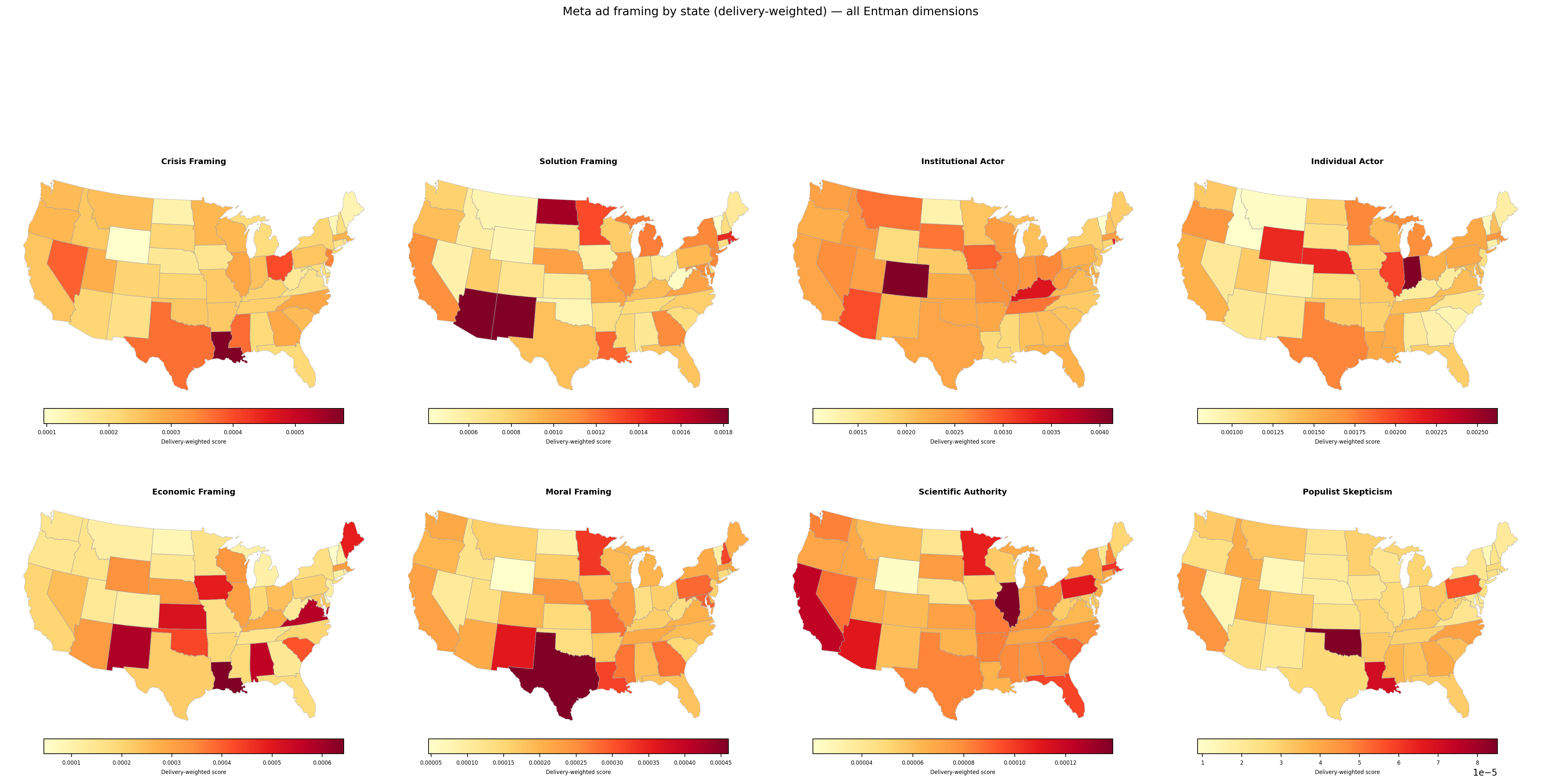}
    \caption{Heatmap of the eight Entman framing dimensions, delivery-weighted by state.}
    \label{fig:meta-framing-map}
\end{figure*}

\paragraph{Geography.}
Fig.~\ref{fig:meta-framing-map} shows delivery-weighted framing 
concentration by state. Scientific authority framing concentrates in 
left-leaning states, consistent with the political-lean pattern above. 
Other framing dimensions show state-level concentrations that are not 
cleanly explained by political lean alone.

\subsection{LIWC Content Categories}
\label{app:liwc-content}

In addition to the four summary variables reported in 
\S\ref{sec:rhetoric}, we analyze eleven LIWC content categories mapped 
to the four Entman framing functions: \textit{tone\_neg}, 
\textit{emo\_anx}, and \textit{risk} for problem definition; 
\textit{cause} and \textit{power} for causal attribution; 
\textit{moral}, \textit{certitude}, and \textit{tentat} for moral 
evaluation; and \textit{achieve} and \textit{focusfuture} for treatment 
recommendation.

\subsection{Theme-Level LIWC Profiles}
\label{app:theme-liwc}

\begin{figure*}
    \centering
    \begin{subfigure}{\linewidth}
        \includegraphics[width=\linewidth]{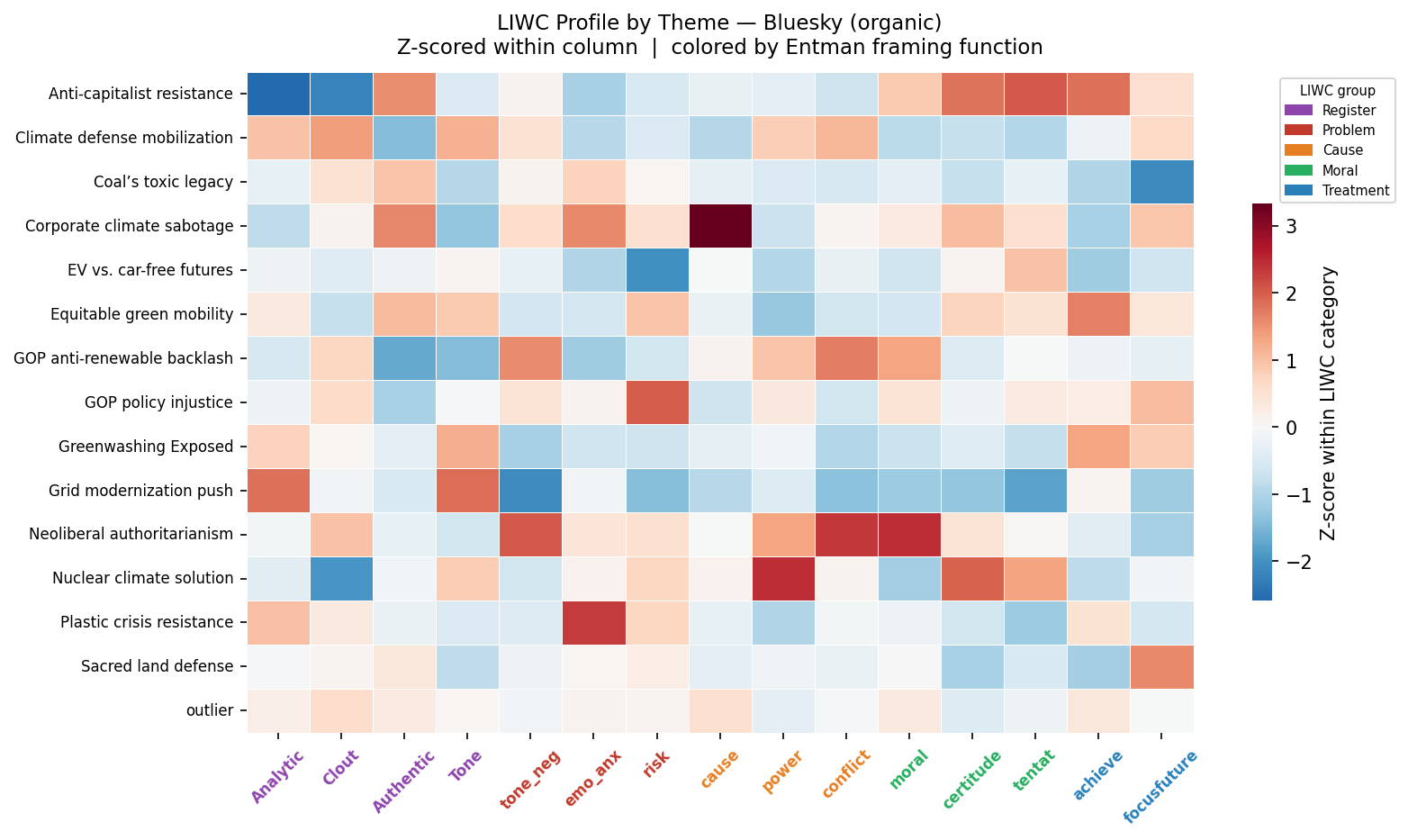}
        \caption{Bluesky (organic).}
    \end{subfigure}
    \vspace{0.5em}
    \begin{subfigure}{\linewidth}
        \includegraphics[width=\linewidth]{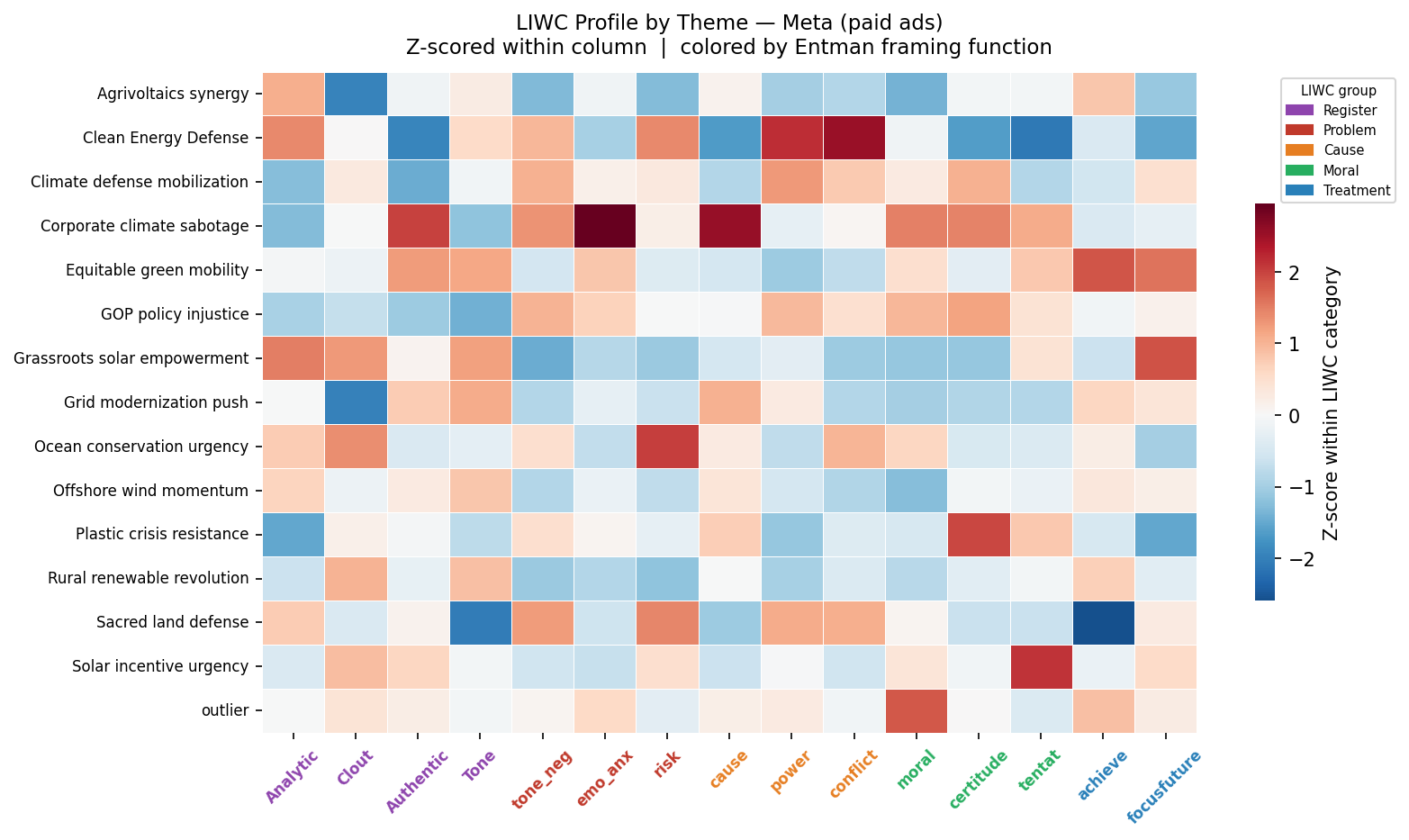}
        \caption{Meta (paid ads).}
    \end{subfigure}
    \caption{LIWC profiles by theme, z-scored within column and colored by Entman framing function.}
    \label{fig:theme_linguistic_heatmaps}
\end{figure*}

Fig.~\ref{fig:theme_linguistic_heatmaps} shows per-theme LIWC profiles 
for Bluesky and Meta, z-scored within each LIWC dimension and 
color-coded by Entman framing function.

\paragraph{Bluesky.} Themes organized around systemic critique exhibit 
distinctive signatures. \textit{Corporate climate sabotage} is the 
strongest outlier on \textit{cause}, reflecting concentrated 
attribution to corporate actors. \textit{Neoliberal authoritarianism} 
peaks on \textit{conflict} and \textit{moral}. \textit{Plastic crisis 
resistance} registers the highest \textit{emo\_anx} score. 
\textit{Anti-capitalist resistance} combines low \textit{Analytic} 
and \textit{Clout} with elevated \textit{certitude}, \textit{achieve}, 
and \textit{tentat}, suggesting an informal, non-authoritative voice 
that mixes strong ideological messages with uncertainty about concrete 
paths forward. \textit{Grid modernization push} is the most 
analytically formal Bluesky theme, diverging from the platform's 
overall register due to the technical nature of the topic.

\paragraph{Meta.} \textit{Clean Energy Defense} exhibits the highest 
\textit{power} and \textit{conflict} scores on the platform. 
\textit{Ocean conservation urgency} registers the highest \textit{risk} 
score. \textit{Corporate climate sabotage} combines high 
\textit{cause}, \textit{emo\_anx}, and \textit{Authentic}, suggesting 
that paid ads invoke corporate blame through an anxious, personally 
engaged voice rather than formal argumentation. \textit{Sacred land 
defense} scores very low on \textit{achieve} and carries negative 
\textit{Tone}, operating less as a prescriptive call to action than 
as a moral grievance framing.

\paragraph{Shared themes.} Themes that appear on both platforms reveal 
that identical conceptual content can be presented through diverging 
linguistic strategies. \textit{Corporate climate sabotage} is a 
high-\textit{cause} theme on both platforms, but Bluesky's rendering 
carries a stronger causal attribution signal while Meta's is 
additionally marked by high \textit{emo\_anx}. \textit{Grid 
modernization push} is analytically formal on Bluesky but 
\textit{Clout}-suppressed on Meta. \textit{Sacred land defense} on 
Meta is \textit{achieve}-suppressed relative to its Bluesky 
counterpart.

\section{Qualitative Analysis Details}

\subsection{Per-Platform Theme Distributions}
\label{app:per-platform-dist}

\begin{figure}[t]
    \centering
    \includegraphics[width=1.0\linewidth]{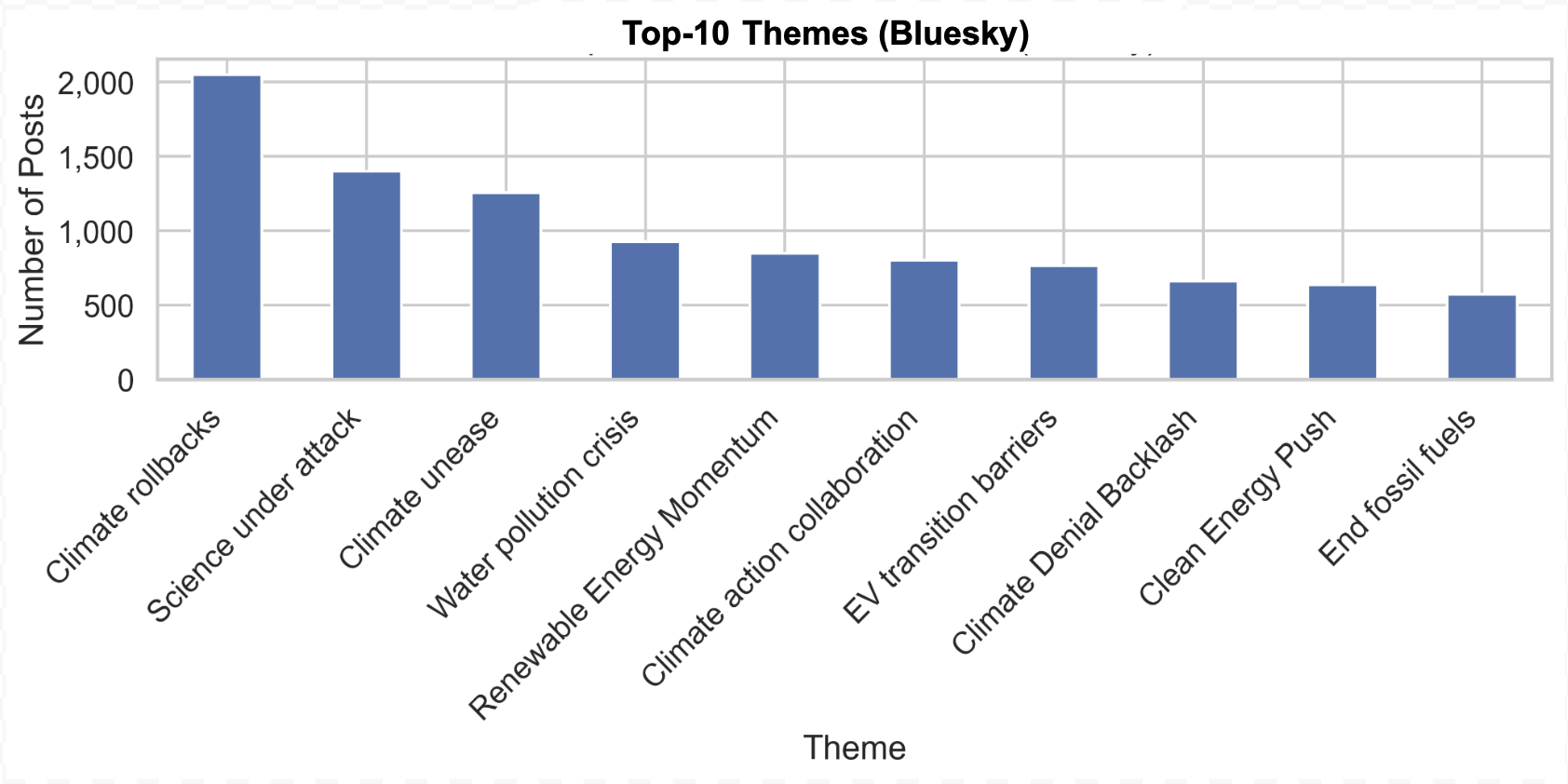}
    \caption{Top 10 themes by number of posts for Bluesky data.}
    \label{fig:bluesky-themes}
\end{figure}

Table~\ref{tab:meta-top-themes} and Fig.~\ref{fig:meta-themes} report 
the top themes by ad count for Meta; Table~\ref{tab:bluesky-top-themes} and 
Fig.~\ref{fig:bluesky-themes} report the corresponding distributions 
for Bluesky.

\begin{table*}
\centering
\begin{tabular}{p{4 cm} r p{10.4 cm}}
\hline
Theme & \# Ads & Example ad text \\
\hline
Local Civic Engagement & 1365 &
Tired of the status quo? Community-led change is THE game-changer. From global crises to climate action, we're your go-to for the latest updates and ways to support local heroes. \\
Clean Energy Advocacy & 1356 &
We’re committed to Environmental Justice \& Social Equity so no community is left behind in our clean energy future. \\
Climate Crisis Advocacy & 1087 &
Mongolia’s herders face rising climate risks. Watch how we are working with partners and the government of Mongolia in solving complex challenges in the country. \\
Pro-infrastructure Investment & 1010 &
These upgrades will make the Bells Mill Road and Valley Green Road bridges more resilient to extreme weather, ensuring safer, stronger connections for generations to come. \\
Solar Energy Advocacy & 929 &
One of the most significant benefits of community solar projects is the potential to reduce energy costs for low-income households. \\
\hline
\end{tabular}
\caption{Top themes for Meta ads.}
\label{tab:meta-top-themes}
\end{table*}

\begin{table*}
\centering
\begin{tabular}{p{4 cm} r p{10.4 cm}}
\hline
Theme & \# Posts & Example text \\
\hline
Climate rollbacks & 2051 & Trump's EPA wants to eliminate regulation for greenhouse gases. \\
Science under attack & 1399 & Science Trump Administration Decommissions Sea Ice Data That Sounded an Alarm on Arctic Climate Change. \\
Climate unease & 1256 & That our kids will come to fear summer rather than look forward to it due to climate change just crushes me every time.
\\
Water pollution crisis & 928 & tudy finds climate change limits River Thames cleanup efforts by accelerating algal growth. \\
Renewable Energy Momentum & 848 & Democrats invested \$400,000,000,000.00 in renewable energy during Biden's term. \\
\hline
\end{tabular}
\caption{Top themes for Bluesky posts. }
\label{tab:bluesky-top-themes}
\end{table*}

\subsection{Full Theme--Stance Correlation}
\label{app:qual_bsky}
\begin{figure*}
\centering
\begin{subfigure}{.9\columnwidth}
  \centering
  \includegraphics[width=\textwidth]{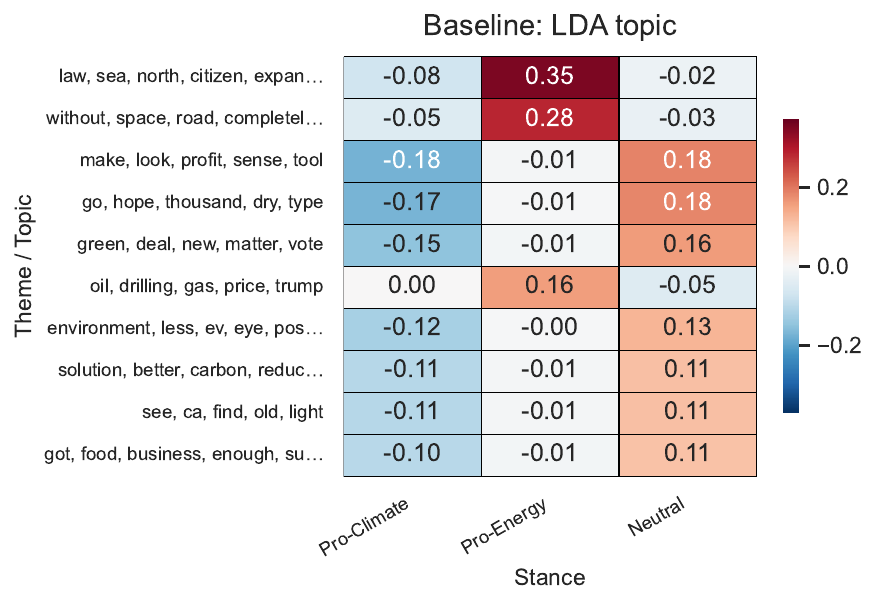}
  \caption{\textbf{Baseline}: LDA.}\label{fig:stance-heatmap-bsky-lda}
\end{subfigure}%
\begin{subfigure}{.9\columnwidth}
  \centering
  \includegraphics[width=\textwidth]{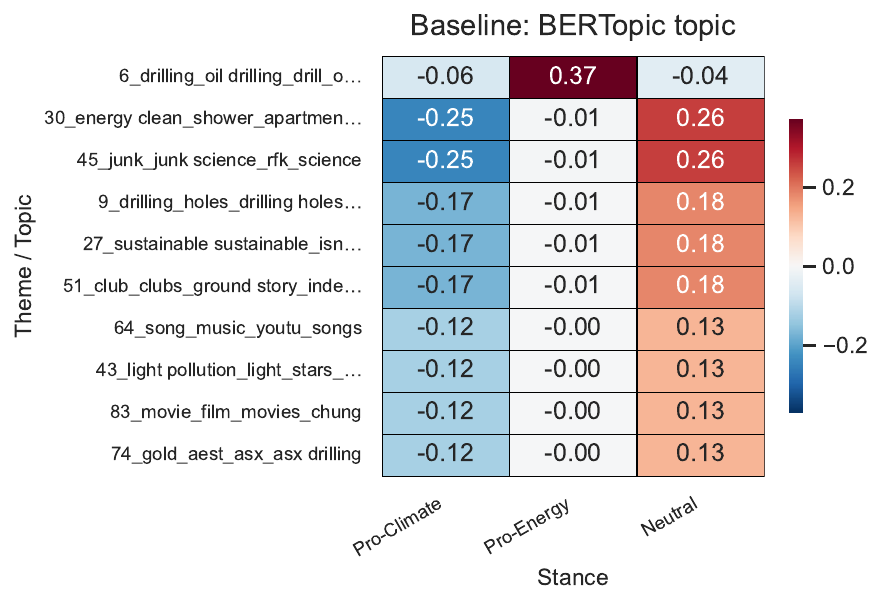}
  \caption{\textbf{Baseline}: BERTopic.}\label{fig:stance-heatmap-bsky-bert}
\end{subfigure}
\begin{subfigure}{.9\columnwidth}
  \centering
  \includegraphics[width=\textwidth]{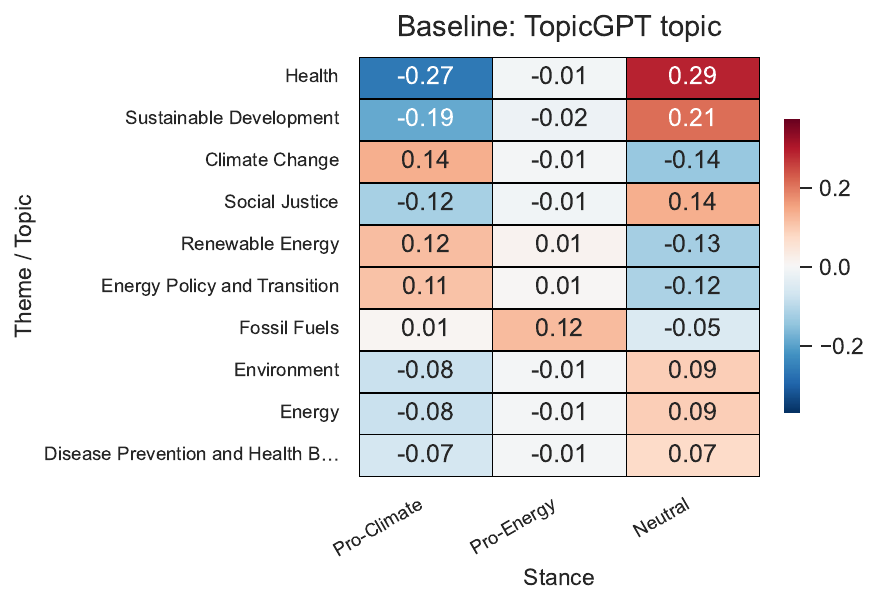}
  \caption{\textbf{Baseline}: TopicGPT.}\label{fig:stance-heatmap-bsky-topicgpt}
\end{subfigure}
\begin{subfigure}{.9\columnwidth}
  \centering
  \includegraphics[width=\textwidth]{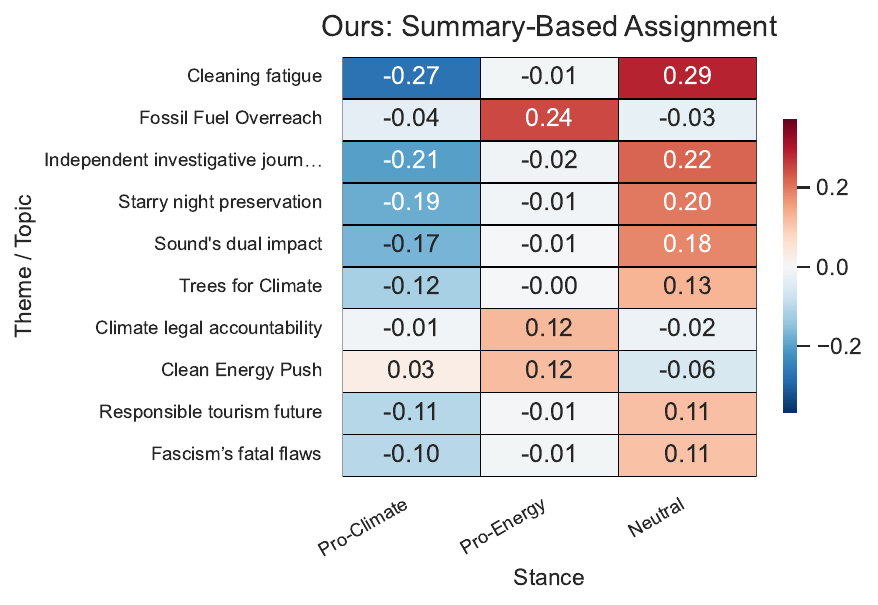}
  \caption{\textbf{Ours}: LLM Summary.}\label{fig:stance-heatmap-bsky-summary}
\end{subfigure}%
\begin{subfigure}{.9\columnwidth}
  \centering
  \includegraphics[width=\textwidth]{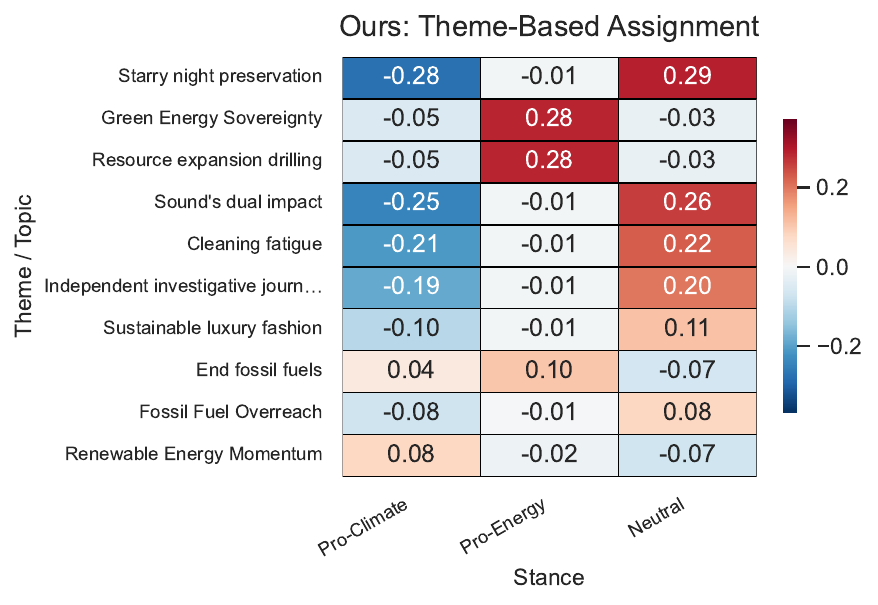}
  \caption{\textbf{Ours}: LLM Theme.}\label{fig:stance-heatmap-bsky-theme}
\end{subfigure}
\caption{Correlations between themes and stances for Bluesky. }
\label{fig:correl_bsky}
\end{figure*}

\begin{figure*}
\centering
\begin{subfigure}{.9\columnwidth}
  \centering
  \includegraphics[width=\textwidth]{figures/stance_meta/stance_heatmap_meta_baseline_lda_topic.pdf}
  \caption{\textbf{Baseline}: LDA.}\label{fig:stance-heatmap-meta-lda-full}
\end{subfigure}%
\begin{subfigure}{.9\columnwidth}
  \centering
  \includegraphics[width=\textwidth]{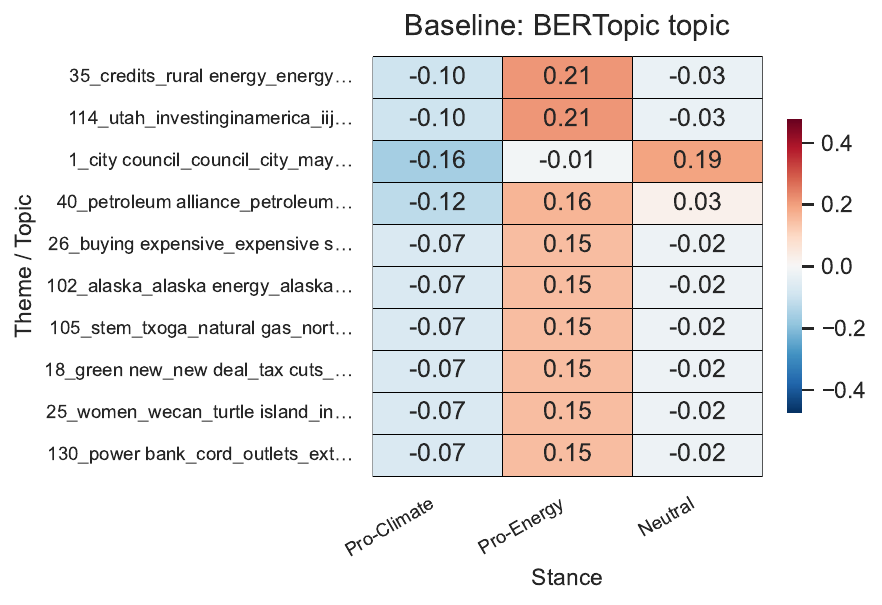}
  \caption{\textbf{Baseline}: BERTopic.}\label{fig:stance-heatmap-meta-bert-full}
\end{subfigure}
\begin{subfigure}{.9\columnwidth}
  \centering
  \includegraphics[width=\textwidth]{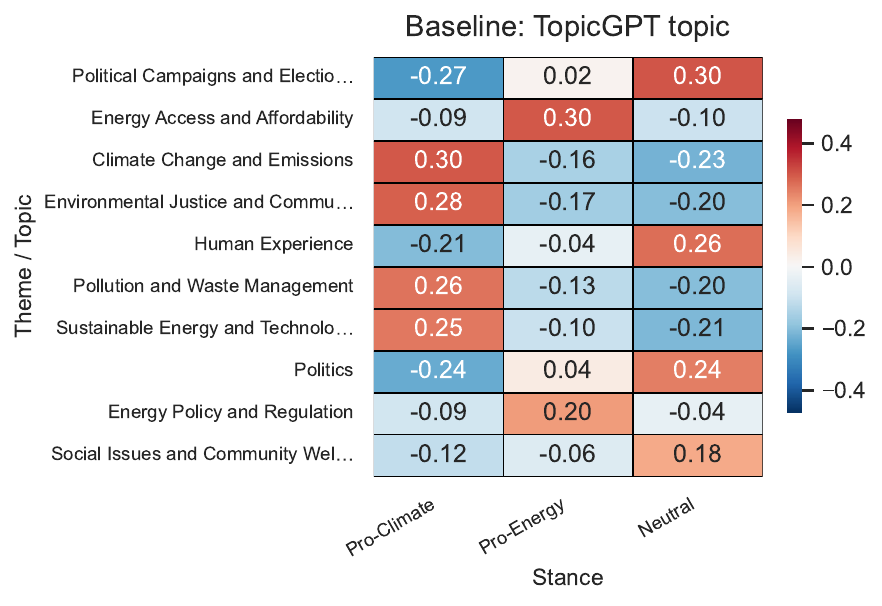}
  \caption{\textbf{Baseline}: TopicGPT.}\label{fig:stance-heatmap-meta-topicgpt-full}
\end{subfigure}
\begin{subfigure}{.9\columnwidth}
  \centering
  \includegraphics[width=\textwidth]{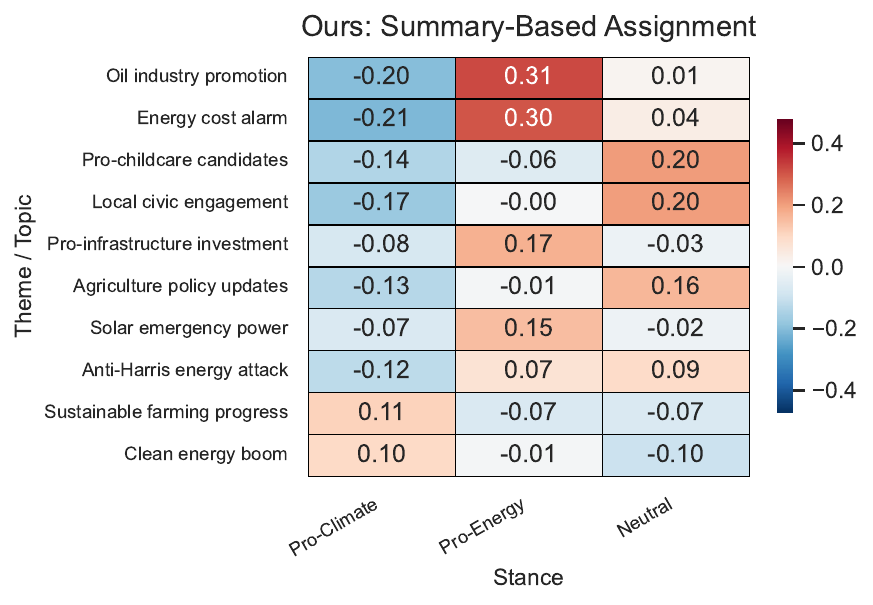}
  \caption{\textbf{Ours}: LLM Summary.}\label{fig:stance-heatmap-meta-summary-full}
\end{subfigure}%
\begin{subfigure}{.9\columnwidth}
  \centering
  \includegraphics[width=\textwidth]{figures/stance_meta/stance_heatmap_meta_ours_theme-based_assignment.pdf}
  \caption{\textbf{Ours}: LLM Theme.}\label{fig:stance-heatmap-meta-theme-full}
\end{subfigure}
\caption{Correlations between themes and stances for Meta (all baselines and methods).}
\label{fig:correl_meta_full}
\end{figure*}

Figs.~\ref{fig:correl_bsky} and ~\ref{fig:correl_meta_full} report
theme--stance correlations for all five methods on Meta and Bluesky,
extending the LDA and Text-to-Theme panels shown in the main text. Note that we manually label $500$ Meta texts and $500$ Bluesky texts as Pro-Climate, Pro-Energy, or Neutral. Although both platforms lean heavily \textit{Pro-Climate}, we see a more balanced structure for Meta than for Bluesky. The breakdown is as follows: for Meta, we have $46$ Pro-Energy, $101$ Neutral, and $353$ Pro-Climate texts. For Bluesky, we have $4$ Pro-Energy, $55$ Neutral, and $441$ Pro-Climate texts. Organic climate discourse on Bluesky is essentially a Pro-Climate-dominated space, consistent with the systemic-critique orientation observed in \S\ref{sec:thematic}. 

\paragraph{Meta.} The two probabilistic baselines yield diffuse,
low-magnitude correlations: LDA peaks near $0.22$ and BERTopic near
$0.21$, with no clear separation between \textit{Pro-Climate} and
\textit{Pro-Energy} stances. TopicGPT is the strongest baseline: its
substantive climate topics align with \textit{Pro-Climate}
(\textit{Climate Change and Emissions} $0.30$, \textit{Environmental
Justice} $0.28$, \textit{Pollution and Waste Management} $0.26$) and its
supply-side topics with \textit{Pro-Energy} (\textit{Energy Access and
Affordability} $0.30$, \textit{Energy Policy and Regulation} $0.20$).
However, much of its signal is carried by broad categories that load onto
\textit{Neutral} (\textit{Political Campaigns and Elections},
\textit{Politics}, \textit{Human Experience}), reflecting a generic
high-level taxonomy rather than dataset-specific structure. Our variants
produce sharper alignment on the \textit{Pro-Energy} pole, where
pro-fossil themes map strongly to \textit{Pro-Energy} (\textit{Oil
industry promotion} $0.48$ for Text-to-Theme, $0.31$ for Text-to-Summary;
\textit{Energy cost alarm} $0.27$--$0.30$). We also see moderate
alignment of pro-climate themes with \textit{Pro-Climate} (\textit{Clean
energy advocacy}, \textit{Big Oil accountability}), and of civic or
non-energy themes with \textit{Neutral} (\textit{Local civic engagement},
\textit{Pro-childcare candidates}). Text-to-Theme is marginally more
concentrated than Text-to-Summary. Our derived themes carry a comparable
number of \textit{Neutral} labels to TopicGPT, but these are more diffuse,
without TopicGPT's reliance on broad catch-all categories.

\paragraph{Bluesky.} Bluesky exhibits extreme stance imbalance ($4$
\textit{Pro-Energy}, $55$ \textit{Neutral}, $441$ \textit{Pro-Climate}),
which limits the stability of theme--stance correlations relative to
Meta. LDA and BERTopic exhibit more variable correlations on Bluesky than
on Meta, but these are more likely driven by noise than by coherent
topic--stance alignment. Our LLM-based methods show reduced absolute
correlation magnitudes relative to Meta but remain directionally
consistent. TopicGPT patterns more like our methods than like LDA and
BERTopic: its climate- and energy-relevant topics are directionally
coherent (\textit{Climate Change}, \textit{Renewable Energy}, and
\textit{Energy Policy} correlate positively with \textit{Pro-Climate},
$0.11$--$0.14$; \textit{Fossil Fuels} is the sole topic with a positive
\textit{Pro-Energy} correlation, $0.12$). However, its strongest
correlations are carried by broad, climate-peripheral topics
(\textit{Health}, \textit{Sustainable Development}) that load onto
\textit{Neutral}, reflecting its reliance on a generic high-level
taxonomy rather than the dataset-specific themes our methods recover.

\subsection{Bluesky: LIWC Event Dynamics}
\label{app:liwc-event-bluesky}

\begin{figure}[t]
\centering
\begin{subfigure}{\linewidth}
    \includegraphics[width=\linewidth]{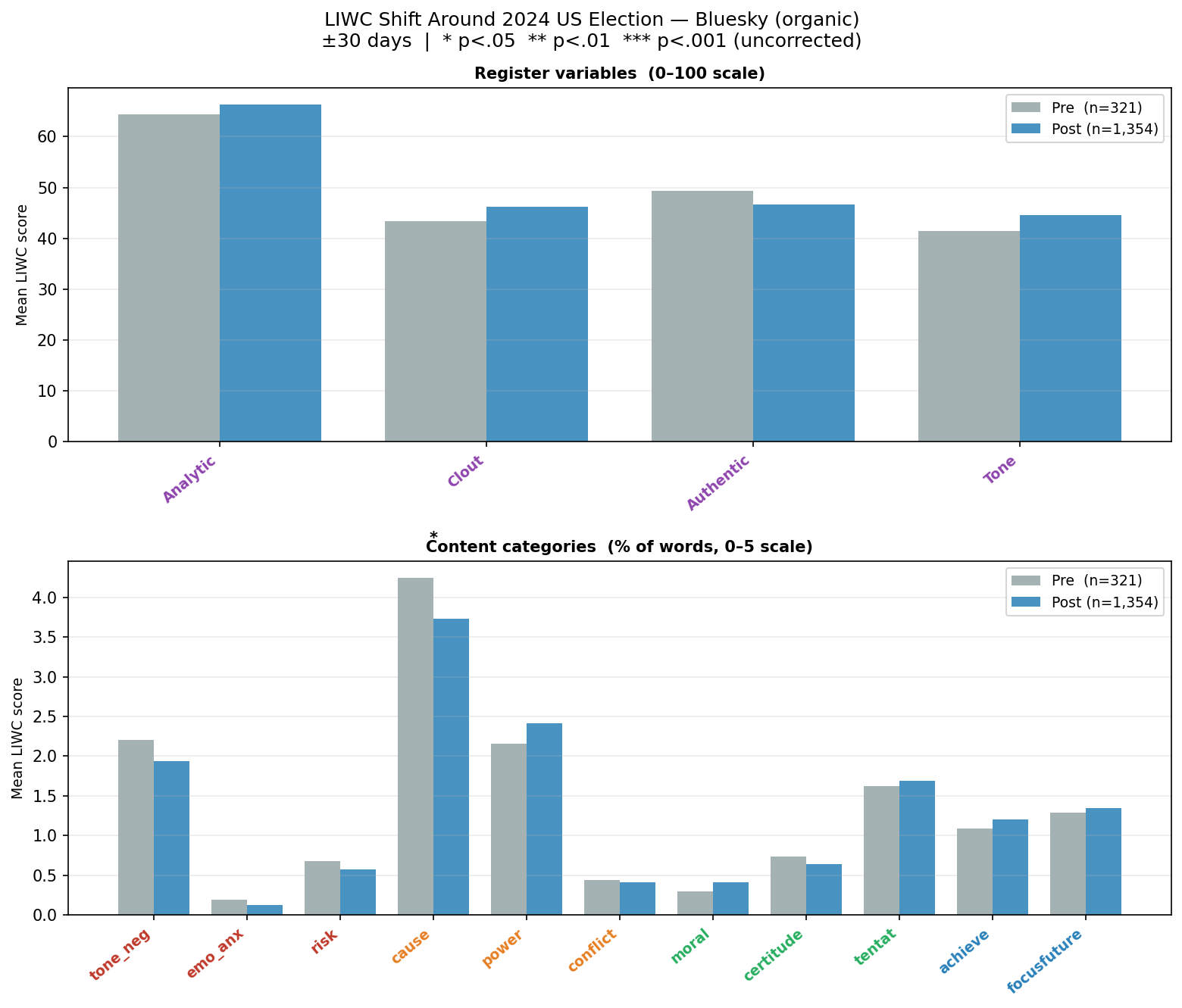}
    \caption{LIWC shift $\pm 30$ days around U.S. election (Bluesky).}
\end{subfigure}
\vspace{0.3em}
\begin{subfigure}{\linewidth}
    \includegraphics[width=\linewidth]{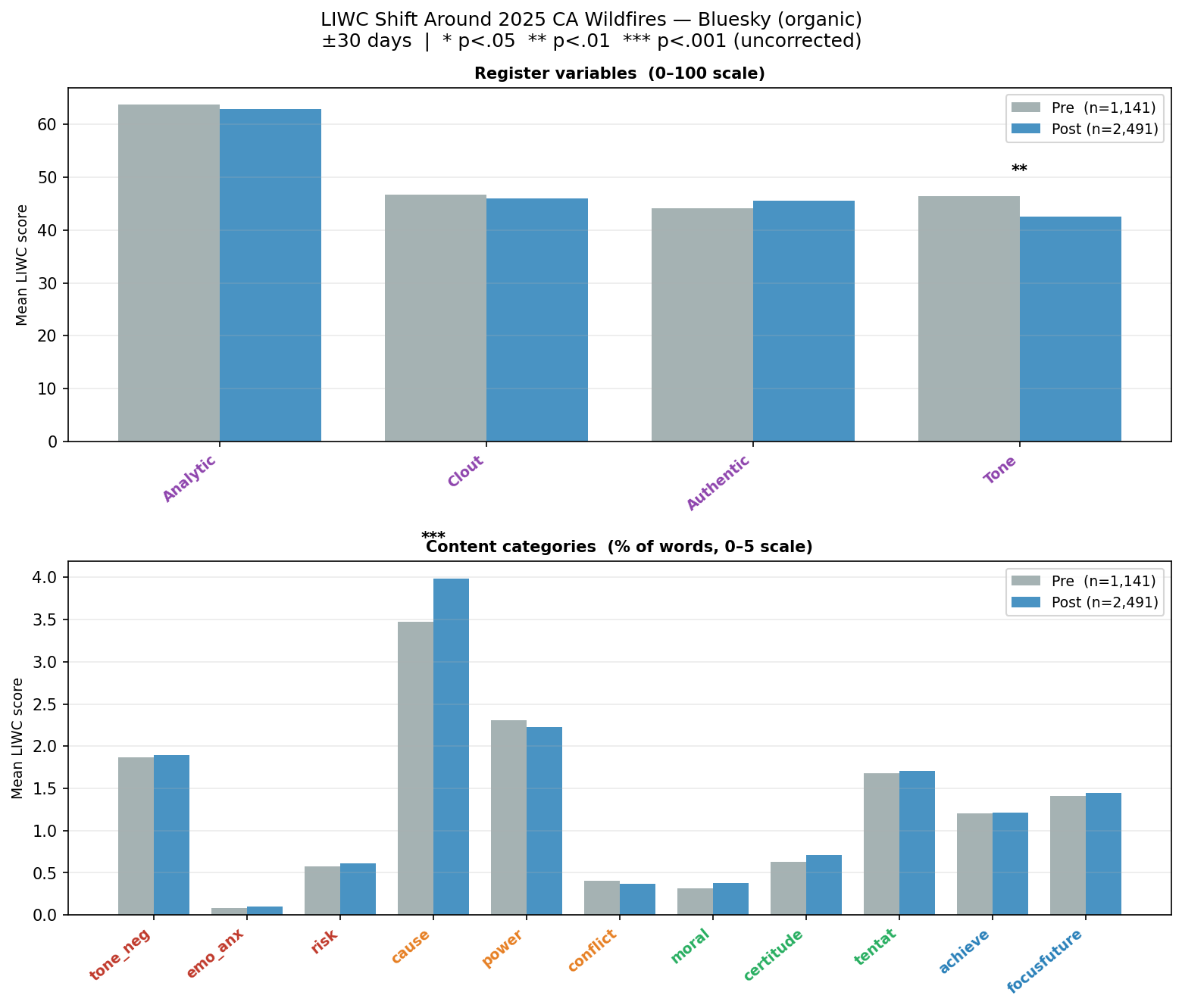}
    \caption{LIWC shift $\pm 30$ days around CA wildfires (Bluesky).}
\end{subfigure}
\caption{LIWC content categories and register variables around the two 
events for Bluesky. Election: \textit{cause} declines despite post-volume 
expansion. Wildfires: \textit{Tone} declines and \textit{cause} increases.}
\label{fig:liwc-bluesky-events}
\end{figure}

Fig.~\ref{fig:liwc-bluesky-events} shows the LIWC shift around both 
2024 U.S. election and 2025 Southern California wildfire events. 
We can see that the only significant shift in the U.S. election event is \textit{Cause}, which is reduced post-election. However, surrounding the wildfire event, \textit{Cause} increases significantly and \textit{Tone} decreases, suggesting that wildfire discourse is characterized by more negative affect and increased causal reasoning, consistent with crisis-driven language.

\subsection{Meta: U.S. Election Event Analysis}
\label{app:meta-election}

\begin{figure}[h]
\centering
\begin{subfigure}[t]{0.48\linewidth}
    \centering
    \includegraphics[width=\linewidth]{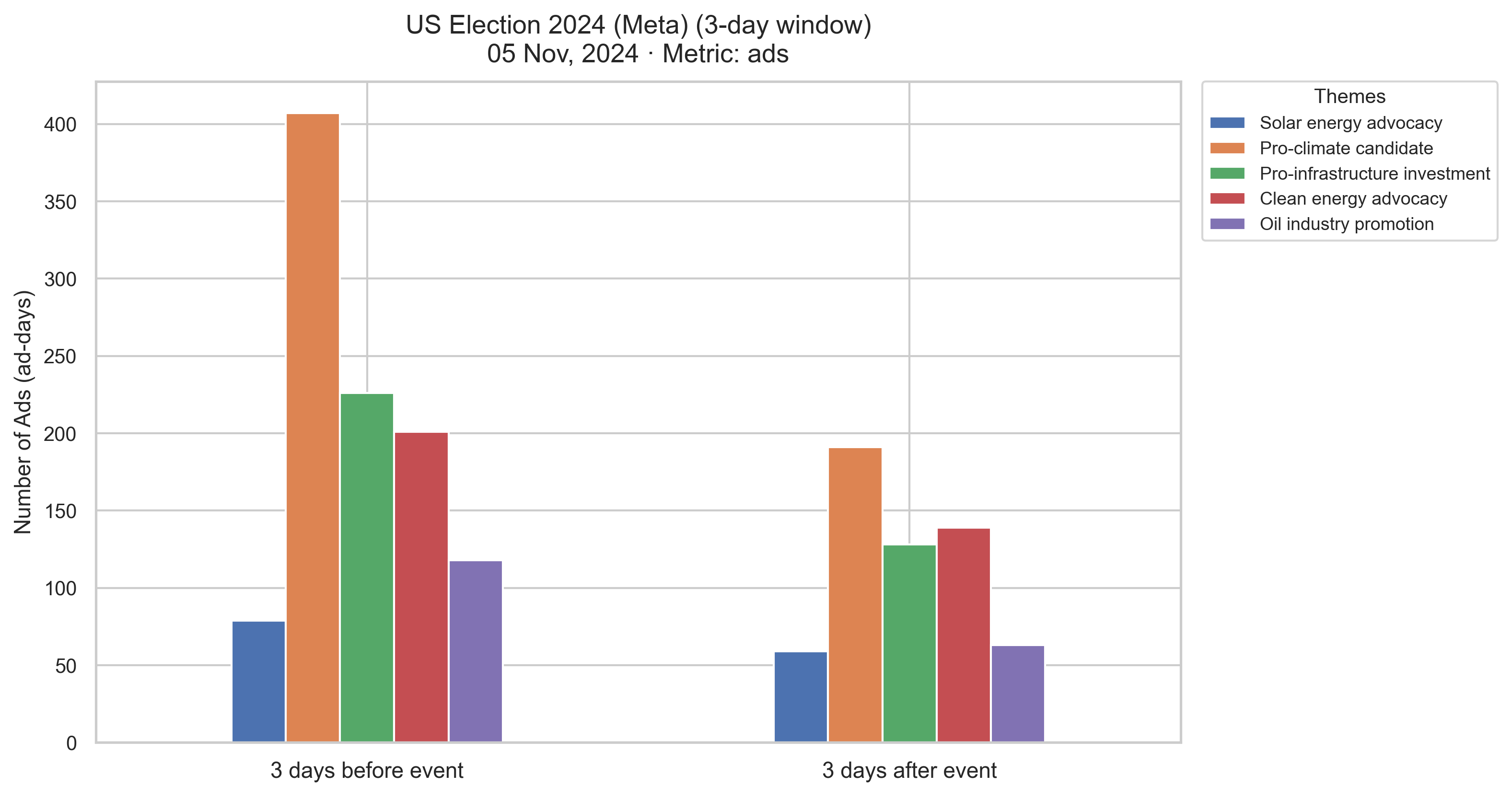}
    \caption{Total number of ads.}
\end{subfigure}
\hfill
\begin{subfigure}[t]{0.48\linewidth}
    \centering
    \includegraphics[width=\linewidth]{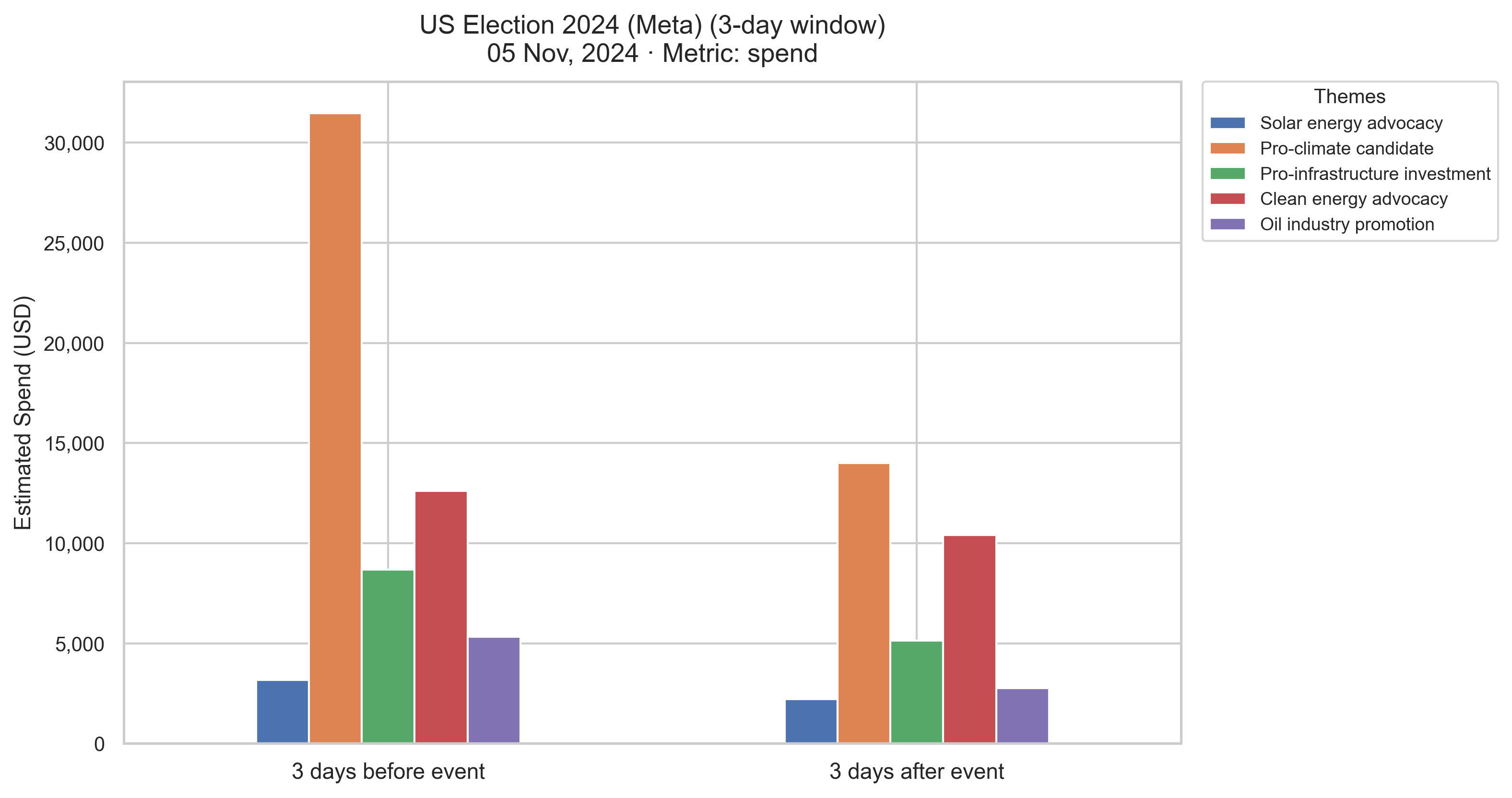}
    \caption{Total spend.}
\end{subfigure}
\vspace{0.3em}
\begin{subfigure}[t]{0.48\linewidth}
    \centering
    \includegraphics[width=\linewidth]{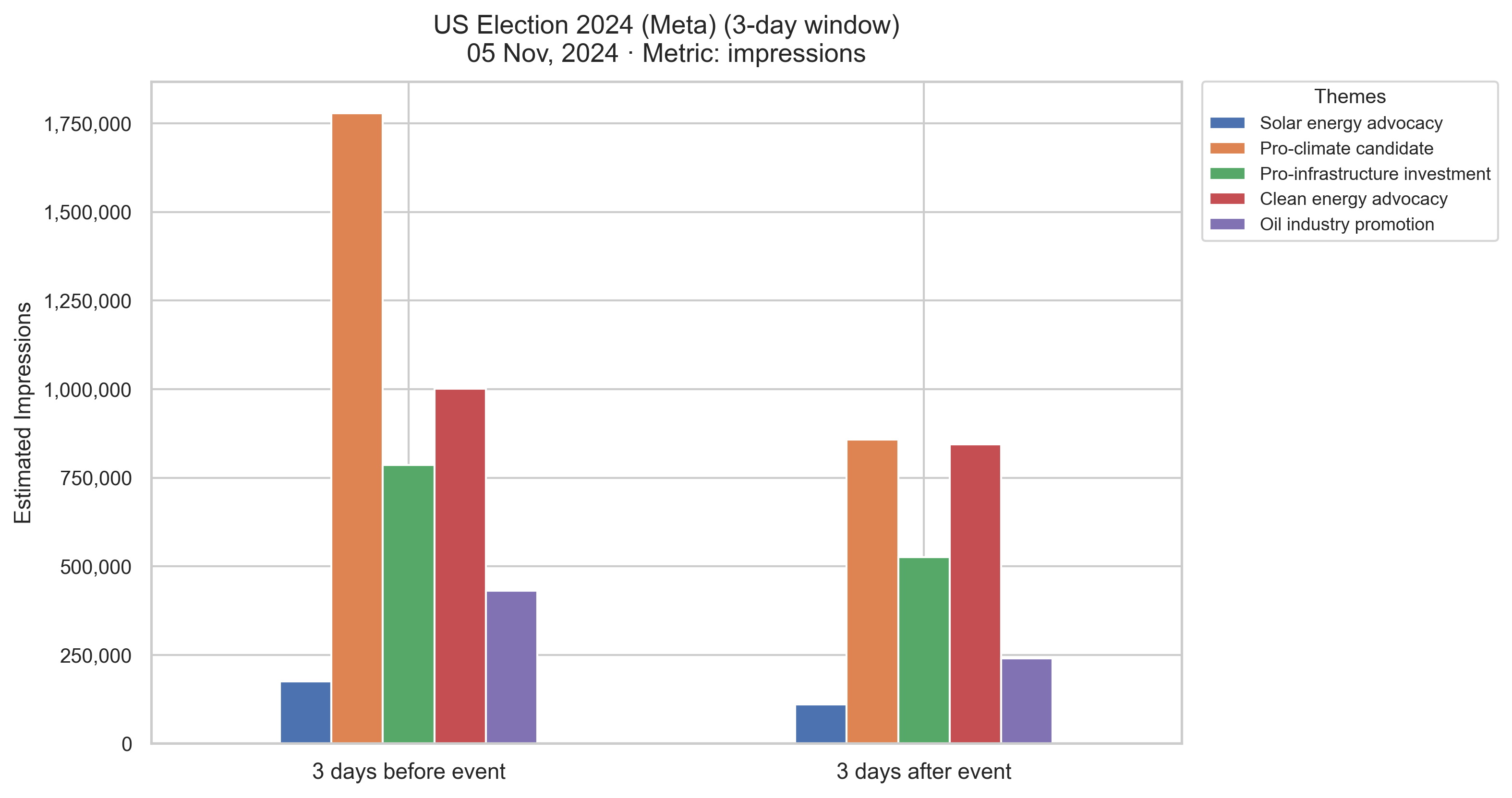}
    \caption{Total impressions.}
\end{subfigure}
\caption{Meta advertising activity $\pm 3$ days around the November 2024 U.S. election.}
\label{fig:election_full}
\end{figure}

\paragraph{Volume, spend, and impressions.}
Fig.~\ref{fig:election_full} reports total number of ads, total 
estimated spend, and total impressions for the five themes with the 
largest before/after shifts in a $\pm 3$-day window around the 
election. All three metrics decline, suggestion a reduction in overall advertising activity post-election.

\paragraph{Longer-horizon shifts.} Within a $30$-day before/after 
window, the theme \textit{Wildfire recovery advocacy} disappears 
entirely following the election. This change may loosely coincide with 
the electoral transition but is also potentially driven by the 
seasonal timing of wildfire-related advertising.

\begin{figure}[t]
    \centering
    \includegraphics[width=1.0\linewidth]{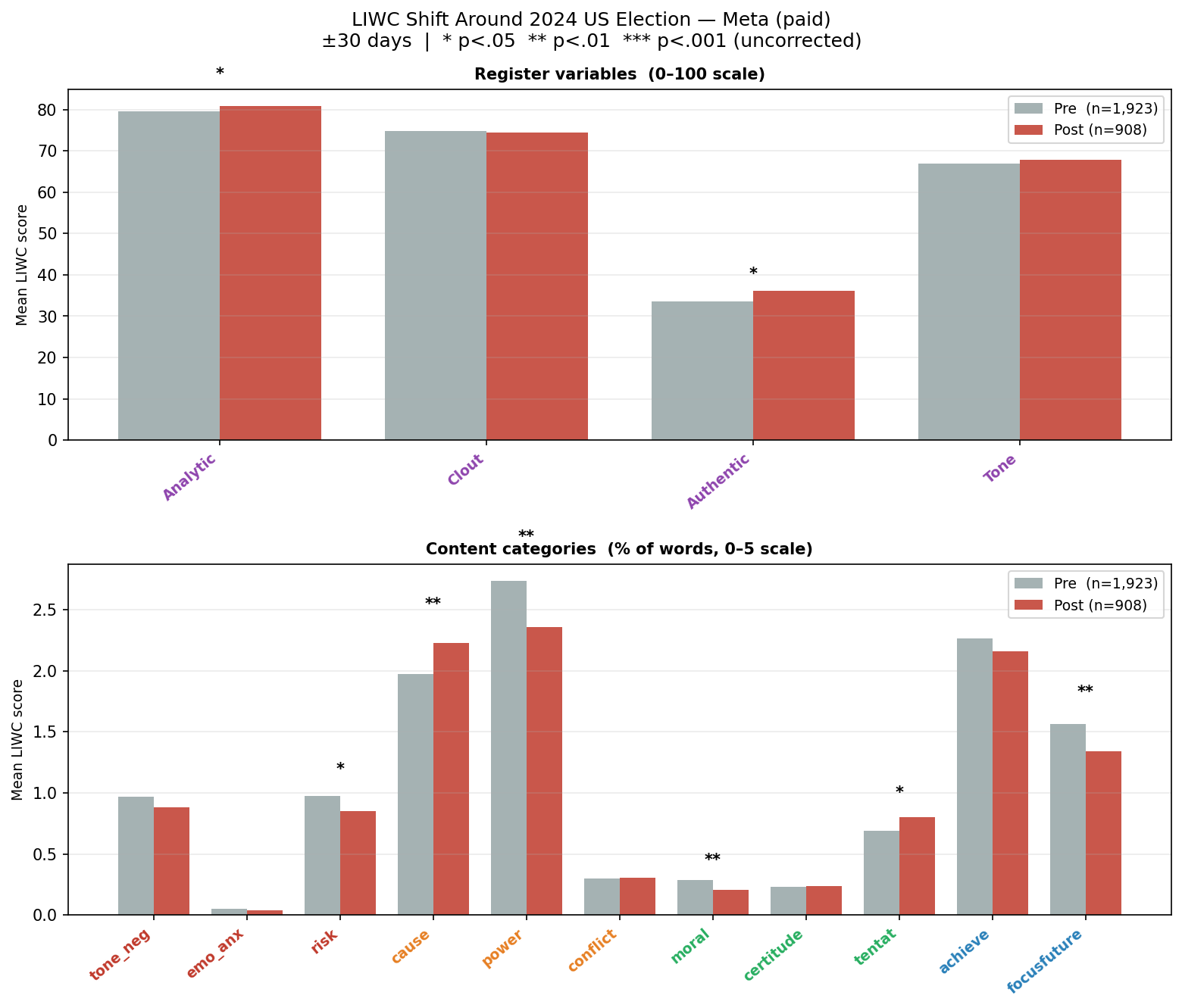}
    \caption{LIWC content categories and register variables $\pm 30$ days 
    around the U.S.\ election (Meta).}
    \label{fig:liwc-election-meta}
\end{figure}

\paragraph{LIWC dynamics.}
Fig.~\ref{fig:liwc-election-meta} shows the LIWC shift around the U.S. election. We see that \textit{moral} ($p<.01$), \textit{risk} ($p <.05$), \textit{power}  ($p<.05$), and \textit{focusfuture} 
($p<.01$) all decline, whereas \textit{Cause} and \textit{tentat} increase significantly. In terms of register variables, there is increase in \textit{Authentic} ($p<.01$) and \textit{Analytic} ($p<.01$). Overall, this linguistic pattern suggests increased tentativeness, authenticity, and causal reasoning post-election. The decline in normative, authoritative, and future-focused language may reflect the rhetorical demands of pre-election campaigning.

\subsection{Meta: California Wildfires Event Analysis}
\label{app:meta-wildfire}

\begin{figure*}
\centering
\begin{subfigure}[t]{0.48\linewidth}
    \centering
    \includegraphics[width=\linewidth]{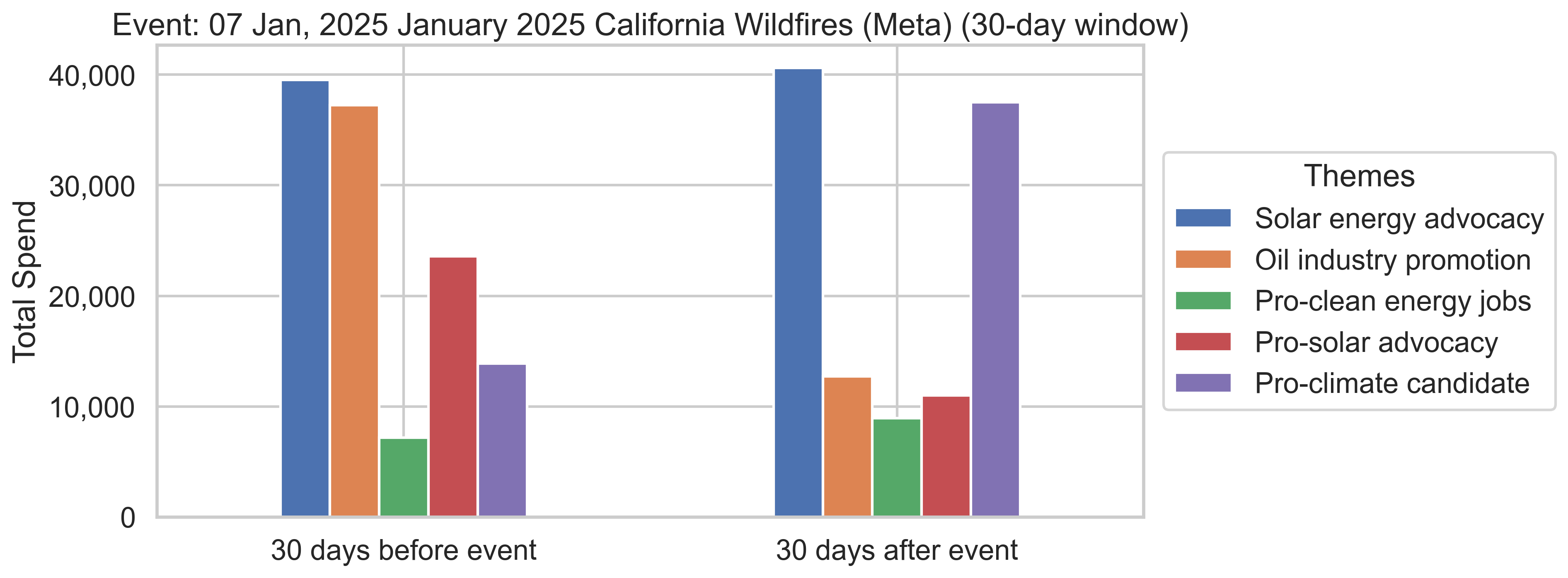}
    \caption{Total spend}
    \label{fig:kirk_spend}
\end{subfigure}
\hfill
\begin{subfigure}[t]{0.48\linewidth}
    \centering
    \includegraphics[width=\linewidth]{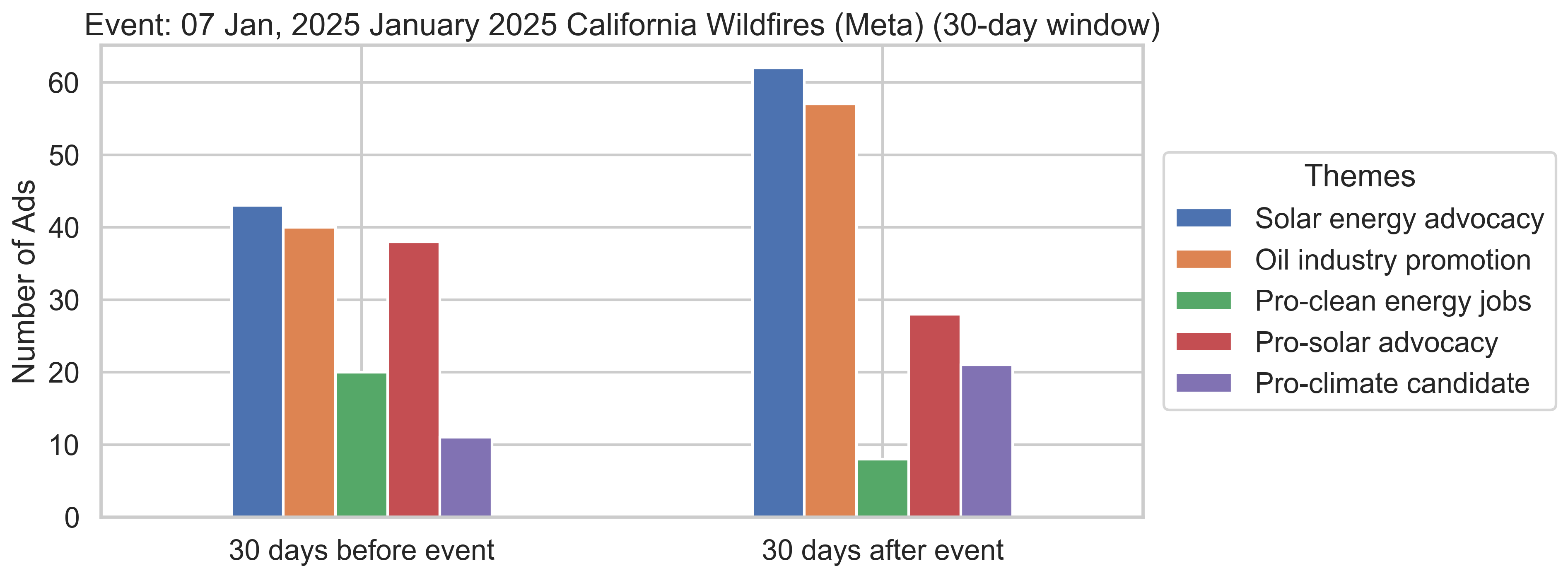}
    \caption{Total number of ads}
    \label{fig:kirk_ads}
\end{subfigure}
\\[1ex]
\begin{subfigure}[t]{0.48\linewidth}
    \centering 
    \includegraphics[width=\linewidth]{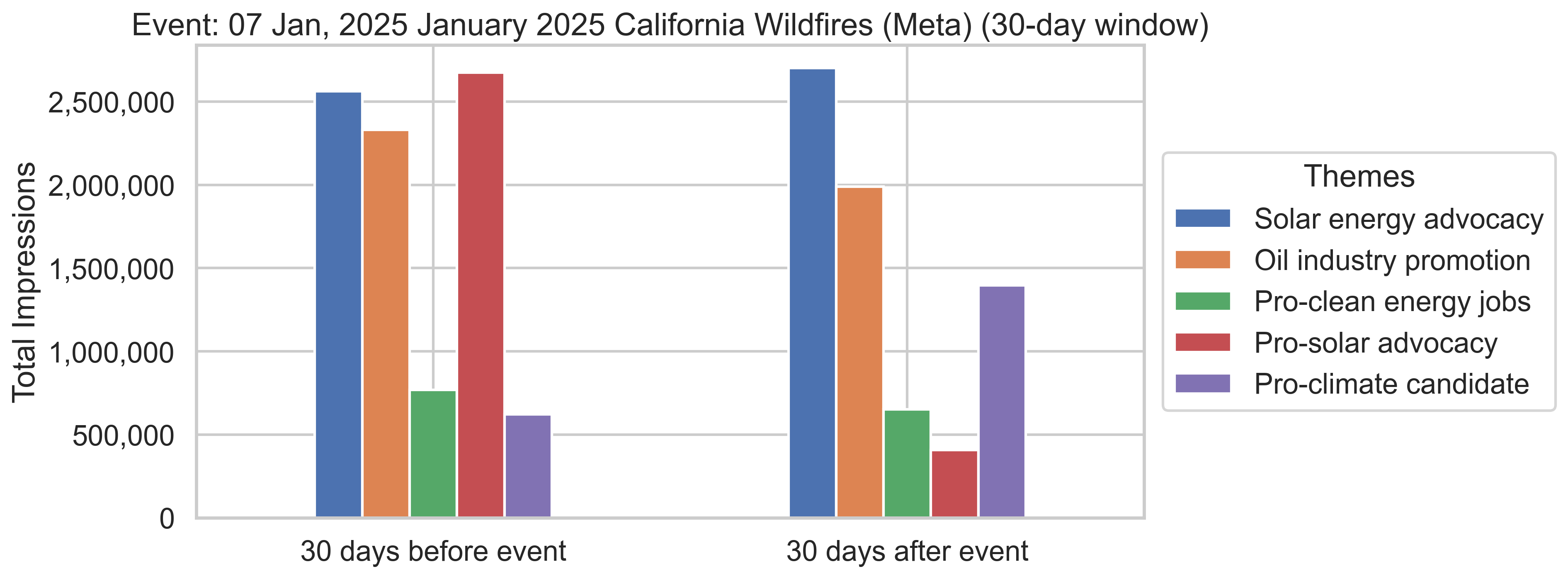}
    \caption{Total number of impressions}
    \label{fig:kirk_impressions}
\end{subfigure}
\begin{subfigure}[t]{0.48\linewidth}
    \centering 
    \includegraphics[width=\linewidth]{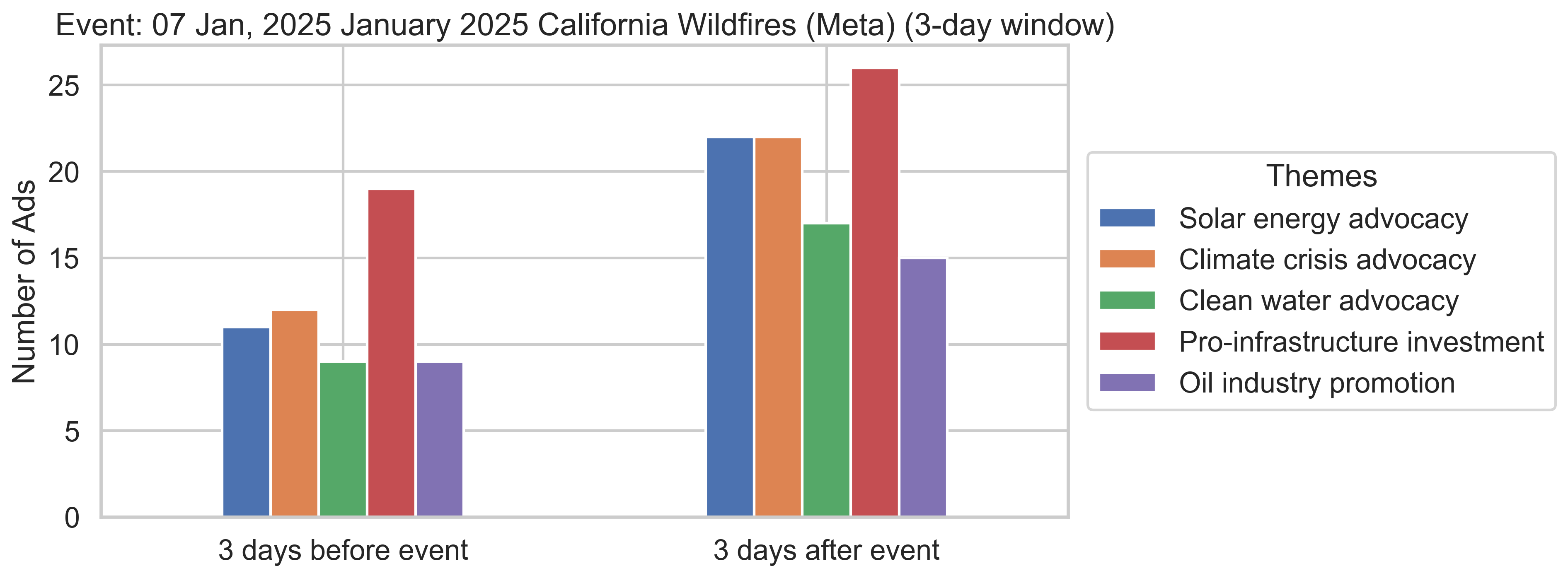}
    \caption{Total number of ads (Short-Horizon)}
    \label{fig:wildfire_shorthorizon}
\end{subfigure}

\caption{Advertising activity before and after the Southern California wildfires: (a) total spend, (b) total number of ads, and (c) total number of impressions.}
\label{fig:kirk_activity}
\end{figure*}

For the January 2025 Southern California wildfires, we utilize an
analysis window of $30$ days before and after the event onset to cover 
the month-long duration of the event. The five themes with the largest 
pre/post shifts in ad volume are \textit{Solar energy advocacy}, 
\textit{Oil industry promotion}, \textit{Pro-clean energy jobs}, 
\textit{Pro-solar advocacy}, and \textit{Pro-climate candidate}. As 
shown in Fig.~\ref{fig:kirk_activity}, the directionality across 
metrics is less uniform than around the election. Ad volume increases 
for \textit{Solar energy advocacy}, \textit{Oil industry promotion}, 
and \textit{Pro-climate candidate}, while \textit{Pro-solar advocacy} 
and \textit{Pro-clean energy jobs} decline. Over a short $3$-day 
horizon, top themes differ from the $30$-day window---\textit{Pro-infrastructure 
investment} and \textit{Climate crisis advocacy} are most active.

\begin{figure}
    \centering
    \includegraphics[width=1.0\linewidth]{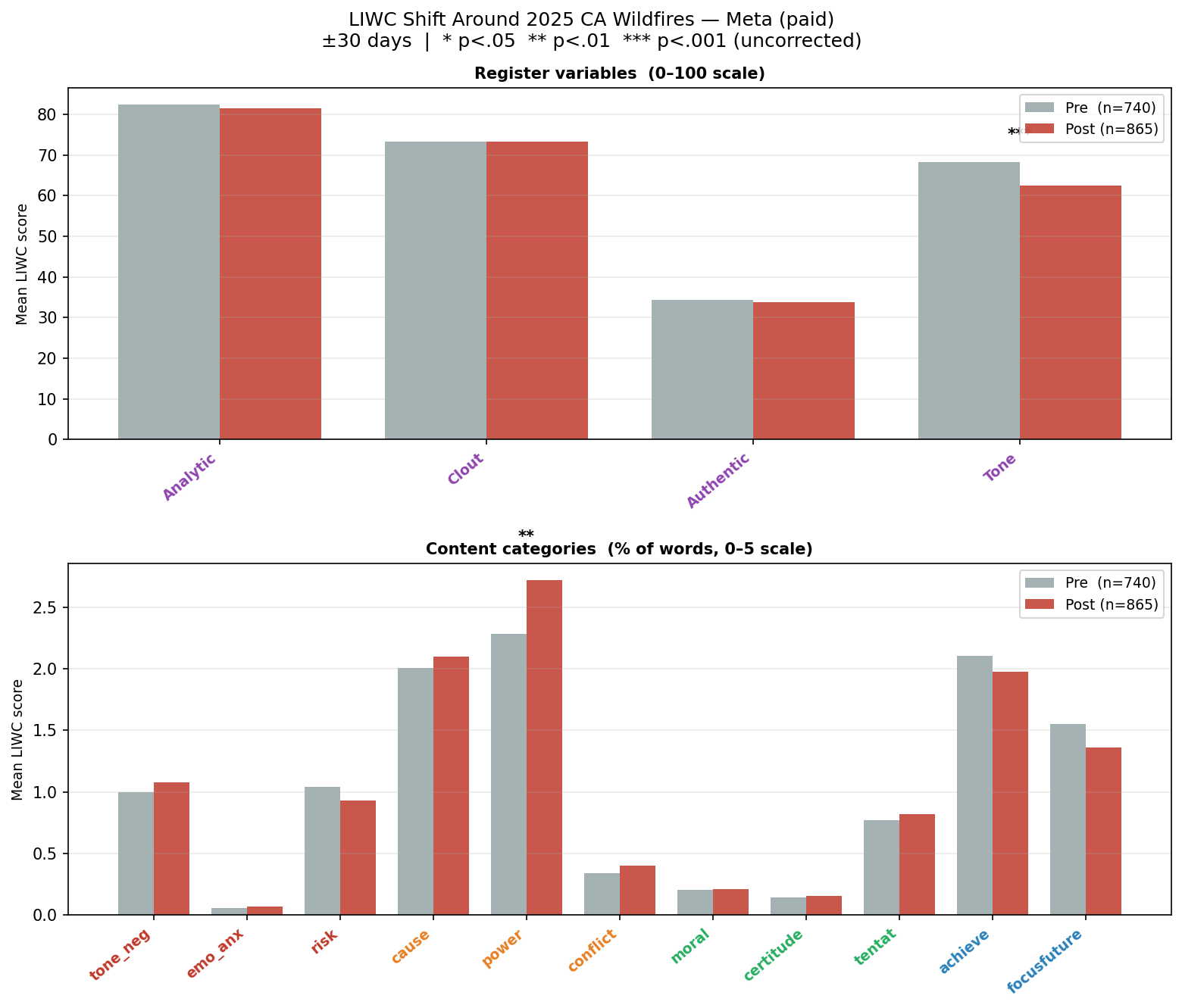}
    \caption{LIWC content categories and register variables 30 days before/after 2025 CA Wildfires for Meta dataset.}
    \label{fig:liwc-wildfire-meta}
\end{figure}

\paragraph{LIWC dynamics.}
Fig.~\ref{fig:liwc-wildfire-meta} shows the LIWC shift around the 
wildfires. \textit{Tone} declines noticeably post-event, suggesting 
a less positive emotional register. \textit{Power} language increases 
significantly post-wildfire ($p<.01$), pointing to a surge in language 
centered on authority, control, and systemic agency. Unlike the 
post-election pattern, \textit{focusfuture} and \textit{moral} remain 
stable, and \textit{risk} does not significantly increase despite the 
salience of the event.

\section{Details of Downstream Tasks}
\subsection{Downstream Task: Stance Prediction}
\label{app:stance}
In our stance prediction downstream task, the model is provided with three input variants: text only (\texttt{txt}), theme only (\texttt{thm}), and their concatenation (\texttt{txt+thm}), and is asked to predict the stance of the text (pro-climate, pro-energy, or neutral).
We split the data using \textbf{stratified} sampling to preserve stance label distributions across splits. Specifically, we first create an $80/20$ stratified split into training–validation and test sets. The training–validation portion is then further split using stratified sampling into training and validation sets ($80/20$). All splits are performed with a fixed random seed for reproducibility.

\begin{figure*}
\centering
\begin{subfigure}{\textwidth}
  \centering
  \includegraphics[width=\textwidth]{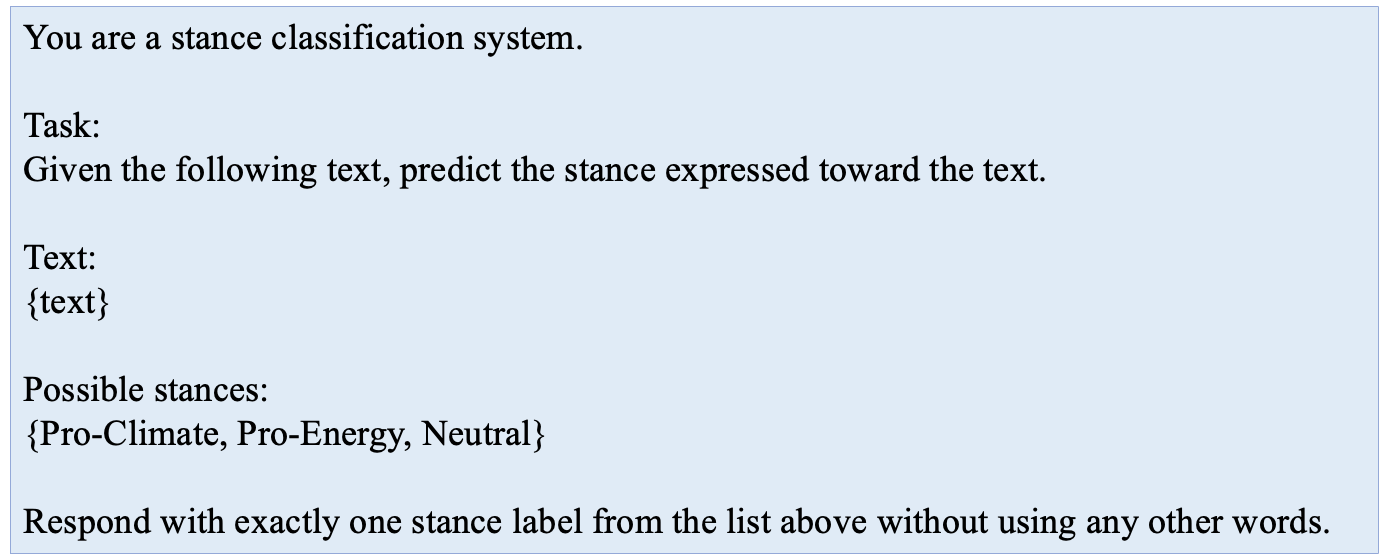}
  \caption{Text only.}
\label{fig:txt}
\end{subfigure}
\begin{subfigure}{\textwidth}
  \centering
  \includegraphics[width=\textwidth]{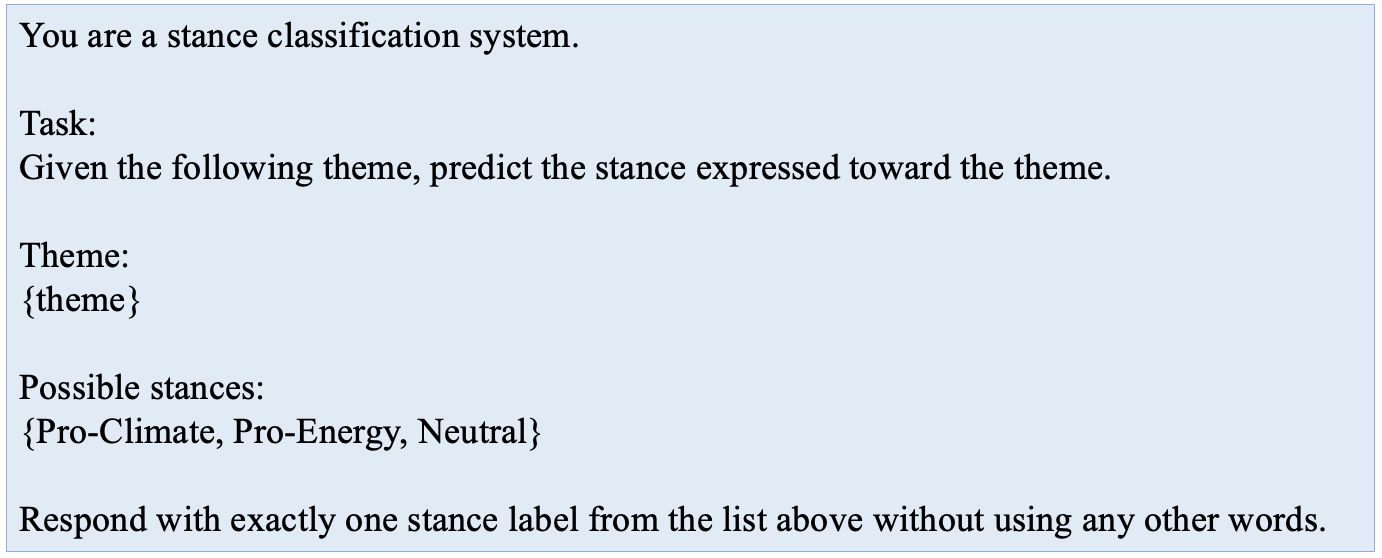}
  \caption{Theme only.}
\label{fig:thm}
\end{subfigure}
\begin{subfigure}{\textwidth}
  \centering
  \includegraphics[width=\textwidth]{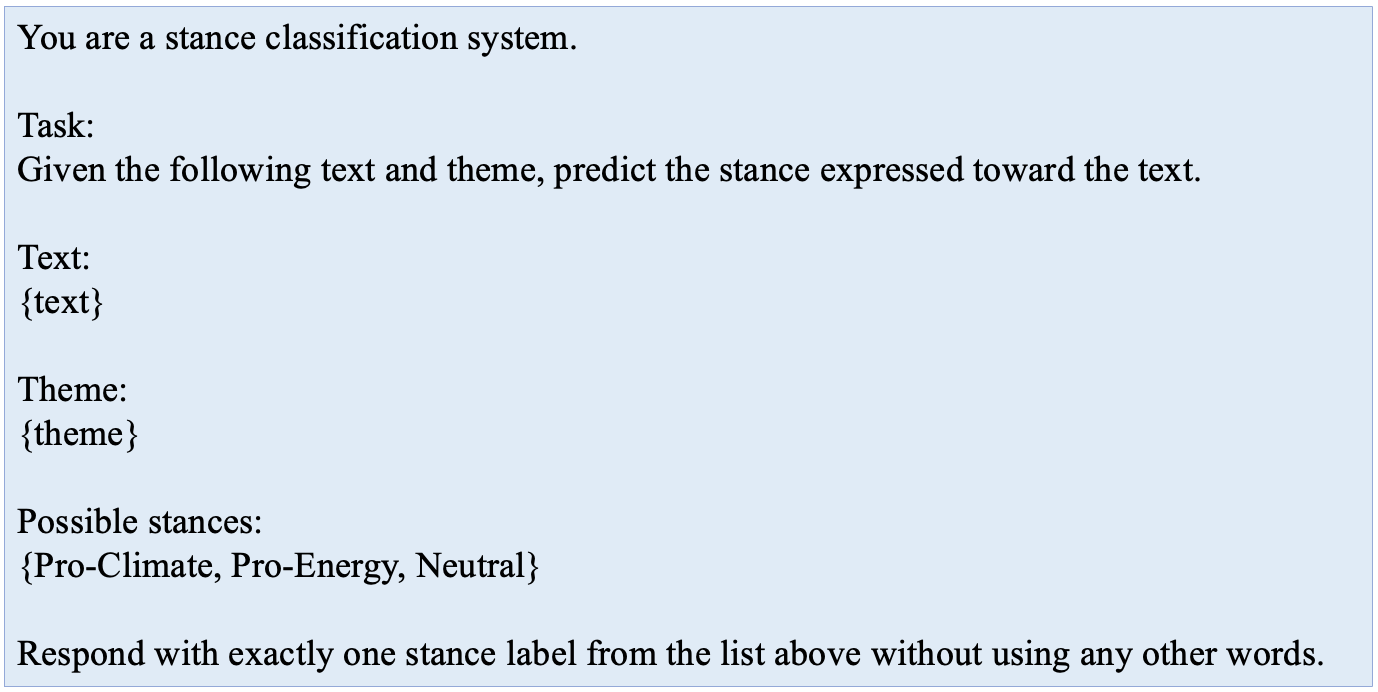}
  \caption{Text+Theme.}
\label{fig:txt_thm}
\end{subfigure}
\caption{Prompts for downstream task: stance prediction.}
\label{fig:pt_dowm}
\end{figure*}

We evaluate open-source LLMs including Llama 3.1 (llama-3.1-8b-instant) \cite{touvron2023llama} and kimi k2 (moonshotai/kimi-k2-instruct-0905) \cite{team2025kimi}. For both models, we use default inference settings with zero-shot prompting and run them via the Groq API\footnote{\url{https://wow.groq.com/}}. Prompts (all three input variants) for the stance prediction downstream task are given in Fig. \ref{fig:pt_dowm}.

In addition, we include a pre-trained language model, RoBERTa (roberta-base)~\cite{liu2019roberta}, as well as a simple logistic regression (LR) classifier~\cite{cox1958regression} trained on term frequency–inverse document frequency (tf-idf) features.

For RoBERTa, we use the `roberta-base' model, batch size = $8$, epoch = $3$, learning rate = $2e-5$, weight decay = $0.01$. Our early stopping criterion is the \textbf{lowest validation loss}. To run the stance classifier, we use a single GPU GeForce GTX $1080$ Ti GPU, with $4$ Intel Core $i5$-$7400$ CPU @ $3.00$ GHz processors, and it takes around $2$ minutes to run the model. 

As a lightweight baseline, we train a multinomial logistic regression classifier using TF-IDF features. For the combined setting, TF-IDF features are independently computed for text and theme and then concatenated. We tune the n-gram range and the regularization strength $C$. We consider unigram and short higher-order n-grams to capture both lexical cues and limited phrasal patterns, and select a compact search space ($\{(1,1), (1,2), (2,3)\}$ for Meta and $\{(1,1), (1,2), (2,3), (3,4)\}$ for Bluesky) to avoid overfitting under class imbalance. The regularization strength  $C$ is chosen from a logarithmic grid ($C \in \{0.1, 1.0, 10.0\}$ for Meta and $C \in \{0.01, 0.1, 1.0, 10.0, 100\}$ for Bluesky) to balance model flexibility and generalization. All hyperparameters are selected based on the performance of \textbf{validation} set.

\subsection{Downstream Task: Theme Guided Retrieval}
\label{app:tsne}
\begin{figure*}
\centering
\begin{subfigure}{.85\textwidth}
    \centering
    \includegraphics[width=\textwidth]{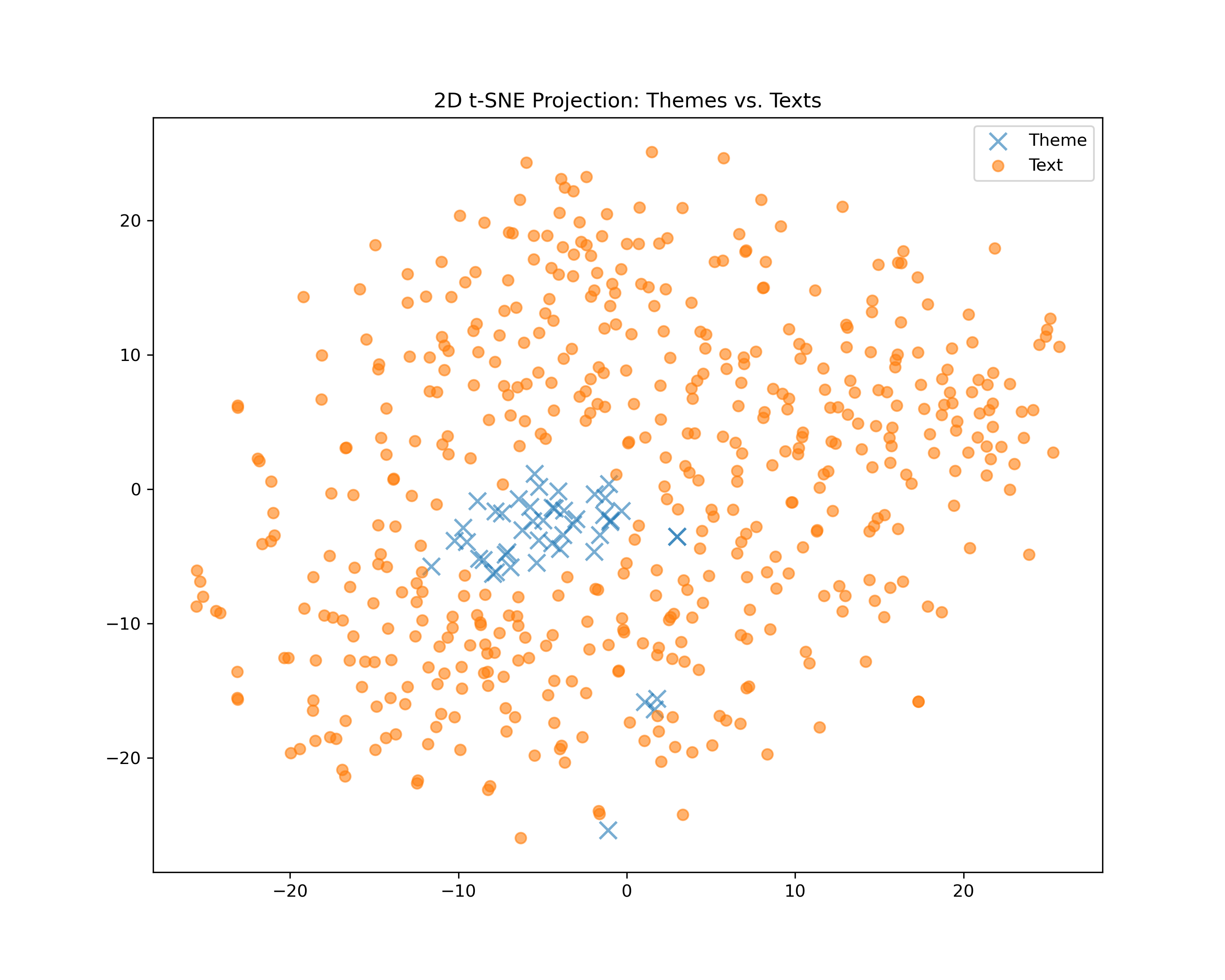}
    \caption{Meta}
    \label{fig:tsne_fb}
\end{subfigure}
\begin{subfigure}{.85\textwidth}
    \centering
    \includegraphics[width=\textwidth]{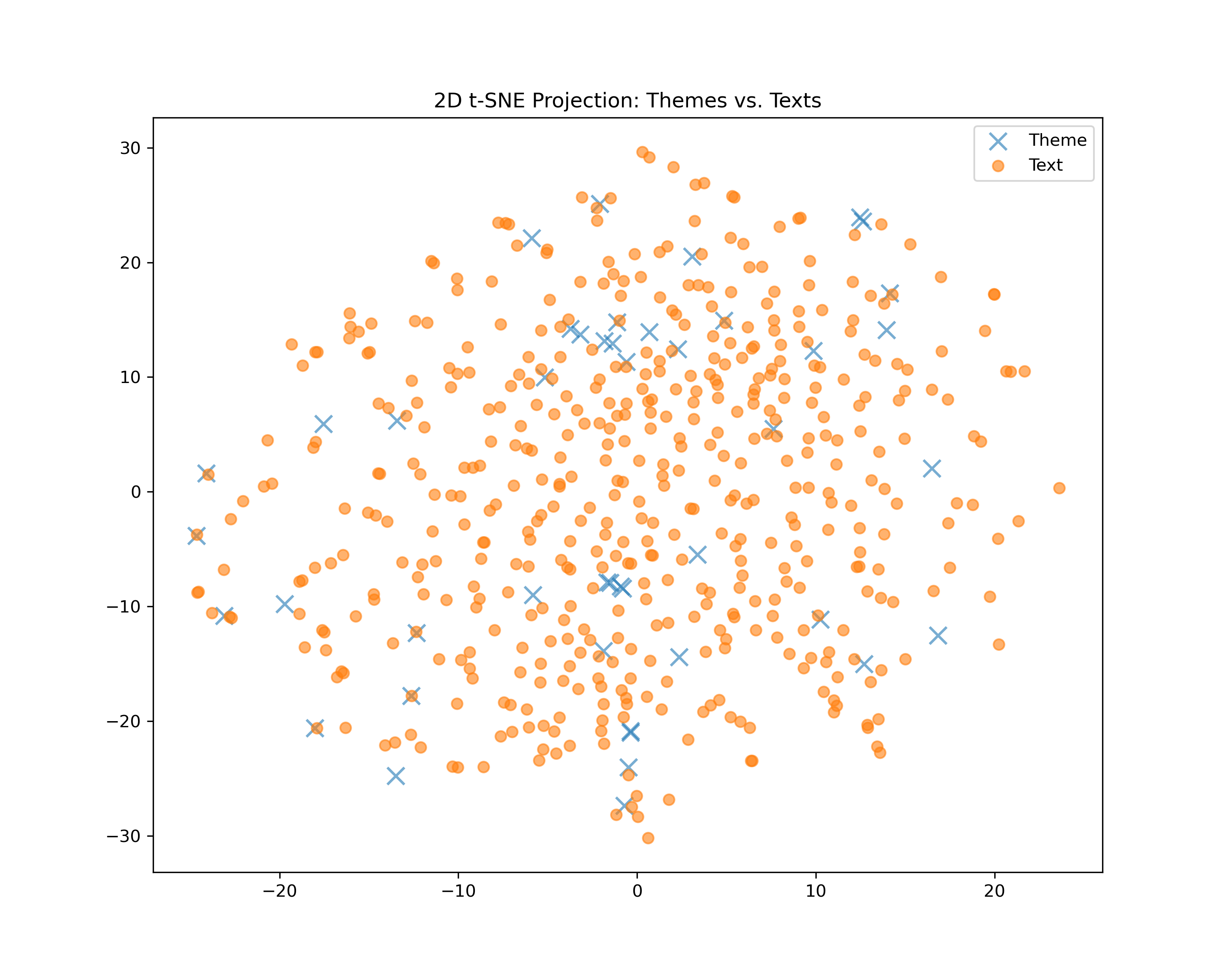}
    \caption{Bluesky}
    \label{fig:tsne_bsky}
\end{subfigure}
\caption{t-SNE projection of post and theme embeddings. Themes (blue \textcolor{blue}{$\times$}) tend to cluster near semantically aligned texts (orange \textcolor{orange}{$\bullet$}), validating theme-text embedding consistency in (a) Meta and (b) Bluesky.}
\label{fig:tsne_theme_text}
\end{figure*}

Fig. ~\ref{fig:tsne_theme_text} shows the t-SNE projection of text and theme embeddings for the same $500$ Meta ads and $500$ Bluesky posts. Meta’s themes and ads exhibit a denser, more centralized semantic structure, likely reflecting a narrower and coordinated message space (e.g., advertisements or campaigns). In contrast, Bluesky exhibits a broader semantic spread, indicating more diverse themes and discourse patterns characteristic of open social platforms.

Qualitative examples (Tables \ref{tab:fb_retrieval_examples} and \ref{tab:bsky_retrieval_examples}) show that theme-guided retrieval surfaces semantically relevant posts across platforms. Meta examples tend to cluster around coordinated messages, while Bluesky samples are more linguistically and topically diverse. Retrieved examples reflect stylistic and semantic differences across platforms, with Meta ads being more coordinated and campaign-like (Table \ref{tab:fb_retrieval_examples}), and Bluesky showcasing open-ended user discourse (Table \ref{tab:bsky_retrieval_examples}).

\begin{table*}
\centering
\begin{tabular}{p{0.28\textwidth} | p{0.65\textwidth}}
\toprule
\textbf{Theme} & \textbf{Top Retrieved Ad} \\
\midrule
Climate crisis advocacy &
Time \& time again we see how most of our elected officials continue to fail at protecting our health \& safety, which is why we need YOUR help to create the world that we want to live in. For more information on how to get involved, click the link in our bio to our landing page to sign up to be involved in upcoming rulemakings and engagement opportunities
\#environment \#climate \#legislation \#policy \#environment \#climatecrisis \#community \#love \#joy \#resilence \#pollution \#solutions \#action \#hope \#climatecrisis \#denver \#denvercolorado \\
\midrule
Ocean conservation call &
Our ocean is everything. From providing most of the oxygen we breathe to shielding us from the worst impacts of climate change, it’s our greatest ally. This Giving Season, your support helps us protect vulnerable marine ecosystems. Give now to multiply your impact for our oceans! \\
\midrule
Methane regulation push &
An emergency brake for our climate: cutting methane pollution is the best thing we can do to stop our planet from boiling. Climate activist Sophia Kianni explains. \\
\midrule
Energy cost alarm & Already struggling with your electrical bill? It might get worse — contact your senator today to tell them we need to keep American energy competitive. \\
\midrule
Oil industry promotion & For nearly five decades, the oil \& gas industry has been the backbone of Alaska's economy. Its impact includes:
Helping create 69,250 additional jobs, statewide
Our state and local governments collecting nearly \$4.4 in oil \& gas revenues
Over \$1 billion added being added to the Permanent Fund
Learn more about the impacts the oil \& gas industry has on Alaska in this video. \\
\bottomrule
\end{tabular}
\caption{Theme-guided retrieval examples from Meta. Ads tend to reflect coordinated or campaign-style messaging.}
\label{tab:fb_retrieval_examples}
\end{table*}

\begin{table*}
\centering
\begin{tabular}{p{0.28\textwidth} | p{0.65\textwidth}}
\toprule
\textbf{Theme} & \textbf{Top Retrieved Post} \\
\midrule
End fossil fuels &
The fossil fuel industry won't go down without a fight. It's too lucrative to keep trying to convince people they're essential to our survival. \\
\midrule
Greenwashing deception &
Authenticity sells. Aligning with nature = no greenwashing. It shows you walk the talk on sustainability and values. \\
\midrule
Science under attack &
We have no time to waste in opposing the worst of President Trump's Cabinet nominees with respect to science and the role it should play in keeping us safe. Urge your senators to vote NO on the `Junk Science Six'. \\
\midrule
Climate action collaboration & Climate action is health action. From reducing waste to supporting renewable energy and green spaces, every step we take protects our health and our planet. Learn how we can work together to champion climate resilience and safeguard our future. \\
\midrule
Sustainable luxury fashion & Sustainable fashion can be considered an antidote to fast fashion. It recognizes the economic and sociocultural contributions of the fashion industry while also advocating for the need to promote sustainability across all aspects of production and consumption. \\
\bottomrule
\end{tabular}
\caption{Theme-guided retrieval examples from Bluesky. Posts show more diverse, user-driven expression and framing.}
\label{tab:bsky_retrieval_examples}
\end{table*}


\end{document}